\newif\ifagibook\agibookfalse
\newif\ifpublished\publishedtrue

\documentclass[12pt,twoside]{article}
\usepackage{latexsym}
\usepackage{makeidx}
\makeindex

\def\,{\mskip 3mu} \def\>{\mskip 4mu plus 2mu minus 4mu} \def\;{\mskip 5mu plus 5mu} \def\!{\mskip-3mu}
\def\dispmuskip{\thinmuskip= 3mu plus 0mu minus 2mu \medmuskip=  4mu plus 2mu minus 2mu \thickmuskip=5mu plus 5mu minus 2mu}
\def\textmuskip{\thinmuskip= 0mu                    \medmuskip=  1mu plus 1mu minus 1mu \thickmuskip=2mu plus 3mu minus 1mu}
\textmuskip
\def\beq{\dispmuskip\begin{equation}}    \def\eeq{\end{equation}\textmuskip}
\def\beqn{\dispmuskip\begin{displaymath}}\def\eeqn{\end{displaymath}\textmuskip}
\def\bqa{\dispmuskip\begin{eqnarray}}    \def\eqa{\end{eqnarray}\textmuskip}
\def\bqan{\dispmuskip\begin{eqnarray*}}  \def\eqan{\end{eqnarray*}\textmuskip}

\newenvironment{keywords}{\centerline{\bf
Keywords}\vspace{0.5ex}\begin{quote}\small}{\par\end{quote}\vskip
1ex}

\renewenvironment{abstract}{\centerline{\bf
Abstract}\vspace{0.5ex}\begin{quote}\small}{\par\end{quote}\vskip
1ex}

\newtheorem{theorem}{Theorem}
\newtheorem{tablex}[theorem]{Table}
\newtheorem{figurex}[theorem]{Figure}
\newtheorem{claim}[theorem]{Claim}
\newtheorem{axiom}[theorem]{Axiom}
\newtheorem{assumption}[theorem]{Assumption}
\newtheorem{corollary}[theorem]{Corollary}
\newtheorem{lemma}[theorem]{Lemma}
\newtheorem{definition}[theorem]{Definition}

\def\ffigurex#1#2#3#4{{#4} \begin{figurex}[#2]\label{#1} #3 \end{figurex} }
\def\fclaim#1#2#3{\begin{claim}[#2]\label{#1} #3 \end{claim} }

\def\fassumption#1#2#3{\begin{assumption}[#2]\label{#1} #3 \end{assumption} }
\def\ftheorem#1#2#3{\begin{theorem}[#2]\label{#1} #3 \end{theorem} }

\def\fdefinition#1#2#3{\begin{definition}[#2]\label{#1} #3 \end{definition} }

\def\aidx#1{\protect\index{#1}} 
\def\idx#1{\index{#1}#1} 
\def\indxs#1#2{\index{#1!#2}\index{#2!#1}} 
\def\paragraph#1{\vspace{1ex}\noindent{\bf #1.}}
\def\toinfty#1{\stackrel{#1\to\infty}{\longrightarrow}}
\def\nq{\hspace{-1em}}

\def\odt{{\textstyle{1\over 2}}}

\def\odm{{\textstyle{1\over m}}}
\def\odn{{\textstyle{1\over n}}}
\def\eps{\varepsilon}
\def\epstr{\epsilon}                    
\def\pb#1{\def\pb{\underline}\underline{#1}} 
\def\hh#1{{\dot{#1}}}                     
\def\best{*}                              
\def\l{{\ell}}                               
\def\approxleq{\mbox{\raisebox{-0.8ex}{$\stackrel{\displaystyle<}\sim$}}} 
\def\equa{\stackrel+=}                 
\def\leqa{\stackrel+\leq}              
\def\eqm{\stackrel\times=}             
\def\leqm{\stackrel\times\leq}

\def\AI{{\text{AI}}}
\def\SP{{\text{SP}}}
\def\CF{{\text{CF}}}
\def\SG{{\text{SG}}}
\def\FM{{\text{FM}}}

\def\M{{\cal M}}
\def\X{{\cal X}}
\def\Y{{\cal Y}}
\def\O{{\cal O}}

\def\R{{\cal R}}
\def\E{{\bf E}}

\def\B{{I\!\!B}}                       
\def\SetN{I\!\!N}  \def\SetR{I\!\!R} 
\def\qmbox#1{{\quad\mbox{#1}\quad}}
\ifagibook\def\Chapter{Section }\else\def\Chapter{Chapter }\fi
\def\mrcp{the chain rule}
\def\text#1{\mbox{\scriptsize{#1}}}    

\begin{document}

\begin{titlepage}

\title{\vspace{-1cm}
{\small Technical Report IDSIA-01-03 \hfill
  \ifpublished In {\it Artificial General Intelligence}, 2007
  \else 17 January 2003 (minor rev.18Oct'04)
  \fi \\[4mm]}
  \Large\bf\hrule height1pt \vskip 3mm
\mbox{{\LARGE U}NIVERSAL {\LARGE A}LGORITHMIC {\LARGE I}NTELLIGENCE} \\[1mm]
\bf A mathematical top$\to$down approach
\vskip 3mm \hrule height1pt}

\author{Marcus Hutter\\[3ex]
\small IDSIA, Galleria 2, CH-6928 Manno-Lugano, Switzerland%
\\
\small marcus@idsia.ch \hspace{9ex} http://www.idsia.ch/$^{_{_\sim}}\!$marcus
}
\ifpublished\date{17 January 2003}\else\date{}\fi

\maketitle              

\begin{keywords}
Artificial intelligence;
algorithmic probability;
sequential decision theory;
rational agents;
value function;
Solomonoff induction;
Kolmogorov complexity;
reinforcement learning;
universal sequence prediction;
strategic games;
function minimization;
supervised learning.
\end{keywords}

\begin{abstract}
Sequential decision theory formally solves the problem of rational
agents in uncertain worlds if the true environmental prior
probability distribution is known. Solomonoff's theory of
universal induction formally solves the problem of sequence
prediction for unknown prior distribution. We combine both ideas
and get a parameter-free theory of universal Artificial
Intelligence. We give strong arguments that the resulting AIXI
model is the most intelligent unbiased agent possible. We outline
how the AIXI model can formally solve a number of problem classes,
including sequence prediction, strategic games, function
minimization, reinforcement and supervised learning. The major
drawback of the AIXI model is that it is uncomputable. To overcome
this problem, we construct a modified algorithm AIXI$tl$ that is
still effectively more intelligent than any other time $t$ and
length $l$ bounded agent. The computation time of AIXI$tl$ is of
the order $t\cdot 2^l$. The discussion includes formal definitions
of intelligence order relations, the horizon problem and relations
of the AIXI theory to other AI approaches.
\end{abstract}

\end{titlepage}

{\parskip=0ex\tableofcontents}

\section{Introduction}\label{secInt}

This \ifagibook chapter \else article \fi gives an introduction to
a mathematical theory for intelligence. We present the AIXI model,
a parameter-free optimal reinforcement learning agent embedded in
an arbitrary unknown environment.

The science of Artificial Intelligence (AI) may be defined as the
construction of intelligent systems and their analysis. A natural
definition of a {\it system} is anything that has an input and an
output stream. Intelligence is more complicated. It can have many
faces like creativity, solving problems, pattern recognition,
classification, learning, induction, deduction, building
analogies, optimization, surviving in an environment, language
processing, knowledge and many more. A formal definition
incorporating every aspect of intelligence, however, seems
difficult. Most, if not all known facets of intelligence can be
formulated as goal-driven or, more precisely, as maximizing some
utility function. It is, therefore, sufficient to study
goal-driven AI; e.g.\ the (biological) goal of animals and humans
is to survive and spread. The goal of AI systems should be to be
useful to humans. The problem is that, except for special cases,
we know neither the utility function nor the environment in which
the agent will operate in advance. The mathematical theory,
coined AIXI, is supposed to solve these problems.

Assume the availability of unlimited computational resources. The
first important observation is that this does not make the AI
problem trivial. Playing chess optimally or solving NP-complete
problems become trivial, but driving a car or surviving in nature
don't. This is because it is a challenge itself to
well-define the latter problems, not to mention presenting an
algorithm. In other words: The AI problem has not yet been
well defined. One may view AIXI as a suggestion for such a
mathematical definition of AI.

AIXI is a universal theory of sequential decision making akin to
Solomonoff's celebrated universal theory of induction. Solomonoff
derived an optimal way of predicting future data, given previous
perceptions, provided the data is sampled from a computable
probability distribution. AIXI extends this approach to an optimal
decision making agent embedded in an unknown environment. The {\em
main idea} is to replace the unknown environmental distribution
$\mu$ in the Bellman equations by a suitably generalized
universal Solomonoff distribution $\xi$. The state space is the
space of complete histories. AIXI is a universal theory without
adjustable parameters, making no assumptions about the environment
except that it is sampled from a computable distribution. From an
algorithmic complexity perspective, the AIXI model generalizes
optimal passive universal induction to the case of active agents.
From a decision-theoretic perspective, AIXI is a suggestion of a
new (implicit) ``learning'' algorithm, which may overcome all
(except computational) problems of previous reinforcement learning
algorithms.

There are strong arguments that AIXI is the most intelligent
unbiased agent possible. We outline for a number of problem
classes, including sequence prediction, strategic games, function
minimization, reinforcement and supervised learning, how the AIXI
model can formally solve them. The major drawback of the AIXI
model is that it is incomputable. To overcome this problem, we
construct a modified algorithm AIXI$tl$ that is still
effectively more intelligent than any other time $t$ and length $l$
bounded agent. The computation time of AIXI$tl$ is of the order $t
\cdot 2^l$. Other discussed topics are a formal definition of
an intelligence order relation, the horizon problem and relations of
the AIXI theory to other AI approaches.

The article is meant to be a gentle introduction to and discussion
of the AIXI model. For a mathematically rigorous treatment, many
subtleties, and proofs see the references to the author's works in
the annotated bibliography section at the end of this \ifagibook
chapter\else article\fi, and in particular the book
\cite{Hutter:04uaibook}. This section also provides references to
introductory textbooks and original publications on algorithmic
information theory and sequential decision theory.

{\em \Chapter \ref{chAImu}} presents the theory of sequential
decisions in a very general form (called AI$\mu$ model) in which
actions and perceptions may depend on arbitrary past events. We
clarify the connection to the Bellman equations and discuss minor
parameters including (the size of) the I/O spaces and the lifetime
of the agent and their universal choice which we have in mind.
Optimality of AI$\mu$ is obvious by construction.

{\em \Chapter \ref{chSU}:} How and in which sense induction is
possible at all has been subject to long philosophical
controversies. Highlights are Epicurus' principle of multiple
explanations, Occam's razor, and probability theory. Solomonoff
elegantly unified all these aspects into one formal theory of
inductive inference based on a universal probability distribution
$\xi$, which is closely related to Kolmogorov complexity $K(x)$,
the length of the shortest program computing $x$. Rapid
convergence of $\xi$ to the unknown true environmental
distribution $\mu$ and tight loss bounds for arbitrary bounded
loss functions and finite alphabet can be shown. Pareto optimality
of $\xi$ in the sense that there is no other predictor that
performs better or equal in all environments and strictly better
in at least one can also be shown. In view of these results it is
fair to say that the problem of sequence prediction possesses a
universally optimal solution.

{\em \Chapter \ref{chAIxi}:} In the active case, reinforcement
learning algorithms are usually used if $\mu$ is unknown. They can
succeed if the state space is either small or has effectively been
made small by generalization techniques. The algorithms work only
in restricted (e.g.\ Markovian) domains, have problems with
optimally trading off exploration versus exploitation, have
nonoptimal learning rate, are prone to diverge, or are otherwise
ad hoc. The formal solution proposed here is to generalize
Solomonoff's universal prior $\xi$ to include action conditions
and replace $\mu$ by $\xi$ in the AI$\mu$ model, resulting in the
AI$\xi\equiv$AIXI model, which we claim to be universally optimal.
We investigate what we can expect from a universally optimal agent
and clarify the meanings of {\em universal}, {\em optimal}, etc.
Other discussed topics are formal definitions of an intelligence
order relation, the horizon problem, and Pareto optimality of
AIXI.

{\em \Chapter \ref{chApply}:} We show how a number of AI problem
classes fit into the general AIXI model. They include sequence
prediction, strategic games, function minimization, and supervised
learning. We first formulate each problem class in its natural way
(for known $\mu$) and then construct a formulation within the
AI$\mu$ model and show their equivalence. We then consider the
consequences of replacing $\mu$ by $\xi$. The main goal is to
understand in which sense the problems are solved by AIXI.

{\em \Chapter \ref{chTime}:} The major drawback of AIXI is that it
is incomputable, or more precisely, only asymptotically
computable, which makes an implementation impossible. To overcome
this problem, we construct a modified model AIXI$tl$, which is
still superior to any other time $t$ and length $l$ bounded
algorithm. The computation time of AIXI$tl$ is of the order
$t\cdot 2^l$. The solution requires an implementation of first-order logic, the definition of a universal Turing machine within
it and a proof theory system.

{\em \Chapter \ref{chDisc}:} Finally we discuss and remark on some
otherwise unmentioned topics of general interest. We
remark on various topics, including concurrent actions and
perceptions, the choice of the I/O spaces, treatment of encrypted
information, and peculiarities of mortal embodies agents. We
continue with an outlook on further research, including
optimality, down-scaling, implementation, approximation, elegance,
extra knowledge, and training of/for AIXI($tl$). We also include
some (personal) remarks on non-computable physics, the number of
wisdom $\Omega$, and consciousness.

\ifagibook An annotated bibliography concludes this chapter.
\else An annotated bibliography and other references conclude this work.
\fi

\section{Agents in Known Probabilistic Environments}\label{chAImu}

The general framework for AI might be viewed as the design and
study of intelligent agents \cite{Russell:03}. An agent is a
cybernetic system with some internal state, which acts with output
$y_k$ on some environment in cycle $k$, perceives some input $x_k$
from the environment and updates its internal state. Then the next
cycle follows. We split the input $x_k$ into a regular part $o_k$
and a reward $r_k$, often called reinforcement feedback. From time
to time the environment provides nonzero reward to the agent. The
task of the agent is to maximize its utility, defined as the sum
of future rewards. A probabilistic environment can be described by
the conditional probability $\mu$ for the inputs $x_1...x_n$ to
the agent under the condition that the agent outputs $y_1...y_n$.
Most, if not all environments are of this type. We give formal
expressions for the outputs of the agent, which maximize the total
$\mu$-expected reward sum, called value. This model is called the
AI$\mu$ model. As every AI problem can be brought into this form,
the problem of maximizing utility is hence being formally solved,
if $\mu$ is known.
Furthermore, we study some special aspects of the AI$\mu$ model.
We introduce factorizable probability distributions describing
environments with independent episodes. They occur in several
problem classes studied in Section~\ref{chApply} and are a special
case of more general separable probability distributions defined
in Section~\ref{secAIsep}. We also clarify the connection to
the Bellman equations of sequential decision theory and discuss
similarities and differences. We discuss minor parameters of our
model, including (the size of) the input and output spaces $\X$
and $\Y$ and the lifetime of the agent, and their universal
choice, which we have in mind.
There is nothing remarkable in this section; it is the essence of
sequential decision theory
\cite{VonNeumann:44,Bellman:57,Bertsekas:96,Sutton:98}, presented
in a new form. Notation and formulas needed in later sections are
simply developed. There are two major remaining problems: the
problem of the unknown true probability distribution $\mu$, which
is solved in Section~\ref{chAIxi}, and computational aspects,
which are addressed in Section~\ref{chTime}.

\subsection{The Cybernetic Agent Model}
A good way to start thinking about intelligent systems is to
consider more generally cybernetic systems,\index{cybernetic
systems} in AI usually called agents.\index{agents} This
avoids having to struggle with the meaning of \idx{intelligence}
from the very beginning. A cybernetic system is a control circuit
with input\index{input} $y$ and output\index{output} $x$
and an internal state.\index{state!internal} From an external
input and the internal state the agent calculates
deterministically or stochastically an output. This output
(\idx{action}) modifies the environment and leads to a new input
(\idx{perception}). This continues ad infinitum or for a finite
number of cycles.

\fdefinition{defAgent}{The Agent Model}{
An agent is a system that interacts with an environment in cycles
$k=1,2,3,...$. In cycle $k$ the action (output) $y_k \in \Y$ of
the agent is determined by a policy $p$ that depends on the
I/O-history $y_1x_1...y_{k-1}x_{k-1}$. The environment reacts to
this action and leads to a new perception (input) $x_k \in \X$
determined by a deterministic function $q$ or probability
distribution $\mu$, which depends on the history
$y_1x_1...y_{k-1}x_{k-1}y_k$. Then the next cycle $k+1$ starts.
}

\noindent As explained in the last section, we need some
\idx{reward} assignment to the cybernetic system. The input $x$ is
divided into two parts, the standard input $o$ and some reward
input $r$. If input and output are represented by strings, a
\idx{deterministic} cybernetic system can be modeled by a
\idx{Turing machine} $p$, where $p$ is called the \idx{policy} of the agent,
which determines the (re)action to a perception. If the environment is
also computable it might be modeled by a Turing machine $q$ as
well. The interaction of the agent with the environment can be
illustrated as follows:

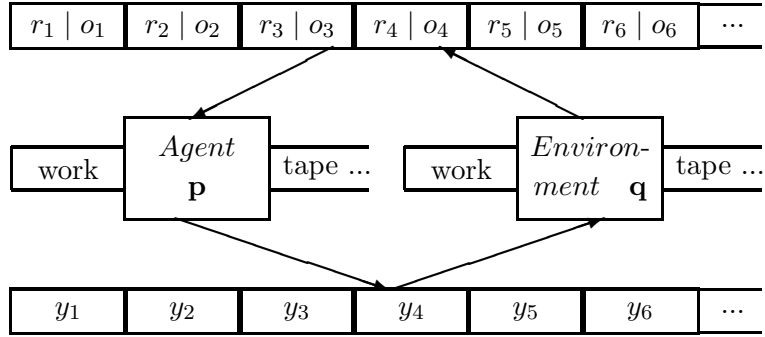
\begin{center}\label{cyberpic}
\small\unitlength=0.95mm\thicklines
\begin{picture}(106,47)
\put( 1,41){\framebox(16,6)[cc]{$r_1 \;|\; o_1$}}
\put(17,41){\framebox(16,6)[cc]{$r_2 \;|\; o_2$}}
\put(33,41){\framebox(16,6)[cc]{$r_3 \;|\; o_3$}}
\put(49,41){\framebox(16,6)[cc]{$r_4 \;|\; o_4$}}
\put(65,41){\framebox(16,6)[cc]{$r_5 \;|\; o_5$}}
\put(81,41){\framebox(16,6)[cc]{$r_6 \;|\; o_6$}}
\put(97,47){\line(1,0){9}}\put(97,41){\line(1,0){9}}\put(102,44){\makebox(0,0)[cc]{...}}
\put( 1,1){\framebox(16,6)[cc]{$y_1$}}
\put(17,1){\framebox(16,6)[cc]{$y_2$}}
\put(33,1){\framebox(16,6)[cc]{$y_3$}}
\put(49,1){\framebox(16,6)[cc]{$y_4$}}
\put(65,1){\framebox(16,6)[cc]{$y_5$}}
\put(81,1){\framebox(16,6)[cc]{$y_6$}}
\put(97,7){\line(1,0){9}}\put(97,1){\line(1,0){9}}\put(102,4){\makebox(0,0)[cc]{...}}
\put(1,21){\line(1,0){16}}
\put(1,27){\line(1,0){16}}
\put(1,21){\line(0,6){6}}
\put(9,24){\makebox(0,0)[cc]{work}}
\put(17,17){\framebox(20,14)[cc]{$\displaystyle{Agent\atop\bf p}$}}
\put(37,27){\line(1,0){14}}
\put(37,21){\line(1,0){14}}
\put(39,24){\makebox(0,0)[lc]{tape ...}}
\put(56,21){\line(1,0){16}}
\put(56,27){\line(1,0){16}}
\put(56,21){\line(0,6){6}}
\put(64,24){\makebox(0,0)[cc]{work}}
\put(72,17){\framebox(20,14)[cc]{$\displaystyle{Environ\mbox{-}\atop ment\quad\bf q}$}}
\put(92,27){\line(1,0){14}}
\put(92,21){\line(1,0){14}}
\put(94,24){\makebox(0,0)[lc]{tape ...}}
%
\put(46,41){\vector(-2,-1){20}}
\put(81,31){\vector(-2,1){20}}
\put(54,7){\vector(3,1){30}}
\put(24,17){\vector(3,-1){30}}
\end{picture}
\end{center}
Both $p$ as well as $q$ have unidirectional input and output tapes
and bidirectional work
tapes.\index{tape!unidirectional}\index{tape!bidirectional} What
entangles the agent with the environment is the fact that the
upper tape serves as input tape for $p$, as well as output tape
for $q$, and that the lower tape serves as output tape for $p$ as
well as input tape for $q$. Further, the reading head must always
be left of the writing head, i.e.\ the symbols must first be
written before they are read. Both $p$ and $q$ have their own mutually
inaccessible work tapes containing their own ``secrets''. The
heads\index{Turing machine!head}\label{defxyr} move in the
following way. In the $k^{th}$ cycle $p$ writes $y_k$, $q$ reads
$y_k$, $q$ writes $x_k \equiv r_k o_k$, $p$ reads $x_k \equiv r_k
o_k$, followed by the $(k + 1)^{th}$ \idx{cycle} and so on. The
whole process starts with the first cycle, all heads on tape start
and work tapes being empty. We call Turing machines
behaving in this way {\it chronological Turing machines}. Before
continuing, some notations on strings are appropriate.

\subsection{Strings}\label{secStrings}\index{strings}\index{sequence}
We denote strings over the \idx{alphabet} $\X$ by $s =
x_1x_2...x_n$, with $x_k \in \X$, where $\X$ is alternatively
interpreted as a nonempty subset of $\SetN$ or itself as a
prefix-free set\index{set!prefix-free} of binary strings. The
length\index{string!length} of $s$ is $\l(s)=\l(x_1) +...+
\l(x_n)$. Analogous definitions hold for $y_k \in \Y$. We call
$x_k$ the $k^{th}$ input word\index{input!word} and $y_k$ the
$k^{th}$ output word\index{output!word} (rather than letter). The
string $s=y_1x_1...y_nx_n$ represents the input/output in
\idx{chronological} order. Due to the \idx{prefix property} of the
$x_k$ and $y_k$, $s$ can be uniquely separated into its words. The
words appearing in strings are always in chronological order. We
further introduce the following abbreviations:
$\epsilon$\index{string!empty} is the empty string,
$x_{n:m}:=x_nx_{n+1}...x_{m-1}x_m$ for $n\leq m$ and $\epsilon$
for $n>m$. $x_{<n}:=x_1... x_{n-1}$. Analogously for $y$. Further,
$y\!x_n :=y_nx_n$, $y\!x_{n:m} := y_nx_n...y_mx_m$, and so on.

\subsection{AI Model for Known Deterministic Environment}
\indxs{environment}{deterministic}%
Let us define for the chronological Turing machine\index{Turing
machine!chronological}\index{chronological!Turing machine} $p$ a
partial function also named $p : \X^* \rightarrow \Y^*$ with
$y_{1:k}=p(x_{<k})$, where $y_{1:k}$ is the output of Turing
machine $p$ on input $x_{<k}$ in cycle $k$, i.e.\ where $p$ has
read up to $x_{k-1}$ but no further.$\!$\footnote{Note that a
possible additional dependence of $p$ on $y_{<k}$ as mentioned in
Definition~\ref{defAgent} can be eliminated by recursive
substitution; see below. Similarly for $q$.} In an analogous way,
we define $q : \Y^* \rightarrow \X^*$ with $x_{1:k}=q(y_{1:k})$.
Conversely, for every partial
recursive\index{chronological!function} chronological function we
can define a corresponding chronological Turing machine. Each
(agent,environment) pair $(p,q)$ produces a unique \idx{I/O
sequence} $\omega^{pq}:=y_1^{pq}x_1^{pq}y_2^{pq}x_2^{pq}...$. When
we look at the definitions of $p$ and $q$ we see a nice symmetry
between the cybernetic system and the environment. Until now, not
much intelligence is in our agent. Now the credit assignment comes
into the game and removes the symmetry somewhat. We split the
input $x_k \in \X := \R \times \O$ into a regular part
\index{input!regular}\index{input!reward}%
\index{reward!maximize}\index{maximize!reward}%
$o_k \in \O$ and a reward $r_k \in \R \subset \SetR$. We define
$x_k \equiv r_k o_k$ and $r_k\equiv r(x_k)$. The goal of the agent
should be to maximize received rewards. This is called
reinforcement learning. The reason for the \idx{asymmetry} is that
eventually we (humans) will be the environment with which the
agent will communicate and {\it we} want to dictate what is good
and what is wrong, not the other way round. This one-way learning,
the agent learns from the environment, and not conversely, neither
prevents the agent from becoming more intelligent than the
environment, nor does it prevent the environment learning from the
agent because the environment can itself interpret the outputs
$y_k$ as a regular and a reward part. The environment is just not
forced to learn, whereas the agent is. In cases where we restrict
the reward to two values $r\in\R=\B:=\{0,1\}$, $r=1$ is
interpreted as a positive
feedback,\index{feedback!positive}\index{feedback!negative} called
{\it good} or {\it correct}, and $r = 0$ a negative feedback,
called {\it bad} or {\it error}. Further, let us restrict for a
while the \label{defHorizon}\idx{lifetime} (number of cycles) $m$
of the agent to a large but finite value. Let
\beqn
  V_{km}^{pq} \;:=\; \sum_{i=k}^m r(x_i^{pq})
\eeqn
be the\index{reward!total}\index{reward!future} future total reward
(called future utility), the agent $p$ receives from the
environment $q$ in the cycles $k$ to $m$. It is now natural to
call the agent $p^\best$ that maximizes $V_{1m}$ (called total
\idx{utility}), the {\it best} one.$\!$\footnote{$\arg\max_p V(p)$ is
the $p$ that maximizes $V(\cdot)$. If there is more than one
maximum we might choose the lexicographically smallest one for
definiteness.}
\beq\label{voptq}
 p^\best:=\arg\max_p V_{1m}^{pq} \quad\Rightarrow\quad
 V_{km}^{p^\best q} \geq V_{km}^{pq} \quad \forall p :
 y_{<k}^{pq}=y_{<k}^{p^\best q}
\eeq
For $k=1$ the condition on $p$ is nil. For $k>1$ it states that
$p$ shall be consistent with $p^*$ in the sense that they have the
same history. If $\X$, $\Y$ and $m$ are finite, the number of
different behaviors of the agent, i.e.\ the search space is
finite. Therefore, because we have assumed that $q$ is known,
$p^\best$ can effectively be determined by pre-analyzing all
behaviors. The main reason for restricting to finite $m$ was not
to ensure computability of $p^\best$ but that the limit
$m\to\infty$ might not exist. The ease with which we defined and
computed the optimal policy $p^\best$ is not remarkable. Just the
(unrealistic) assumption of a completely known deterministic
environment $q$ has trivialized everything.

\subsection{AI Model for Known Prior Probability}\label{secAIfunc}
\indxs{environment}{probabilistic}%
Let us now weaken our assumptions by replacing the deterministic environment $q$
with a probability distribution $\mu(q)$ over chronological
functions. Here $\mu$ might be interpreted in two ways. Either the
environment itself behaves stochastically defined by $\mu$ or the
true environment is deterministic, but we only have subjective
(probabilistic) information of which environment is the true
environment. Combinations of both cases are also possible. We
assume here that $\mu$ is known and describes the true stochastic
behavior of the environment. The case of unknown $\mu$ with the
agent having some beliefs about the environment lies at the heart
of the AI$\xi$ model described in Section~\ref{chAIxi}.

The {\em best} or {\em most intelligent} agent is now the one
that maximizes the {\em expected} utility (called value function)
$
  V_\mu^p\equiv V_{1m}^{p\mu}:=\sum_q\mu(q)V_{1m}^{pq}
$.
This defines the AI$\mu$ model.

\fdefinition{defAImu}{The AI$\mu$ model}{
The AI$\mu$ model is the agent with policy $p^\mu$ that maximizes
the $\mu$-expected total reward $r_1 +...+ r_m$, i.e.\
$p^\best\equiv p^\mu:=\arg\max_p V_\mu^p$. Its value is
$V_\mu^\best:=V_\mu^{p^\mu}$.
}

\noindent We need the concept of a {\em value function} in a
slightly more general form.

\fdefinition{defValue}{The $\mu$/true/generating value function}{
The agent's perception $x$ consists of a regular observation $o \in \O$
and a reward $r \in \R \subset \SetR$. In cycle $k$ the {\em
value} $V_{km}^{p\mu}(y\!x_{<k})$ is defined as the
$\mu$-expectation of the future reward sum $r_k +...+ r_m$ with
actions generated by policy $p$, and fixed history $y\!x_{<k}$. We
say that $V_{km}^{p\mu}(y\!x_{<k})$ is the (future) {\em value} of
policy $p$ in environment $\mu$ given history $y\!x_{<k}$, or
shorter, the $\mu$ or true or generating value of $p$ given
$y\!x_{<k}$. $V_\mu^p:=V_{1m}^{p\mu}$ is the (total) value of $p$.
}

\noindent We now give a more formal definition for
$V_{km}^{p\mu}$. Let us assume we are in cycle $k$ with history
$\hh y\!\hh x_1...\hh y\!\hh x_{k-1}$ and ask for the {\it best}
output $y_k$. Further, let $\hh Q_k := \{q:q(\hh y_{<k})=\hh
x_{<k}\}$ be the set of all environments producing the above
\idx{history}. We say that $q\in\hh Q_k$ is {\em consistent} with
history $\hh y\!\hh x_{<k}$. The expected reward for the next
$m-k+1$ cycles (given the above history) is called the value of
policy $p$ and is given by a conditional probability:
\beq\label{eefunc}
  V_{km}^{p\mu}(\hh y\!\hh x_{<k}) \;:=\;
  { \sum_{q\in \hh Q_k} \mu(q)V_{km}^{pq} \over
    \sum_{q\in \hh Q_k} \mu(q) }.
\eeq
Policy $p$ and environment $\mu$ do not determine history
$\hh y\!\hh x_{<k}$, unlike the
deterministic case, because the history is no longer
deterministically determined by $p$ and $q$, but depends on $p$
and $\mu$ {\it and} on the outcome of a stochastic process.
Every new cycle adds new information ($\hh x_i$) to the
agent. This is indicated by the dots over the symbols.
In cycle $k$ we have to maximize the expected future
rewards, taking into account the information in the history $\hh
y\!\hh x_{<k}$. This information is not already present
in $p$ and $q/\mu$ at the agent's start, unlike in the deterministic
case.

Furthermore, we want to generalize the finite lifetime $m$ to a
dynamic (computable) farsightedness\index{farsightedness!dynamic}
$h_k \equiv m_k - k + 1 \geq 1$, called horizon. For
$m_k = m$ we have our original finite lifetime; for
$h_k = h$ the agent maximizes in every cycle the
next $h$ expected rewards. A discussion of the choices for $m_k$ is
delayed to Section~\ref{secHorizon}.
The next $h_k$ rewards are maximized by
\beqn
  p_k^\best \;:=\; \arg\max_{p\in \hh P_k} V_{km_k}^{p\mu}(\hh y\!\hh
  x_{<k}),
\eeqn
where $\hh P_k := \{p:\exists y_k:p(\hh x_{<k})=\hh y_{<k}y_k\}$
is the set of systems consistent with the current history.
Note that $p_k^\best$ depends on $k$ and is used only in step $k$ to
determine $\hh y_k$ by $ p_k^\best(\hh x_{<k}|\hh y_{<k}) = \hh
y_{<k}\hh y_k$. After writing $\hh y_k$ the environment replies
with $\hh x_k$ with (conditional) probability $\mu(\hh
Q_{k+1})/\mu(\hh Q_k)$. This probabilistic outcome provides new
information to the agent. The cycle $k + 1$ starts with
determining $\hh y_{k+1}$ from $p_{k+1}^\best$ (which can differ
from $p_k^\best$ for dynamic $m_k$) and so on. Note that
$p_k^\best$ implicitly also depends on $\hh y_{<k}$ because $\hh
P_k$ and $\hh Q_k$ do so. But recursively inserting
$p_{k-1}^\best$ and so on, we can define
\beq\label{pbestfunc}
  p^\best(\hh x_{<k}) \;:=\;
  p_k^\best(\hh x_{<k}|p_{k-1}^\best(\hh x_{<k-1}|...p_1^\best))
\eeq
It is a chronological function and computable if $\X$, $\Y$ and
$m_k$ are finite and $\mu$ is computable. For constant $m$ one can
show that the policy (\ref{pbestfunc}) coincides with the AI$\mu$
model (Definition \ref{defAImu}). This also proves
\beq\label{voptmu}
  V_{km}^{\best\mu}(y\!x_{<k}) \geq V_{km}^{p\mu}(y\!x_{<k})
  \quad\forall p \;\mbox{consistent with}\; y\!x_{<k}
\eeq
similarly to (\ref{voptq}). For $k = 1$ this is obvious. We also
call (\ref{pbestfunc}) AI$\mu$ model. For
deterministic\footnote{We call a probability distribution
deterministic if it assumes values 0 and 1 only.} $\mu$ this model
reduces to the deterministic case discussed in the last
subsection.

It is important to maximize the sum of future rewards and not, for
instance, to be \idx{greedy} and only maximize the next reward, as is
done e.g.\ in sequence prediction. For example, let the environment
be a sequence of chess games, and each cycle corresponds to one
move. Only at the end of each game is a positive reward $r = 1$
given to the agent if it won the game (and made no illegal move).
For the agent, maximizing all future rewards means trying to win
as many games in as short as possible time (and avoiding illegal
moves). The same performance is reached if we choose $h_k$ much
larger than the typical game lengths. Maximization of only the
next reward would be a very bad chess playing agent. Even if we
would make our reward $r$ finer, e.g.\ by evaluating the number of
chessmen, the agent would play very bad chess for $h_k = 1$,
indeed.

The AI$\mu$ model still depends on $\mu$ and $m_k$; $m_k$ is
addressed in Section~\ref{secHorizon}. To get our final universal
AI model the idea is to replace $\mu$ by the universal probability
$\xi$, defined later. This is motivated by the fact that $\xi$
converges to $\mu$ in a certain sense for any $\mu$. With $\xi$
instead of $\mu$ our model no longer depends on any parameters, so
it is truly universal. It remains to show that it behaves
intelligently. But let us continue step by step. In the following
we develop an alternative but equivalent formulation of the
AI$\mu$ model. Whereas the \idx{functional form} presented above
is more suitable for theoretical considerations, especially for
the development of a time-bounded version in
Section~\ref{secAIXItl}, the iterative and recursive\index{iterative
formulation}\index{recursive formulation} formulation of the next
subsections will be more appropriate for the explicit calculations
in most of the other sections.

\subsection{Probability Distributions}\label{secPD}
\index{probability!distribution}\index{probability distribution}%
We use Greek letters for probability distributions,
and \idx{underline} their
arguments to indicate that they are probability arguments. Let
$\rho_n(\pb x_1...\pb x_n)$ be the probability that an (infinite)
string starts with $x_1...x_n$. We drop the index on $\rho$ if it
is clear from its arguments:
\beq\label{prop}
  \sum_{x_n\in \X}\rho(\pb x_{1:n}) \equiv
  \sum_{x_n}\rho_n(\pb x_{1:n}) =
  \rho_{n-1}(\pb x_{<n}) \equiv
  \rho(\pb x_{<n})
  ,\quad
  \rho(\epsilon) \equiv \rho_0(\epsilon)=1.
\eeq
We also need conditional probabilities\index{probability
distribution!conditional} derived from the \idx{chain rule}. We prefer a
notation that preserves the chronological
order\index{chronological!order} of the words, in contrast to the
standard notation $\rho(\cdot|\cdot)$ that flips it. We extend
the definition of $\rho$ to the conditional case with the
following convention for its arguments: An underlined argument
$\pb x_k$ is a probability variable, and other non-underlined
arguments $x_k$ represent conditions. With this convention, the
conditional probability has the form $\rho(x_{<n}\pb x_n) =
\rho(\pb x_{1:n})/\rho(\pb x_{<n})$. The equation states that the
probability that a string $x_1...x_{n-1}$ is followed by $x_n$ is
equal to the probability of $x_1...x_n*$ divided by the
probability of $x_1...x_{n-1}*$. We use $x*$ as an abbreviation
for `strings starting with $x$'.

The introduced notation is also suitable for defining the
conditional probability $\rho(y_1\pb x_1...y_n\pb x_n)$ that the
environment reacts with $x_1...x_n$ under the condition that the
output of the agent is $y_1...y_n$. The environment is
chronological, i.e.\ input $x_i$ depends on $y\!x_{<i}y_i$ only. In
the probabilistic case this means that $\rho(y\!\pb
x_{<k}y_k) := \sum_{x_k}\rho(y\!\pb x_{1:k})$ is independent of
$y_k$, hence a tailing $y_k$ in the arguments of $\rho$ can be
dropped. Probability distributions with this property will be
called {\it chronological}. The $y$ are always conditions, i.e.\
are never underlined, whereas additional conditioning for the $x$ can
be obtained with \mrcp
\bqa\label{bayes2}
  \rho(y\!x_{<n}y\!\pb x_n) & = &
  \rho(y\!\pb x_{1:n})/\rho(y\!\pb x_{<n}) \quad\mbox{and}
  \\ \nonumber
  \rho(y\!\pb x_{1:n}) & = &
  \rho(y\!\pb x_1)\!\cdot\!\rho(y\!x_1y\!\pb x_2)\!\cdot...\cdot\!
  \rho(y\!x_{<n}y\!\pb x_n).
\eqa
The second equation is the first equation applied $n$ times.

\subsection{Explicit Form of the AI$\mu$ Model}\label{secAImurec}
\index{AI$\mu $ model!recursive \& iterative form}%
Let us define the AI$\mu$ model $p^\best$ in a different way:
Let $\mu(y\!x_{<k}y\!\pb x_k)$ be the true probability of input
$x_k$ in cycle $k$, given the history $y\!x_{<k}y_k$; $\mu(y\!\pb
x_{1:k})$ is the true chronological prior probability that the
environment reacts with $x_{1:k}$ if provided with actions
$y_{1:k}$ from the agent. We assume the cybernetic model depicted
on page \pageref{cyberpic} to be valid. Next we define the value
$V_{k+1,m}^{\best\mu}(y\!x_{1:k})$ to be the $\mu$-expected reward
sum $r_{k+1} +...+ r_m$ in cycles $k + 1$ to $m$ with outputs
$y_i$ generated by agent $p^\best$ that maximizes the expected
reward sum, and responses $x_i$ from the environment, drawn
according to $\mu$. Adding $r(x_k) \equiv r_k$ we get the reward including
cycle $k$. The probability of $x_k$, given $y\!x_{<k}y_k$, is
given by the conditional probability $\mu(y\!x_{<k}y\!\pb x_k)$. So
the expected reward sum in cycles $k$ to $m$ given $y\!x_{<k}y_k$
is
\beq\label{ebesty}
  V_{km}^{\best\mu}(y\!x_{<k}y_k) \;:=\;
  \sum_{x_k}[r(x_k)+V_{k+1,m}^{\best\mu}(y\!x_{1:k})] \!\cdot\!
  \mu(y\!x_{<k}y\!\pb x_k)
\eeq
Now we ask how $p^\best$ chooses $y_k$: It should choose $y_k$ as
to maximize the future rewards. So the expected reward in cycles
$k$ to $m$ given $y\!x_{<k}$ and $y_k$ chosen by $p^\best$ is $
V_{km}^{\best\mu}(y\!x_{<k})
 := \max_{y_k}V_{km}^{\best\mu}(y\!x_{<k}y_k)$ (see Figure~\ref{figexmax}).

\begin{figure}[tbh]
\ffigurex{figexmax}{Expectimax Tree/Algorithm for $\O=\Y=\B$}{}{%
\small
\unitlength=1mm
\begin{center}
\begin{picture}(145,60)(10,10)
\thicklines
\put(50,60){\circle*{1.5}}
\put(50,60){\line(-1,-1){20}}
\put(39,47){\makebox(0,0)[lc]{$y_k\!=0$}}
\put(50,60){\line(1,-1){20}}
\put(62,47){\makebox(0,0)[rc]{$y_k\!=1$}}
\put(50,55){\makebox(0,0)[ct]{$\underbrace{\;max\;}$}}
\put(55,59){\makebox(0,0)[lc]{$\displaystyle V_{km}^{\best\mu}(y\!x_{<k})=\max_{y_k}V_{km}^{\best\mu}(y\!x_{<k}y_k)$}}
\put(65,50){\makebox(0,0)[lc]{action $y_k$ with max value.}}
\put(30,40){\circle*{1}}
\put(30,40){\line(-1,-2){10}}
\put(24,27){\makebox(0,0)[lc]{$o_k\!=0$}}
\put(23,23){\makebox(0,0)[lc]{$r_k\!=...$}}
\put(30,40){\line(1,-2){10}}
\put(37.5,27){\makebox(0,0)[lc]{$o_k\!=1$}}
\put(39,23){\makebox(0,0)[lc]{$r_k\!=...$}}
\put(30,35){\makebox(0,0)[ct]{$\underbrace{\bf E}$}}
\put(70,40){\circle*{1}}
\put(70,40){\line(-1,-2){10}}
\put(62.5,27){\makebox(0,0)[rc]{$o_k\!=0$}}
\put(61,23){\makebox(0,0)[rc]{$r_k\!=...$}}
\put(70,40){\line(1,-2){10}}
\put(76,27){\makebox(0,0)[rc]{$o_k\!=1$}}
\put(77.5,23){\makebox(0,0)[rc]{$r_k\!=...$}}
\put(70,35){\makebox(0,0)[ct]{$\underbrace{\bf E}$}}
\put(75,39){\makebox(0,0)[lc]{$\displaystyle V_{km}^{\best\mu}(y\!x_{<k}y_k)=\sum_{x_k}[r_k+V_{k+1,m}^{\best\mu}(y\!x_{1:k})]\mu(y\!x_{<k}y\!\pb x_k)$}}
\put(80,30){\makebox(0,0)[lc]{$\mu$-expected reward $r_k$ and observation $o_k$.}}
\thinlines
\put(20,20){\circle*{0.8}}
\put(20,20){\line(-1,-2){5}}
\put(20,20){\line(1,-2){5}}
\put(20,14){\makebox(0,0)[ct]{${{\scriptscriptstyle m\!a\!x}\atop \smile}$}}
\put(30,15){\makebox(0,0)[cc]{$\scriptstyle y_{k+1}$}}
\put(40,20){\circle*{0.8}}
\put(40,20){\line(-1,-2){5}}
\put(40,20){\line(1,-2){5}}
\put(40,14){\makebox(0,0)[ct]{${{\scriptscriptstyle m\!a\!x}\atop \smile}$}}
\put(50,15){\makebox(0,0)[cc]{$\scriptstyle y_{k+1}$}}
\put(60,20){\circle*{0.8}}
\put(60,20){\line(-1,-2){5}}
\put(60,20){\line(1,-2){5}}
\put(60,14){\makebox(0,0)[ct]{${{\scriptscriptstyle m\!a\!x}\atop \smile}$}}
\put(70,15){\makebox(0,0)[cc]{$\scriptstyle y_{k+1}$}}
\put(80,20){\circle*{0.8}}
\put(80,20){\line(-1,-2){5}}
\put(80,20){\line(1,-2){5}}
\put(80,14){\makebox(0,0)[ct]{${{\scriptscriptstyle m\!a\!x}\atop \smile}$}}
\put(85,19){\makebox(0,0)[lc]{\scriptsize $\displaystyle V_{k+1,m}^{\best\mu}(y\!x_{1:k})=\max_{y_{k+1}}V_{k+1,m}^{\best\mu}(y\!x_{1:k}y_{k+1})$}}
\put(15,8){\makebox(0,0)[cc]{$\cdots$}}
\put(25,8){\makebox(0,0)[cc]{$\cdots$}}
\put(35,8){\makebox(0,0)[cc]{$\cdots$}}
\put(45,8){\makebox(0,0)[cc]{$\cdots$}}
\put(55,8){\makebox(0,0)[cc]{$\cdots$}}
\put(65,8){\makebox(0,0)[cc]{$\cdots$}}
\put(75,8){\makebox(0,0)[cc]{$\cdots$}}
\put(85,8){\makebox(0,0)[cc]{$\cdots$}}
\end{picture}
\end{center}}
\end{figure}

Together with the induction start
\beq\label{ee0}
  V_{m+1,m}^{\best\mu}(y\!x_{1:m}) \;:=\; 0
\eeq
$V_{km}^{\best\mu}$ is completely defined.
We might summarize one cycle into the formula
\beq\label{defAImuVi}
  V_{km}^{\best\mu}(y\!x_{<k}) \;=\;
  \max_{y_k}\sum_{x_k}
  [r(x_k)+V_{k+1,m}^{\best\mu}(y\!x_{1:k})] \!\cdot\!
  \mu(y\!x_{<k}y\!\pb x_k)
\eeq
We introduce a dynamic (computable)
farsightedness\index{farsightedness!dynamic} $h_k \equiv m_k - k +
1 \geq 1$, called horizon\index{horizon}. For $m_k=m$, where $m$
is the lifetime of the agent, we achieve optimal behavior, for
limited farsightedness $h_k = h$ ($m = m_k =h + k - 1$), the agent
maximizes in every cycle the next $h$ expected rewards. A
discussion of the choices for $m_k$ is delayed to
Section~\ref{secHorizon}. If $m_k$ is our horizon function of
$p^\best$ and $\hh y\!\hh x_{<k}$ is the actual history in cycle
$k$, the output $\hh y_k$ of the agent is explicitly given by
\beq\label{pbestrec}
  \hh y_k \;=\; \arg\max_{y_k}V_{km_k}^{\best\mu}
  (\hh y\!\hh x_{<k}y_k)
\eeq
which in turn defines the policy $p^\best$. Then the environment responds $\hh x_k$ with
probability $\mu(\hh y\!\hh x_{<k}\hh y\!\pb{\hh
x}_k)$. Then cycle $k + 1$ starts. We might
unfold the recursion (\ref{defAImuVi}) further and give $\hh y_k$
nonrecursively as
\beq\label{ydotrec}
  \hh y_k \;\equiv\; \hh y_k^\mu \;:=\;
  \arg\max_{y_k}\sum_{x_k}\max_{y_{k+1}}\sum_{x_{k+1}}\;...\;
  \max_{y_{m_k}}\sum_{x_{m_k}}
  (r(x_k)\!+...+\!r(x_{m_k})) \!\cdot\!
  \mu(\hh y\!\hh x_{<k}y\!\pb x_{k:m_k})
\eeq
This has a direct interpretation: The probability of inputs
$x_{k:m_k}$ in cycle $k$ when the agent outputs $y_{k:m_k}$ with
actual history $\hh y\!\hh x_{<k}$ is $\mu(\hh y\!\hh x_{<k}y\!\pb
x_{k:m_k})$. The future reward in this case is $r(x_k) +...+
r(x_{m_k})$. The best expected reward is obtained by averaging
over the $x_i$ ($\sum_{x_i}$) and maximizing over the $y_i$. This
has to be done in chronological order to correctly incorporate the
dependencies of $x_i$ and $y_i$ on the history. This is essentially
the expectimax
algorithm/tree\index{expectimax!algorithm}\index{expectimax!tree}
\cite{Michie:66,Russell:03}. The AI$\mu$ model is {\it optimal} in
the sense that no other policy leads to higher expected reward.
The value for a general policy $p$ can be written in the form
\beq\label{vdefpmu}
  V_{km}^{p\mu}(y\!x_{<k}) \;:=\;
  \sum_{x_{1:m}}(r_k\!+...+\!r_m)\mu(y\!x_{<k}y\!\pb
  x_{k:m})_{|y_{1:m}=p(x_{<m})}
\eeq
\index{AI$\mu $ model!equivalence}%
As is clear from their interpretations, the iterative
environmental probability $\mu$ relates to the functional form in
the following way:
\beq\label{mufr}
  \mu(y\!\pb x_{1:k}) \;=\;
  \nq\sum_{q:q(y_{1:k})=x_{1:k}}\nq \mu(q)
\eeq
With this identification one can show \cite{Hutter:00kcunai,Hutter:04uaibook} the
following:

\ftheorem{thEqAI}{Equivalence of functional and explicit AI model}{
The actions of the functional AI model (\ref{pbestfunc}) coincide
with the actions of the explicit (recursive/iterative) AI model
(\ref{defAImuVi})--(\ref{ydotrec}) with environments identified by
(\ref{mufr}).
}

\label{secAIspec}
\index{AI$\mu $ model!special aspects}

\subsection{Factorizable Environments}\label{secFacmu}
\index{factorizable!environment}%
Up to now we have made no restrictions on the form of the prior
probability $\mu$ apart from being a chronological probability
distribution. On the other hand, we will see that, in order to
prove rigorous reward bounds, the prior probability must satisfy
some separability condition to be defined later. Here we introduce
a very strong form of separability, when $\mu$ factorizes into
products.

Assume that the cycles are grouped into
independent episodes $r = 1,2,3,...$, where each episode $r$
consists of the cycles $k = n_r + 1,...,n_{r+1}$ for some
$0=n_0<n_1<...<n_s=n$:
\beq\label{facmu}
  \mu(y\!\pb x_{1:n}) \;=\;
  \prod_{r=0}^{s-1} \mu_r(y\!\pb x_{n_r+1:n_{r+1}})
\eeq
(In the simplest case, when all episodes have the
same length $l$ then $n_r=r \cdot l$). Then $\hh y_k$ depends on
$\mu_r$ and $x$ and $y$ of episode $r$ only, with $r$ such
that $n_r < k \leq n_{r+1}$. One can show that

\beq\label{facydot}
  \hh y_k \;=\; \arg\max_{y_k}V_{km_k}^{\best\mu}
  (\hh y\!\hh x_{<k}y_k) \;=\;
  \arg\max_{y_k}V_{kt}^{\best\mu}(\hh y\!\hh x_{<k}y_k)
\eeq
with $t := \min\{m_k,n_{r+1}\}$. The different episodes
are\index{independent!episodes} completely independent in the
sense that the inputs $x_k$ of different episodes are
statistically independent and depend only on the outputs $y_k$ of
the same \idx{episode}. The outputs $y_k$ depend on the $x$ and
$y$ of the corresponding episode $r$ only, and are independent of
the actual I/O of the other episodes.

Note that $\hh y_k$ is also independent of the choice of $m_k$,
as long as $m_k$ is sufficiently large. If all episodes have a
length of at most $l$, i.e.\ $n_{r+1} - n_r \leq l$ and if we
choose the horizon $h_k$ to be at least $l$, then $m_k \geq k + l
- 1 \geq n_r + l \geq n_{r+1}$ and hence $t=n_{r+1}$ independent
of $m_k$. This means that for factorizable $\mu$ there is no
problem in taking the limit $m_k \to \infty$. Maybe this limit can
also be performed in the more general case of a sufficiently
separable $\mu$. The (problem of the) choice of $m_k$ will be
discussed in more detail later.

Although factorizable $\mu$ are too restrictive to cover all AI
problems, they often occur in practice in the form of repeated
problem solving, and hence, are worthy of study. For example, if
the agent has to play games like chess repeatedly, or has to
minimize different functions, the different games/functions might
be completely independent, i.e.\ the environmental probability
factorizes, where each factor corresponds to a game/function
minimization. For details, see the appropriate sections on
strategic games and function minimization.

Further, for factorizable $\mu$ it is probably easier to derive
suitable reward bounds for the universal AI$\xi$ model defined in
the next section, than for the separable cases that will be
introduced later. This could be a first step toward a definition
and proof for the general case of separable problems. One goal of
this paragraph was to show that the notion of a factorizable
$\mu$ could be the first step toward a definition and analysis of
the general case of separable $\mu$.

\subsection{Constants and Limits}\label{secCL}\index{constants}\index{limits}
We have in mind a universal agent with complex interactions that
is at least as intelligent and complex as a \idx{human} being. One
might think of an agent whose input $y_k$ comes from a digital
\idx{video camera}, and the output $x_k$ is some \idx{image} to a
\idx{monitor},$\!$\footnote{Humans can only simulate a screen as
output device\index{output!device}\index{input!device} by drawing pictures.}
only for the rewards we might restrict to the most primitive
binary ones, i.e.\ $r_k \in \B$. So we think of the following
constant sizes:
\beqn
\begin{array}{ccccccccc}
  1 & \ll & \langle \l(y_kx_k)\rangle & \ll & k & \leq & m & \ll & |\Y\times \X| \\
  1 & \ll & 2^{16} & \ll & 2^{24} & \le & 2^{32} & \ll & 2^{65536}
\end{array}
\eeqn
The first two limits say that the actual number $k$ of
inputs/outputs should be reasonably large compared to the typical
length $\langle\l\rangle$ of the input/output words, which itself
should be rather sizeable. The last limit expresses the fact that
the total \idx{lifetime} $m$ (number of I/O cycles) of the agent
is far too small to allow every possible input to occur, or to try
every possible output, or to make use of identically repeated
inputs or outputs. We do not expect any useful outputs for
$k\approxleq\langle\l\rangle$. More interesting than the lengths
of the inputs is the complexity $K(x_1...x_k)$ of all inputs until
now, to be defined later. The environment is usually not
``\idx{perfect}''. The agent could either interact with an
\idx{imperfect} human or tackle a \idx{nondeterministic world}
(due to\index{quantum physics}\index{physics!quantum} quantum
mechanics or \idx{chaos}).$\!$\footnote{Whether there exist truly
stochastic processes at all is a difficult question. At least the
quantum indeterminacy comes very close to it.} In either case, the
sequence contains some \idx{noise}, leading to
$K(x_1...x_k)\propto \langle\l\rangle \cdot k$. The complexity of
the probability distribution of the input
sequence\index{complexity!input sequence} is something different.
We assume that this \idx{noisy world} operates according to some
simple computable rules. $K(\mu_k)\ll \langle\l\rangle \cdot k$,
i.e.\ the rules of the world can be highly compressed. We may
allow environments in which new aspects appear for $k \to \infty$,
causing a non-bounded $K(\mu_k)$.

In the following we never use these limits, except when explicitly
stated. In some simpler models and examples the size of the
constants will even violate these limits (e.g.\
$\l(x_k)=\l(y_k)=1$), but it is the limits above that the reader
should bear in mind. We are only interested in theorems that do
not degenerate under the above limits. In order to avoid
cumbersome convergence and existence considerations we make the
following assumptions throughout this work:

\fassumption{assFinite}{Finiteness}{
We assume that
\begin{itemize}\parskip=0ex\parsep=0ex\itemsep=0ex
\item the input/perception space $\X$ is finite,
\item the output/action space $\Y$ is finite,
\item the rewards are nonnegative and bounded,
      i.e.\ $r_k \in \R \subseteq [0,r_{max}]$,
\item the horizon $m$ is finite.
\end{itemize}
}
\index{input space!choice}\index{output space!choice}\index{horizon!choice}%
Finite $\X$ and bounded $\R$ (each separately) ensure existence of
$\mu$-expectations but are sometimes needed together. Finite $\Y$
ensures that $\arg\max_{y_k\in\Y}[...]$ exists, i.e.\ that maxima
are attained, while finite $m$ avoids various technical and
philosophical problems (Section~\ref{secHorizon}),
and positive rewards are needed for the time-bounded AIXI$tl$
model (Section~\ref{secAIXItl}). Many theorems can be generalized by
relaxing some or all of the above finiteness assumptions.

\subsection{Sequential Decision Theory}
One can relate (\ref{defAImuVi}) to the Bellman equations
\cite{Bellman:57} of sequential decision theory by identifying
complete histories $y\!x_{<k}$ with states, $\mu(y\!x_{<k}y\!\pb
x_k)$ with the state transition matrix, $V_\mu^\best$ with the
value function, and $y_k$ with the action in cycle $k$
\cite{Bertsekas:96,Russell:03}. Due to the use of complete
histories as state space, the AI$\mu$ model neither assumes
stationarity, nor the Markov property, nor complete accessibility
of the environment. Every state occurs at most once in the
lifetime of the system. For this and other reasons the explicit
formulation (\ref{ydotrec}) is more natural and useful here than
to enforce a pseudo-recursive Bellman equation form.

As we have in mind a universal system with complex interactions,
the action and perception spaces $\Y$ and $\X$ are huge (e.g.\
video images), and every action or perception itself occurs
usually only once in the lifespan $m$ of the agent. As there is no
(obvious) universal similarity relation on the state space, an
effective reduction of its size is impossible, but there is no
principle problem in determining $y_k$ from (\ref{ydotrec}) as
long as $\mu$ is known and computable, and $\X$, $\Y$ and $m$ are
finite.

\index{optimal!policy}
\index{policy!optimal}
\index{policy}
\index{utility!expected}
\index{expected!utility}
\index{complete!history}
\index{history!complete}
\index{state!environmental}
\index{stationarity}
\index{Markov}
\index{accessibility}
\index{universe}
\index{reduction!state space}
\index{value iteration}
\index{policy iteration}
\index{reinforcement learning}
\index{learning!by reinforcement}
\index{generalization techniques}
\index{restricted domains}
\index{exploration}
\index{exploitation}
\index{learning!rate}

Things drastically change if $\mu$ is unknown. Reinforcement
learning algorithms \cite{Kaelbling:96,Sutton:98,Bertsekas:96} are
commonly used in this case to learn the unknown $\mu$ or directly
its value. They succeed if the state space is either small or has
effectively been made small by generalization or function
approximation techniques. In any case, the solutions are either
{\it ad hoc}, work in restricted domains only, have serious
problems with state space exploration versus exploitation, or are
prone to diverge, or have nonoptimal learning rates. There is no
universal and optimal solution to this problem so far. The central
theme of this article is to present a new model and argue that it
formally solves all these problems in an optimal way. The true
probability distribution $\mu$ will not be learned directly, but
will be replaced by some generalized universal prior $\xi$, which
converges to $\mu$.

\section{Universal Sequence Prediction}\label{chSP}\label{chSU}

This section deals with the question of how to make predictions
in unknown environments.
Following a brief description of important philosophical attitudes
regarding inductive reasoning and inference, we describe more
accurately what we mean by induction, and motivate why we can
focus on sequence prediction tasks. The most important concept is
Occam's razor (simplicity) principle.
Indeed, one can show that the best way to make predictions is
based on the shortest ($\widehat=$ simplest) description of the
data sequence seen so far. The most general effective descriptions
can be obtained with the help of general recursive functions, or
equivalently by using programs on Turing machines, especially on
the universal Turing machine. The length of the shortest program
describing the data is called the Kolmogorov complexity of the
data.
Probability theory is needed to deal with uncertainty.
The environment may be a stochastic process (e.g.\ gambling houses
or quantum physics) that can be described by ``objective''
probabilities. But also uncertain knowledge about the environment,
which leads to beliefs about it, can be modeled by ``subjective''
probabilities.
The old question left open by subjectivists of how to choose the a
priori probabilities is solved by Solomonoff's universal prior,
which is closely related to Kolmogorov complexity.
Solomonoff's major result is that the universal (subjective) posterior
converges to the true (objective) environment(al probability)
$\mu$. The only assumption on $\mu$ is that $\mu$ (which needs not
be known!) is computable. The problem of the unknown environment
$\mu$ is hence solved for all problems of inductive type, like
sequence prediction and classification.

\subsection{Introduction}

An important and highly nontrivial aspect of intelligence
is inductive inference. Simply speaking, induction is the process
of predicting the future from the past, or more precisely, it is
the process of finding rules in (past) data and using these rules
to guess future data. Weather or stock-market forecasting, or
continuing number series in an IQ test are nontrivial examples.
Making good predictions plays a central role in natural and
artificial intelligence in general, and in machine learning in
particular.
All induction problems can be phrased as sequence prediction
tasks. This is, for instance, obvious for time-series prediction,
but also includes classification tasks. Having observed data $x_t$
at times $t<n$, the task is to predict the $n^{th}$ symbol $x_n$
from sequence $x_1...x_{n-1}$. This {\em\idx{prequential
approach}} \cite{Dawid:84} skips over the intermediate step of
learning a model based on observed data $x_1...x_{n-1}$ and then
using this model to predict $x_n$. The prequential approach avoids
problems of model consistency, how to separate noise from useful
data, and many other issues. The goal is to make ``good''
predictions, where the prediction quality is usually measured by a
loss function, which shall be minimized.
The key concept to well-define and solve induction problems is
{\em Occam's razor} (simplicity) principle, which says that ``{\em
Entities should not be multiplied beyond necessity},'' which may
be interpreted as to keep the simplest theory consistent with the
observations $x_1...x_{n-1}$ and to use this theory to predict
$x_n$.
Before we can present Solomonoff's formal solution, we have to
quantify Occam's razor in terms of Kolmogorov complexity, and
introduce the notion of subjective/objective probabilities.

\subsection{Algorithmic Information Theory}
\index{Kolmogorov complexity}\index{complexity!Kolmogorov}

Intuitively, a string is simple if it can be described in a few
words, like ``the string of one million ones'', and is complex if
there is no such short description, like for a random string whose
shortest description is specifying it bit by bit. We can restrict
the discussion to binary strings, since for other (non-stringy
mathematical) objects we may assume some default coding as binary
strings. Furthermore, we are only interested in effective
descriptions, and hence restrict decoders to be Turing machines.
Let us choose some universal (so-called prefix) {\em Turing
machine $U$} with unidirectional binary input and output tapes and
a bidirectional work tape \cite{Li:97,Hutter:04uaibook}. We can
then define the (conditional) {\em prefix Kolmogorov complexity}
\cite{Chaitin:75,Gacs:74,Kolmogorov:65,Levin:74} of a binary
string $x$ as the length $l$ of the shortest program $p$, for
which $U$ outputs the binary string $x$ (given $y$)

\fdefinition{defKC}{Kolmogorov complexity}{
Let $U$ be a universal prefix Turing machine $U$. The
(conditional) prefix Kolmogorov
complexity\index{Kolmogorov complexity} is defined as the shortest
program $p$, for which $U$ outputs $x$ (given $y$):
\beqn
  K(x) \;:=\; \min_p\{\l(p): U(p)=x\},\quad
  K(x|y) \;:=\; \min_p\{\l(p): U(y,p)=x\}
\eeqn\vspace{-3ex}
}
Simple strings like 000...0 can be generated by short programs,
and hence have low Kolmogorov complexity, but irregular (e.g.\
random) strings are their own shortest description, and hence have
high Kolmogorov complexity. An important property of $K$ is that
it is nearly independent of the choice of $U$. Furthermore, it
shares many properties with Shannon's entropy (information
measure) $S$, but $K$ is superior to $S$ in many respects. To be
brief, $K$ is an excellent universal complexity measure, suitable
for quantifying Occam's razor. There is (only) one severe
disadvantage: $K$ is not finitely computable. The major
algorithmic property of $K$ is that it is (only) co-enumerable,
i.e.\ it is approximable from above.

For general (non-string) objects one can specify some default
coding $\langle\cdot\rangle$ and define $K(\mbox{\it
object}):=K(\langle\mbox{\it object}\rangle)$, especially for
numbers and pairs, e.g.\ we abbreviate $K(x,y):=K(\langle
x,y\rangle)$. The most important information-theoretic properties
of $K$ are listed below, where we abbreviate $f(x)\leq g(x)+O(1)$
by $f(x)\leqa g(x)$. We also later abbreviate $f(x)=O(g(x))$ by
$f(x)\leqm g(x)$.

\ftheorem{thKCprop}{Information properties of Kolmogorov complexity}{\hfill
\begin{itemize}\parskip=0ex\parsep=0ex\itemsep=1ex
\item[$i)$] $K(x) \;\leqa\; \l(x) + 2\log\,\l(x),\qquad
             K(n) \;\leqa\; \log n + 2\log\log n$
\item[$ii)$] $\sum_x 2^{-K(x)}\;\leq\;1,\quad$
             $K(x)\;\geq\;l(x)$ for `most' $x$,
             $\quad K(n)\to\infty$ for $n\to\infty$.
\item[$iii)$] $K(x|y) \;\leqa\; K(x) \;\leqa\; K(x,y)$
\item[$iv)$] $K(x,y) \;\leqa\; K(x)+K(y),\qquad
            K(xy) \;\leqa\; K(x)+K(y)$
\item[$v)$] $K(x|y,K(y))+K(y) \;\equa\; K(x,y)
              \;\equa\; K(y,x) \;\equa\; K(y|x,K(x))+K(x)$
\item[$vi)$] $K(f(x)) \;\leqa\; K(x)+K(f)\;$
     if $\;f:\B^*\to\B^*$ is recursive/computable
\item[$vii)$] $K(x) \;\leqa\; -\log_2P(x)+K(P)$ if $P:\B^*\to[0,1]$
     is recursive and $\sum_x P(x)\leq 1$ 
\end{itemize}
}

\noindent All (in)equalities remain valid if $K$ is (further)
conditioned under some $z$, i.e.\ $K(...)\leadsto K(...|z)$ and
$K(...|y)\leadsto K(...|y,z)$. Those stated are all valid within
an additive constant of size $O(1)$, but there are others which
are only valid to logarithmic accuracy.
$K$ has many properties in common with Shannon entropy as it
should be, since both measure the information content of a
string.
Property $(i)$ gives an upper bound on $K$, and property $(ii)$ is
Kraft's inequality which implies a lower bound on $K$ valid for
`most' $n$, where `most' means that there are only $o(N)$
exceptions for $n\in\{1,..., N\}$. Providing side information $y$
can never increase code length, requiring extra information $y$
can never decrease code length $(iii)$. Coding $x$ and $y$
separately never helps $(iv)$, and transforming $x$ does not
increase its information content $(vi)$. Property $(vi)$ also
shows that if $x$ codes some object $o$, switching from one coding
scheme to another by means of a recursive bijection leaves $K$
unchanged within additive $O(1)$ terms. The first nontrivial
result is the symmetry of information $(v)$, which is the analogue
of the multiplication/chain rule for conditional probabilities.
Property $(vii)$ is at the heart of the MDL principle
\cite{Rissanen:89}, which approximates $K(x)$ by
$-\log_2P(x)+K(P)$. See \cite{Li:97} for proofs.

\subsection{Uncertainty \& Probabilities}

For the {\em\idx{objectivist}} probabilities are real aspects of
the world. The outcome of an observation or an \idx{experiment} is
not deterministic, but involves \idx{physical random processes}.
Kolmogorov's axioms of probability theory formalize the properties
that probabilities should have. In the case of i.i.d.\
experiments the probabilities assigned to events can be
interpreted as limiting frequencies ({\em frequentist} view), but
applications are not limited to this case. Conditionalizing
probabilities and Bayes' rule\index{chain rule} are the major tools in
computing posterior probabilities from prior ones.
For instance, given the initial binary sequence $x_1...x_{n-1}$,
what is the probability of the next bit being $1$? The probability
of observing $x_n$ at time $n$, given past observations
$x_1...x_{n-1}$ can be computed with the multiplication or chain
rule\footnote{Strictly speaking it is just the definition of
conditional probabilities.} if the true generating distribution
$\mu$ of the sequences $x_1x_2x_3...$ is known: $\mu(x_{<n}\pb
x_n)=\mu(\pb x_{1:n})/\mu(\pb x_{<n})$ (see
Sections~\ref{secStrings} and \ref{secPD}). The problem, however,
is that one often does not know the true distribution $\mu$ (e.g.\
in the cases of weather and stock-market forecasting).

The {\em\idx{subjectivist}} uses probabilities to characterize an
agent's \idx{degree of belief} in (or plausibility of) something,
rather than to characterize physical random processes. This is the
most relevant interpretation of probabilities in AI.
It is somewhat surprising that plausibilities can be shown to also
respect Kolmogorov's axioms of probability and \mrcp\ for
conditional probabilities by assuming only a few plausible
qualitative rules they should follow \cite{Cox:46}. Hence, if the
plausibility of $x_{1:n}$ is $\xi(\pb x_{1:n})$, the degree of
belief in $x_n$ given $x_{<n}$ is, again, given by the conditional
probability: $\xi(x_{<n}\pb x_n)=\xi(\pb x_{1:n})/\xi(\pb
x_{<n})$.

The \mrcp\ allows determining posterior
probabilities/plausibilities from prior ones, but leaves open the
question of how to determine the priors themselves. In statistical
physics, the principle of indifference (symmetry principle) and
the maximum entropy principle can often be exploited to determine
prior probabilities, but only Occam's razor is general enough to
assign prior probabilities in {\em every} situation, especially to
cope with complex domains typical for AI.


\subsection{Algorithmic Probability \& Universal Induction}

Occam's razor (appropriately interpreted and in compromise with
Epicurus' principle of indifference) tells us to assign high/low a
priori plausibility to simple/complex strings $x$. Using $K$ as
the complexity measure, any monotone decreasing function of $K$,
e.g.\ $\xi(\pb x)=2^{-K(x)}$ would satisfy this criterion. But $\xi$
also has to satisfy the probability axioms, so we have to be a bit
more careful.
Solomonoff \cite{Solomonoff:64,Solomonoff:78} defined the {\em
universal prior} $\xi(\pb x)$ as the probability that the output
of a universal Turing machine $U$ starts with $x$ when provided
with \idx{fair coin flips} on the input tape. Formally, $\xi$
can be defined as
\beq\label{xidef}
  \xi(\pb x)\;:=\;\sum_{p\;:\;U(p)=x*}\nq 2^{-\l(p)} \;\geq\; 2^{-K(x)}
\eeq
where the sum is over all (so-called minimal) programs $p$ for
which $U$ outputs a string starting with $x$.
The inequality follows by dropping all terms in $\sum_p$ except
for the shortest $p$ computing $x$. Strictly speaking $\xi$ is only
a {\em semimeasure} since it is not normalized to 1, but this is
acceptable/correctable.
We derive the following bound:
\beqn
  \sum_{t=1}^\infty(1\!-\!\xi(x_{<t}\pb x_t))^2 \;\leq\;
  -\odt \sum_{t=1}^\infty\ln \xi(x_{<t}\pb x_t) \;=\;
  -\odt\ln\xi(\pb x_{1:\infty}) \;\leq\;
  \odt\ln 2\!\cdot\!K(x_{1:\infty})
\eeqn
In the first inequality we have used $(1-a)^2\leq-\odt\ln a$ for
$0\leq a\leq 1$. In the equality we exchanged the sum with the
logarithm and eliminated the resulting product by \mrcp\
(\ref{bayes2}). In the last inequality we used (\ref{xidef}). If
$x_{1:\infty}$ is a computable sequence, then $K(x_{1:\infty})$ is
finite, which implies $\xi(x_{<t}\pb x_t)\to 1$
($\sum_{t=1}^\infty(1-a_t)^2<\infty\Rightarrow a_t\to 1$). This
means, that if the environment is a computable sequence
(whichsoever, e.g.\ the digits of $\pi$ or $e$ in binary
representation), after having seen the first few digits, $\xi$
correctly predicts the next digit with high probability, i.e.\ it
recognizes the structure of the sequence.

Assume now that the true sequence is drawn from the distribution
$\mu$, i.e.\ the true (objective) probability of $x_{1:n}$ is
$\mu(\pb x_{1:n})$, but $\mu$ is unknown. How is the posterior
(subjective) belief $\xi(x_{<n}\pb x_n)=\xi(\pb x_n)/\xi(\pb
x_{<n})$ related to the true (objective) posterior probability
$\mu(x_{<n}\pb x_n)$?
Solomonoff's \cite{Solomonoff:78} crucial result is that the
posterior (subjective) beliefs converge to the true (objective)
posterior probabilities, if the latter are computable. More precisely, he showed that
\beq\label{eukdistxi}
  \sum_{t=1}^\infty\sum_{x_{<t}}\mu(\pb x_{<t})
  \Big(\xi(x_{<t}\pb 0)-\mu(x_{<t}\pb 0)\Big)^2 \;\leqa\;
  {\odt}\ln 2\!\cdot\!K(\mu),
\eeq
$K(\mu)$ is finite if $\mu$ is computable, but the infinite sum on
the l.h.s.\ can only be finite if the difference $\xi(x_{<t}\pb
0)-\mu(x_{<t}\pb 0)$ tends to zero for $t\to\infty$ with
$\mu$-probability $1$. This shows that using $\xi$ as an estimate
for $\mu$ may be a reasonable thing to do.

\subsection{Loss Bounds \& Pareto Optimality}\label{secLBPO}\label{secErr}

Most predictions are eventually used as a basis for some decision
or action, which itself leads to some reward or loss. Let
$\ell_{x_t y_t}\in[0,1]\subset\SetR$ be the received loss when
performing prediction/decision/action $y_t\in\Y$ and $x_t\in\X$ is
the $t^{th}$ symbol of the sequence. Let $y_t^\Lambda \in \Y$ be
the prediction of a (causal) prediction scheme $\Lambda$. The true
probability of the next symbol being $x_t$, given $x_{<t}$, is
$\mu(x_{<t}\pb x_t)$. The expected loss when predicting $y_t$ is
$\E[\ell_{x_t y_t}]$. The total $\mu$-expected loss suffered by
the $\Lambda$ scheme in the first $n$ predictions is
\beq\label{rholossi}
  L_n^\Lambda \;:=\; \sum_{t=1}^n
  \E[\ell_{x_t y_t^\Lambda}] \;=\;
  \sum_{t=1}^n\sum_{x_{1:t}\in\X^t} \mu(\pb x_{1:t}) \ell_{x_t y_t^\Lambda}
\eeq
For instance, for the error-loss $\l_{xy}=1$ if $x=y$ and 0 else,
$L_n^\Lambda$ is the expected number of prediction errors, which
we denote by $E_n^\Lambda$. The goal is to minimize the expected
loss. More generally, we define the $\Lambda_\rho$ sequence
prediction scheme (later also called SP$\rho$)
$
  y_t^{\smash{\Lambda_\rho}} :=
  \arg\min_{y_t\in\Y}\sum_{x_t}\rho(x_{<t}\pb x_t)\ell_{x_t y_t}
$
which minimizes the $\rho$-expected loss.
If $\mu$ is known, $\Lambda_\mu$ is obviously the
best prediction scheme in the sense of achieving minimal expected
loss ($L_n^{\smash{\Lambda_\mu}}\leq L_n^\Lambda$ for any $\Lambda$).
One can prove the following loss bound for the universal
$\Lambda_\xi$ predictor \cite{Hutter:01loss,Hutter:01alpha,Hutter:02spupper}
\beq\label{thULoss}
  0 \;\leq\; L_n^{\smash{\Lambda_\xi}}-L_n^{\smash{\Lambda_\mu}} \;\leq\;
  2\ln 2\!\cdot\! K(\mu)+2\sqrt{L_n^{\smash{\Lambda_\mu}}\ln 2\!\cdot\! K(\mu)}
\eeq
Together with $L_n\leq n$ this shows that $\odn
L_n^{\smash{\Lambda_\xi}}-\odn L_n^{\smash{\Lambda_\mu}}=O(n^{-1/2})$, i.e.\
asymptotically $\Lambda_\xi$ achieves the optimal average loss of
$\Lambda_\mu$ with rapid convergence. Moreover
$L_\infty^{\smash{\Lambda_\xi}}$ is finite if $L_\infty^{\smash{\Lambda_\mu}}$ is
finite and $L_n^{\smash{\Lambda_\xi}}/L_n^{\smash{\Lambda_\mu}}\to 1$ if
$L_\infty^{\smash{\Lambda_\mu}}$ is not finite. Bound (\ref{thULoss}) also
implies $L_n^\Lambda\geq L_n^{\smash{\Lambda_\xi}} -
2\sqrt{L_n^{\smash{\Lambda_\xi}}\ln 2\cdot K(\mu)}$, which shows that {\em
no} (causal) predictor $\Lambda$ whatsoever achieves significantly
less (expected) loss than $\Lambda_\xi$. In view of these results
it is fair to say that, ignoring computational issues, the problem
of sequence prediction has been solved in a universal way.

A different kind of optimality is {\em Pareto optimality}. The
universal prior $\xi$ is Pareto optimal in the sense that there is
no other predictor that leads to equal or smaller loss in {\em
all} environments. Any improvement achieved by some predictor
$\Lambda$ over $\Lambda_\xi$ in some environments is
balanced by a deterioration in other environments
\cite{Hutter:03optisp}.

\section{The Universal Algorithmic Agent AIXI}\label{chAIxi}

Active systems, like game playing (SG) and optimization (FM),
cannot be reduced to induction systems. The {\it main idea of this
work} is to generalize universal induction to the general agent
model described in Section~\ref{chAImu}. For this, we generalize
$\xi$ to include actions as conditions and replace $\mu$ by $\xi$ in the
rational agent model, resulting in the AI$\xi$(=AIXI) model. In
this way the problem that the true prior probability $\mu$ is
usually unknown is solved. Convergence of $\xi\to\mu$ can be
shown, indicating that the AI$\xi$ model could behave optimally
in any computable but unknown environment with reinforcement
feedback.

The main focus of this section is to investigate what we can
expect from a universally optimal agent and to clarify the
meanings of {\em universal}, {\em optimal}, etc. Unfortunately
bounds similar to the loss bound (\ref{thULoss}) in the SP case
can hold for {\em no} active agent. This forces us to lower our
expectation about universally optimal agents and to introduce
other (weaker) performance measures.
Finally, we show that AI$\xi$ is Pareto optimal in the sense that
there is no other policy yielding higher or equal value in {\em
all} environments and a strictly higher value in at least one.

\subsection{The Universal AI$\xi$ Model}\label{secAIxi}
\index{AI$\xi $ model}\index{universal!AI$\xi$ model}
\index{model!universal}\index{model!AI$\xi$}

\paragraph{Definition of the AI$\xi$ model}
We have developed enough formalism to suggest our universal
AI$\xi$ model. All we have to do is to suitably generalize the
universal semimeasure $\xi$ from the last section and
replace the true but unknown prior probability $\mu^\AI$ in the
AI$\mu$ model by this generalized $\xi^\AI$. In what sense
this AI$\xi$ model is universal will be discussed subsequently.
\index{universal!generalized prior}
\index{generalized universal prior}

In the functional formulation we define the universal probability
$\xi^\AI$ of an environment $q$ just as $2^{-\l(q)}$
\beqn
  \xi(q) \;:=\; 2^{-\l(q)}
\eeqn
The definition could not be easier\footnote{It is not necessary to
use $2^{-K(q)}$ or something similar as some readers may expect at
this point, because for every program $q$ there exists
a functionally equivalent program $\tilde q$ with
$K(q)\equa\l(\tilde q)$.}!\footnote{Here and later we identify
objects with their coding relative to some fixed Turing machine
$U$. For example, if $q$ is a function $K(q):=K(\langle q\rangle)$
with $\langle q\rangle$ being a binary coding of $q$ such that
$U(\langle q\rangle,y)=q(y)$. Reversely, if $q$ already is a
binary string we define $q(y) :=U(q,y)$.} Collecting the formulas
of Section~\ref{secAIfunc} and replacing $\mu(q)$ by $\xi(q)$ we
get the definition of the AI$\xi$ agent in functional form. Given
the history $\hh y\!\hh x_{<k}$ the policy $p^\xi$ of the
functional AI$\xi$ agent is given by
\beq\label{eefuncxi}
  \hh y_k \;:=\;
  \arg\max_{y_k}\max_{p:p(\hh x_{<k})=\hh y_{<k}y_k}
  \sum_{q:q(\hh y_{<k})=\hh x_{<k}}
  \nq 2^{-\l(q)}\cdot V_{km_k}^{pq}
\eeq
in cycle $k$, where $V_{km_k}^{pq}$ is the total reward of cycles $k$ to $m_k$ when
agent $p$ interacts with environment $q$. We have dropped the
denominator $\sum_q\mu(q)$ from (\ref{eefunc}) as it is
independent of the $p \in \hh P_k$ and a constant multiplicative
factor does not change $\arg\max_{y_k}$.

For the iterative formulation, the universal probability
$\xi$ can be obtained by inserting the functional $\xi(q)$ into
(\ref{mufr})
\beq\label{uniMAI}
  \xi(y\!\pb x_{1:k}) \;=\;
  \nq\sum_{q:q(y_{1:k})=x_{1:k}}\nq 2^{-\l(q)}
\eeq
Replacing $\mu$ by $\xi$ in (\ref{ydotrec}) the
iterative AI$\xi$ agent outputs
\beq\label{ydotxi}
  \hh y_k \;\equiv\; \hh y_k^\xi \;:=\;
  \arg\max_{y_k}\sum_{x_k}\max_{y_{k+1}}\sum_{x_{k+1}}\;...\;
  \max_{y_{m_k}}\sum_{x_{m_k}}
  (r(x_k)\!+...+\!r(x_{m_k})) \!\cdot\!
  \xi(\hh y\!\hh x_{<k}y\!\pb x_{k:m_k})
\eeq
in cycle $k$ given the history $\hh y\!\hh x_{<k}$.

The equivalence of the functional and iterative AI model (Theorem
\ref{thEqAI}) is true for every chronological semimeasure $\rho$,
especially for $\xi$, hence we can talk about {\it the} AI$\xi$
model in this respect. It (slightly) depends on the choice of the
universal Turing machine. $\l(\langle q\rangle)$ is defined only up
to an additive constant. The AI$\xi$ model also depends on the
choice of $\X = \R \times \O$ and $\Y$, but we do not expect any
\idx{bias} when the spaces are chosen sufficiently simple, e.g.\
all strings of length $2^{16}$. Choosing $\SetN$ as the word
space would be ideal, but whether the maxima (suprema) exist in
this case, has to be shown beforehand. The only nontrivial
dependence is on the horizon function $m_k$ which will be
discussed later. So apart from $m_k$ and unimportant details the
AI$\xi$ agent is uniquely defined by (\ref{eefuncxi}) or
(\ref{ydotxi}). It does not depend on any assumption about the
environment apart from being generated by some computable (but
unknown!) probability distribution.

\paragraph{Convergence of $\xi$ to $\mu$}
Similarly to (\ref{eukdistxi}) one can show that the $\mu$-expected
squared difference of $\mu$ and $\xi$ is finite for computable
$\mu$. This, in turn, shows that $\xi(y\!x_{<k}y\!\pb x_k)$
converges rapidly to $\mu(y\!x_{<k}y\!\pb x_k)$ for $k \to \infty$ with
$\mu$-probability 1. The line of reasoning is the same; the $y$
are pure spectators. This will change when we analyze
loss/reward bounds analogous to (\ref{thULoss}).
More generally, one can show \cite{Hutter:04uaibook} that\footnote{Here,
and everywhere else, with $\xi_k\to\mu_k$ we mean $\xi_k-\mu_k\to
0$, and not that $\mu_k$ (and $\xi_k$) itself converge to a
limiting value.}
\beq\label{aixitomu}
  \xi(y\!x_{<k}y\!\pb x_{k:m_k}) \toinfty{k} \mu(y\!x_{<k}y\!\pb x_{k:m_k})
\eeq
This gives hope that the outputs $\hh y_k$ of the AI$\xi$ model
(\ref{ydotxi}) could converge to the outputs $\hh y_k$ from the
AI$\mu$ model (\ref{ydotrec}).

\indxs{solvable}{problem}\indxs{learnable}{task}
\indxs{universal}{optimality}
We want to call an AI model {\it universal}, if it is
$\mu$-independent (unbiased, model-free) and is able
to solve any solvable problem and learn any learnable task.
Further, we call a universal model, {\it universally optimal}, if
there is no program, which can solve or learn significantly faster
(in terms of interaction cycles). Indeed, the AI$\xi$ model is
parameter free, $\xi$ converges to $\mu$ (\ref{aixitomu}), the
AI$\mu$ model is itself optimal, and we expect no other model to
converge faster to AI$\mu$ by analogy to SP (\ref{thULoss}):

\fclaim{clMain}{We expect AIXI to be universally optimal}{\vspace{-1ex}}
%
This is our main claim. In a sense, the intention of the remaining
sections is to define this statement more rigorously and to give
further support.

\paragraph{Intelligence order relation}\label{secIOR}
\indxs{intelligence}{order relation}%
\indxs{consistent}{policy}\indxs{inconsistent}{policy}%
We define the $\xi$-expected reward in cycles $k$ to $m$ of a
policy $p$ similar to (\ref{eefunc}) and (\ref{eefuncxi}).
We extend the definition to programs $p \not\in \hh P_k$ that
are not consistent with the current history.
\beq\label{cxi}
  V_{km}^{p\xi}(\hh y\!\hh x_{<k}) \;:=\;
  {1\over\cal N}
  \sum_{q:q(\hh y_{<k})=\hh x_{<k}}
  \nq 2^{-\l(q)}\cdot V_{km}^{\tilde p q}
\eeq
The normalization $\cal N$ is again only necessary for
interpreting $V_{km}$ as the expected reward but is otherwise
unneeded. For consistent policies $p\in\hh P_k$ we define $\tilde
p := p$. For $p \not\in \hh P_k$, $\tilde p$ is a modification of
$p$ in such a way that its outputs are consistent with the current
history $\hh y\!\hh x_{<k}$, hence $\tilde p \in \hh P_k$, but
unaltered for the current and future cycles $\geq k$. Using this
definition of $V_{km}$ we could take the maximium over all
policies $p$ in (\ref{eefuncxi}), rather than only the consistent
ones.

\fdefinition{defaiorder}{Intelligence order relation}{
We call a policy $p$ {\it more or equally intelligent} than $p'$ and write
\beqn
  p\succeq p' \quad:\Leftrightarrow\quad
  \forall k\forall\hh y\!\hh x_{<k}:
  V_{km_k}^{p\xi}(\hh y\!\hh x_{<k}) \geq
  V_{km_k}^{p'\xi}(\hh y\!\hh x_{<k}).
\eeqn
i.e.\ if $p$ yields in any circumstance higher $\xi$-expected
reward than $p'$.
}

\indxs{most intelligent}{agent}
\indxs{universal}{order relation}
\indxs{intermediate}{intelligence}%
\noindent As the algorithm $p^\best$ behind the AI$\xi$
agent maximizes $V_{km_k}^{p\xi}$ we have $p^\xi\succeq p$ for all
$p$. The AI$\xi$ model is hence the most intelligent agent
w.r.t.\ $\succeq$. Relation $\succeq$ is a universal order relation in the
sense that it is free of any parameters (except $m_k$) or specific
assumptions about the environment. A proof, that $\succeq$ is a
reliable intelligence order (which we believe to be true), would
prove that AI$\xi$ is universally optimal. We could further ask:
How useful is $\succeq$ for ordering policies of practical
interest with intermediate intelligence, or how can $\succeq$ help
to guide toward constructing more intelligent systems with
reasonable computation time? An effective intelligence order
relation $\succeq^c$ will be defined in Section~\ref{secAIXItl},
which is more useful from a practical point of view.

\subsection{On the Optimality of AIXI}\label{secOOAIXI}
\indxs{optimality}{AI$\xi$ model}
\indxs{adaptive}{control}

In this section we outline ways toward an optimality proof of
AIXI. Sources of inspiration are the SP loss bounds proven in
Section~\ref{chSP} and optimality criteria from the adaptive control
literature (mainly) for linear systems \cite{Kumar:86}.
The value bounds for AIXI are expected to be, in a sense, weaker
than the SP loss bounds because the problem class covered by AIXI
is much larger than the class of induction problems. Convergence
of $\xi$ to $\mu$ has already been proven, but is not sufficient
to establish convergence of the behavior of the AIXI model to the
behavior of the AI$\mu$ model. We will focus on three approaches
toward a general optimality proof:

\indxs{universal}{optimality}
\paragraph{What is meant by universal optimality}
The first step is to investigate what we can expect from AIXI,
i.e.\ what is meant by {\it universal optimality}. A ``learner''
(like AIXI) may converge to the optimal informed decision-maker
(like AI$\mu$) in several senses. Possibly relevant concepts from
statistics are, {\em\idx{consistency}},
{\em\idx{self-tunability}}, {\em\idx{self-optimization}},
{\em\idx{efficiency}}, {\em\idx{unbiasedness}}, {\em asymptotic}
or\indxs{asymptotic}{convergence}\indxs{finite}{convergence} {\em
finite} convergence \cite{Kumar:86}, \idx{Pareto optimality}, and
some more defined in Section~\ref{secAIsep}. Some concepts are
stronger than necessary, others are weaker than desirable but
suitable to start with. Self-optimization is defined as the
asymptotic convergence of the average true value $\odm
V_{1m}^{p^\xi \mu}$ of AI$\xi$ to the optimal value $\odm
V_{1m}^{\best\mu}$. Apart from convergence speed,
self-optimization of AIXI would most closely correspond to the
loss bounds proven for SP. We investigate which properties are
desirable and under which circumstances the AIXI model satisfies
these properties. We will show that no universal model, including
AIXI, can in general be self-optimizing. On the other hand, we show
that AIXI is Pareto optimal in the sense that there is no other
policy that performs better or equal in all environments, and
strictly better in at least one.

\indxs{limited}{environmental class}
\indxs{restricted}{concept class}
\paragraph{Limited environmental classes}
The problem of defining and proving general value bounds becomes
more feasible by considering, in a first step, restricted concept
classes. We analyze AIXI for known classes (like Markovian or
factorizable environments) and especially for the new classes
(forgetful, relevant, asymptotically learnable, farsighted,
uniform, pseudo-passive, and passive) defined later in
Section~\ref{secAIsep}. In Section~\ref{chApply} we study the
behavior of AIXI in various standard problem classes, including
sequence prediction, strategic games, function minimization, and
supervised learning.

\indxs{general}{Bayes mixture}
\indxs{general Bayes mixture}{AI$\xi$ model}
\paragraph{Generalization of AIXI to general Bayes mixtures}
The other approach is to generalize AIXI to AI$\zeta$, where
$\zeta()=\sum_{\nu\in\M}w_\nu\nu()$ is a general Bayes mixture of
distributions $\nu$ in some class $\M$. If $\M$ is the multi-set
of enumerable semimeasures enumerated by a Turing machine, then
AI$\zeta$ coincides with AIXI. If $\M$ is the (multi)set of
passive effective environments, then AIXI reduces to the
$\Lambda_\xi$ predictor that has been shown to perform well. One
can show that these loss/value bounds generalize to wider classes,
at least asymptotically \cite{Hutter:02selfopt}. Promising
classes are, again, the ones described in Section~\ref{secAIsep}.
In particular, for ergodic {\sc mdp}s we showed that AI$\zeta$ is
self-optimizing\indxs{self-optimizing}{policy}. Obviously, the
least we must demand from $\M$ to have a chance of finding a
self-optimizing policy is that there exists some self-optimizing
policy at all. The key result in \cite{Hutter:02selfopt} is that
this necessary condition is also sufficient. More generally, the key is
not to prove absolute results for specific problem classes, but to
prove relative results of the form ``if there exists a policy with
certain desirable properties, then AI$\zeta$ also possesses these
desirable properties''. If there are tasks that cannot be solved
by any policy, AI$\zeta$ cannot be blamed for failing.
Environmental classes that allow for self-optimizing policies
include bandits, i.i.d.\ processes, classification tasks, certain
classes of {\sc pomdp}s, $k^{th}$-order ergodic {\sc mdp}s,
factorizable environments, repeated games, and prediction
problems. Note that in this approach we have for each
environmental class a corresponding model AI$\zeta$, whereas in
the approach pursued in this article the same universal AIXI model
is analyzed for all environmental classes.

\index{optimality!by construction}
\index{Bandit problem}
\paragraph{Optimality by construction}
A possible further approach toward an optimality ``proof'' is to
regard AIXI as {\em optimal by construction}.
This perspective is common in various (simpler) settings.
For instance, in bandit problems, where pulling arm $i$ leads to
reward $1$ ($0$) with unknown probability $p_i$ ($1-p_i$), the
traditional Bayesian solution to the uncertainty about $p_i$ is to
assume a uniform (or Beta) prior over $p_i$ and to maximize
the (subjectively) expected reward sum over multiple trials. The
exact solution (in terms of Gittins indices) is widely regarded as
``optimal'', although justified alternative approaches exist.
Similarly, but simpler, assuming a uniform subjective prior over
the Bernoulli parameter $p_{(i)}\in[0,1]$, one arrives at the
reasonable, but more controversial, Laplace rule for predicting
i.i.d.\ sequences.
AIXI is similar in the sense that the unknown $\mu\in\M$ is the
analogue of the unknown $p\in[0,1]$, and the prior beliefs
$w_\nu=2^{-K(\nu)}$ justified by Occam's razor are the analogue of
a uniform distribution over [0,1].
In the same sense as Gittins' solution to the bandit problem and
Laplace' rule for Bernoulli sequences, AIXI may also be
regarded as optimal by construction.
Theorems relating AIXI to AI$\mu$ would not be regarded as
optimality proofs of AIXI, but just as how much harder it becomes
to operate when $\mu$ is unknown, i.e.\
the achievements of the first three approaches are simply
reinterpreted.

\subsection{Value Bounds and Separability Concepts}\label{secAIsep}
\indxs{value}{bound}\indxs{separability}{concepts}

\paragraph{Introduction}
The values $V_{km}$ associated with the AI systems correspond
roughly to the negative loss $-L_n^\Lambda$ of the SP
systems. In SP, we were interested in small bounds for the loss
excess $L_n^{\smash{\Lambda_\xi}} - L_n^\Lambda$. Unfortunately, simple
value bounds for AI$\xi$ in terms of $V_{km}$ analogous to the
loss bound (\ref{thULoss}) do not hold. We even have difficulties in
specifying what we can expect to hold for AI$\xi$ or any AI system
that claims to be universally optimal. Consequently, we cannot
have a proof if we don't know what to prove. In SP, the only
important property of $\mu$ for proving loss bounds was its
complexity $K(\mu)$. We will see that in the AI case, there are no
useful bounds in terms of $K(\mu)$ only. We either have to study
restricted problem classes or consider bounds depending on other
properties of $\mu$, rather than on its complexity only. In the
following, we will exhibit the difficulties by two examples and
introduce concepts that may be useful for proving value bounds.
Despite the difficulties in even claiming useful value bounds, we
nevertheless, firmly believe that the order relation
(Definition~\ref{defaiorder}) correctly formalizes the intuitive meaning of
intelligence and, hence, that the AI$\xi$ agent is universally optimal.

\paragraph{(Pseudo) Passive $\mu$ and the HeavenHell example}
\indxs{pseudo-passive}{environment}\index{HeavenHell example}%
In the following we choose $m_k = m$. We want to compare the
true, i.e.\ $\mu$-expected value $V^\mu_{1m}$ of a $\mu$-independent
universal policy $p^{best}$ with any other policy $p$.
Naively, we might expect the existence of a policy $p^{best}$
that maximizes $V^\mu_{1m}$, apart from additive corrections of
lower order for $m\to\infty$
\beq\label{cximu}
  V_{1m}^{p^{best}\mu} \;\geq\; V_{1m}^{p\mu} - o(...)
  \quad \forall\mu,p
\eeq
Such policies are sometimes called self-optimizing
\cite{Kumar:86}. Note that $V_{1m}^{p^\mu\mu} \geq
V_{1m}^{p\mu}\,\forall p$, but $p^\mu$ is not a candidate for (a
universal) $p^{best}$ as it depends on $\mu$. On the other hand,
the policy $p^\xi $ of the AI$\xi$ agent maximizes $V^\xi_{1m}$ by
definition ($p^\xi \succeq p$). As $V^\xi_{1m}$ is thought to be a
guess of $V^\mu_{1m}$, we might expect $p^{best} = p^\xi $ to
approximately maximize $V^\mu_{1m}$, i.e.\ (\ref{cximu}) to hold.
Let us consider the problem class (set of environments)
$\M=\{\mu_0,\mu_1\}$ with $\Y = \R =\{0,1\}$ and $r_k
=\delta_{iy_1}$ in environment $\mu_i$, where the Kronecker symbol
$\delta_{xy}$ is defined as 1 for $x=y$ and 0 otherwise. The first
action $y_1$ decides whether you go to heaven with all future
rewards $r_k$ being $1$ (good) or to hell with all future rewards
being $0$ (bad). Note that $\mu_i$ are
(deterministic, non-ergodic) {\sc mdp}s:

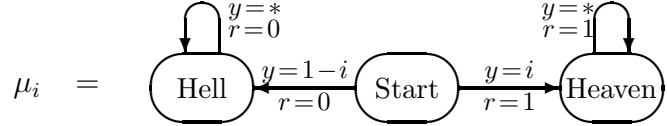
\begin{center}
\footnotesize\unitlength=1.5ex
\begin{picture}(34,7)(0,0)
\thicklines
\put(1,2){\makebox(0,0)[rc]{\normalsize$\mu_i\quad=$}}
\put(7,2){\oval(6,4)}\put(7,2){\makebox(0,0)[cc]{\small Hell}}
\put(7,6){\oval(2,2)[t]}\put(6,6){\vector(0,-1){2}}\put(8,6){\line(0,-1){2}}
\put(8,6){\makebox(0,0)[lb]{ $y=*$}}
\put(8,6){\makebox(0,0)[lt]{ $r=0$}}
\put(16,2){\vector(-1,0){6}}
\put(13,2.3){\makebox(0,0)[cb]{$y=1-i$}}
\put(13,1.7){\makebox(0,0)[ct]{$r=0$}}
\put(19,2){\oval(6,4)}\put(19,2){\makebox(0,0)[cc]{\small Start}}
\put(22,2){\vector(1,0){6}}
\put(25,2.3){\makebox(0,0)[cb]{$y=i$}}
\put(25,1.7){\makebox(0,0)[ct]{$r=1$}}
\put(31,2){\oval(6,4)}\put(31,2){\makebox(0,0)[cc]{\small Heaven}}
\put(31,6){\oval(2,2)[t]}\put(32,6){\vector(0,-1){2}}\put(30,6){\line(0,-1){2}}
\put(30,6){\makebox(0,0)[rb]{ $y=*$}}
\put(30,6){\makebox(0,0)[rt]{ $r=1$}}
\end{picture}
\end{center}
\noindent It is clear that if $\mu_i$, i.e.\ $i$ is known,
the optimal policy $p^{\mu_i}$ is to output $y_1 = i$ in the first
cycle with $V_{1m}^{p^{\mu_i}\mu} = m$. On the other hand, any
unbiased policy $p^{best}$ independent of the actual $\mu$ either
outputs $y_1 = 1$ or $y_1 = 0$. Independent of the actual choice
$y_1$, there is always an environment ($\mu = \mu_{1-y_1}$) for
which this choice is catastrophic ($V_{1m}^{p^{best}\mu} = 0$). No
single agent can perform well in both environments $\mu_0$ {\it
and} $\mu_1$. The r.h.s.\ of (\ref{cximu}) equals $m - o(m)$ for
$p = p^\mu$. For all $p^{best}$ there is a $\mu$ for which the
l.h.s.\ is zero. We have shown that no $p^{best}$ can satisfy
(\ref{cximu}) for all $\mu$ and $p$, so we cannot expect $p^\xi $
to do so. Nevertheless, there are problem classes for which
(\ref{cximu}) holds, for instance SP. For SP, (\ref{cximu}) is
just a reformulation of (\ref{thULoss}) with an appropriate choice
for $p^{best}$, namely $\Lambda_\xi$ (which differs from $p^\xi $,
see next section). We expect (\ref{cximu}) to hold for all
inductive problems in which the environment is not
influenced\footnote{Of course, the reward feedback $r_k$ depends
on the agent's output. What we have in mind is, like in sequence
prediction, that the true sequence is not influenced by the
agent.} by the output of the agent.
\indxs{passive}{environment}\indxs{inductive}{environment}%
We want\indxs{pseudo-passive}{environment} to call these $\mu$,
{\it passive} or {\it inductive} environments. Further, we want to
call $\M$ and $\mu\in\M$ satisfying (\ref{cximu}) with $p^{best} =
p^\xi $ {\it pseudo-passive}. So we expect inductive $\mu$ to be
pseudo-passive.

\paragraph{The OnlyOne example}\index{OnlyOne example}
Let us give a further example to demonstrate the difficulties in
establishing value bounds. Let $\X=\R =\{0,1\}$ and $|\Y|$ be large.
We consider all (deterministic) environments in which a single
complex output $y^*$ is correct ($r = 1$) and all others are wrong
($r = 0$). The problem class $\M$ is defined by
\beqn
  \M:=\{\mu_{y^*}: y^*\!\in\!\Y,\; K(y^*)=_\lfloor\!\!\log|\Y|_\rfloor\},
  \qmbox{where} \mu_{y^*}(y\!x_{<k}y_k\pb 1):=\delta_{y_ky^*}\,\forall k.
\eeqn
There are $N\eqm|\Y|$ such $y^*$. The only way a $\mu$-independent
policy $p$ can find the correct $y^*$, is by trying one $y$ after
the other in a certain order. In the first $N - 1$ cycles, at most
$N - 1$ different $y$ are tested. As there are $N$ different
possible $y^*$, there is always a $\mu\in\M$ for which $p$ gives
erroneous outputs in the first $N - 1$ cycles. The number of
errors is $E_\infty^p \geq N - 1 \eqm|\Y|\eqm 2^{K(y^*)}\eqm
2^{K(\mu)}$ for this $\mu$. As this is true for any $p$, it is
also true for the AI$\xi$ model, hence $E_k^{p^\xi} \leq
2^{K(\mu)}$ is the best possible error bound we can expect that
depends on $K(\mu)$ only. Actually, we will derive such a bound in
Section~\ref{secSP} for inductive environments. Unfortunately, as
we are mainly interested in the cycle region $k\ll|\Y|\eqm
2^{K(\mu)}$ (see Section~\ref{secCL}) this bound is vacuous. There
are no interesting bounds for deterministic $\mu$ depending on
$K(\mu)$ only, unlike the SP case. Bounds must either depend on
additional properties of $\mu$ or we have to consider specialized
bounds for restricted problem classes. The case of probabilistic
$\mu$ is similar. Whereas for SP there are useful bounds in terms
of $L_k^{\smash{\Lambda_\mu}}$ and $K(\mu)$, there are no such
bounds for AI$\xi$. Again, this is not a drawback of AI$\xi$ since
for no unbiased AI system could the errors/rewards be bound in
terms of $K(\mu)$ and the errors/rewards of AI$\mu$ only.

\index{posterization}\index{bound!boost}\index{boosting!bound}%
There is a way to make use of gross (e.g.\ $2^{K(\mu)}$) bounds.
Assume that after a reasonable number of cycles $k$, the
information $\hh x_{<k}$ perceived by the AI$\xi$ agent contains a
lot of information about the true environment $\mu$. The
information in $\hh x_{<k}$ might be coded in any form. Let us
assume that the complexity $K(\mu|\hh x_{<k})$ of $\mu$ under the
condition that $\hh x_{<k}$ is known, is of order 1. Consider a
theorem, bounding the sum of rewards or of other quantities over
cycles $1...\infty$ in terms of $f(K(\mu))$ for a function $f$
with $f(O(1)) = O(1)$, like $f(n) = 2^n$. Then, there will be a
bound for cycles $k...\infty$ in terms of $\approx f(K(\mu|\hh
x_{<k})) = O(1)$. Hence, a bound like $2^{K(\mu)}$ can be replaced
by small bound $\approx 2^{K(\mu|\hh x_{<k})} = O(1)$ after $k$
cycles. All one has to show/ensure/assume is that enough
information about $\mu$ is presented (in any form) in the first
$k$ cycles. In this way, even a gross bound could become useful.
In Section~\ref{secEX} we use a similar argument to prove that
AI$\xi$ is able to learn supervised.

\paragraph{Asymptotic learnability}
\index{asymptotic!learnability}\index{learnable!asymptotically}%
In the following, we weaken (\ref{cximu}) in the hope of getting a
bound applicable to wider problem classes than the passive one.
Consider the I/O sequence $\hh y_1\hh x_1...\hh y_n\hh x_n$ caused
by AI$\xi$. On history $\hh y\!\hh x_{<k}$, AI$\xi$ will output
$\hh y_k \equiv\hh y^\xi_k$ in cycle $k$. Let us compare this to
$\hh y^\mu_k$ what AI$\mu$ would output, still on the same history
$\hh y\!\hh x_{<k}$ produced by AI$\xi$. As AI$\mu$ maximizes the
$\mu$-expected value, AI$\xi$ causes lower (or at best equal)
$V_{km_k}^\mu$ if $\hh y^\xi_k$ differs from $\hh y^\mu_k$. Let
$D_{n\mu\xi} := \E[\sum_{k=1}^n 1 - \delta_{\hh y^\mu_k,\hh
y^\xi_k}]$ be the $\mu$-expected number of suboptimal
choices of AI$\xi$, i.e.\ outputs different from AI$\mu$ in the
first $n$ cycles. One might weigh the deviating cases by their
severity. In particular, when the $\mu$-expected rewards
$V_{km_k}^{p\mu}$ for $\hh y^\xi_k$ and $\hh y^\mu_k$ are equal or
close to each other, this should be taken into account in a
definition of $D_{n\mu\xi}$, e.g.\ by a weight factor
$[V_{km}^{\best\mu}(y\!x_{<k}) - V_{km}^{p^\xi\mu}(y\!x_{<k})]$.
These details do not matter in the following qualitative
discussion.\indxs{suboptimal}{decision}\index{decision!wrong} The
important difference to (\ref{cximu}) is that here we stick to the
history produced by AI$\xi$ and count a wrong decision as, at
most, one error. The wrong decision in the HeavenHell example in
the first cycle no longer counts as losing $m$ rewards, but counts
as one wrong decision. In a sense, this is fairer. One shouldn't
blame somebody too much who makes a single wrong decision for
which he just has too little information available, in order to
make a correct decision. The AI$\xi$ model would deserve to be
called asymptotically optimal if the probability of making a
wrong decision tends to zero, i.e.\ if
\beq\label{Doon}
  D_{n\mu\xi}/n\to 0 \qmbox{for} n\to\infty, \quad\mbox{i.e.}\quad
  D_{n\mu\xi} \;=\; o(n).
\eeq
We say that $\mu$ can be {\it asymptotically learned} (by AI$\xi$)
if (\ref{Doon}) is satisfied. We claim that AI$\xi$ (for $m_k \to
\infty$) can asymptotically learn every problem $\mu$ of
relevance, i.e.\ AI$\xi$ is asymptotically
optimal.\indxs{relevant}{problem} We included the qualifier {\it
of relevance}, as we are not sure whether there could be strange
$\mu$ spoiling (\ref{Doon}) but we expect those $\mu$ to be
irrelevant from the perspective of AI. In the field of Learning,
there are many asymptotic learnability theorems, often not too
difficult to prove. So a proof of (\ref{Doon}) might also be
feasible. Unfortunately, asymptotic learnability theorems are
often too weak to be useful from a practical point of view.
Nevertheless, they point in the right direction.

\paragraph{Uniform $\mu$}\indxs{uniform}{environment}
From the convergence (\ref{aixitomu}) of $\xi\to\mu$ we might
expect $V_{km_k}^{p\xi} \to V_{km_k}^{p\mu}$ for all $p$, 
and hence we might also expect $\hh y^\xi_k$
defined in (\ref{ydotxi}) to converge to $\hh y^\mu_k$ defined in
(\ref{ydotrec}) for $k \to \infty$. The
first problem is that if the $V_{km_k}$ for the different choices
of $y_k$ are nearly equal, then even if $V_{km_k}^{p\xi} \approx
V_{km_k}^{p\mu}$, $\hh y^\xi_k \neq \hh y^\mu_k$ is possible due to
the non-continuity of $\arg\max_{y_k}$. This can be cured by a
weighted $D_{n\mu\xi}$ as described above. More serious is the
second problem we explain for $h_k = 1$ and $\X = \R = \{0,1\}$.
For $\hh y^\xi_k \equiv \arg\max_{y_k}\xi(\hh y\!\hh r_{<k}y_k\pb
1)$ to converge to $\hh y^\mu_k \equiv \arg\max_{y_k}\mu(\hh
y\!\hh r_{<k}y_k\pb 1)$, it is not sufficient to know that
$\xi(\hh y\!\hh r_{<k}\hh y\!\hh{\pb r}_k) \to \mu(\hh y\!\hh
r_{<k}\hh y\!\hh{\pb r}_k)$ as proven in (\ref{aixitomu}). We need
convergence not only for the true output $\hh y_k$,
but also for alternative outputs $y_k$. $\hh
y^\xi_k$ converges to $\hh y^\mu_k$ if $\xi$ converges uniformly
to $\mu$, i.e.\ if in addition to (\ref{aixitomu})
\indxs{uniform}{convergence}
\beq\label{uniform}
  \big|\mu(y\!x_{<k}y'_k\pb x'_k)-\xi(y\!x_{<k}y'_k\pb x'_k)\big|
  \;<\; c\!\cdot\!
  \big|\mu(y\!x_{<k}y\!\pb x_k)-\xi(y\!x_{<k}y\!\pb x_k)\big|
  \quad\forall y'_kx'_k
\eeq
holds for some constant $c$ (at least in a $\mu$-expected sense).
We call $\mu$ satisfying (\ref{uniform}) {\it uniform}. For
uniform $\mu$ one can show (\ref{Doon}) with appropriately
weighted $D_{n\mu\xi}$ and bounded horizon $h_k < h_{max}$.
Unfortunately there are relevant $\mu$ that are not uniform.

\paragraph{Other concepts}
\indxs{Markov}{environment}\index{Markov!$k$-th order}%
\indxs{factorizable}{environment}%
\indxs{stationary}{environment}%
\indxs{forgetful}{environment}%
\indxs{farsighted}{environment}%
In the following, we briefly mention some further concepts. A {\it
Markovian} $\mu$ is defined as depending only on the last cycle,
i.e.\ $\mu(y\!x_{<k}y\!\pb x_k) = \mu_k(x_{k-1}y\!\pb x_k)$. We
say $\mu$ is {\it generalized ($l^{th}$-order) Markovian}, if
$\mu(y\!x_{<k}y\!\pb x_k) = \mu_k(x_{k-l}y\!x_{k-l+1:k-1}y\!\pb
x_k)$ for fixed $l$. This property has some similarities to {\it
factorizable} $\mu$ defined in (\ref{facmu}). If further $\mu_k
\equiv \mu_1\forall k$, $\mu$ is called {\it stationary}. Further,
we call $\mu$ ($\xi$) {\it forgetful} if $\mu(y\!x_{<k}y\!\pb
x_k)$ ($\xi(y\!x_{<k}y\!\pb x_k)$) become(s) independent of $y\!x_{<l}$
for fixed $l$ and $k\to\infty$ with $\mu$-probability 1. Further,
we say $\mu$ is {\it farsighted} if $\lim_{m_k\to\infty}\hh
y_k^{(m_k)}$ exists. More details will be given in
Section~\ref{secHorizon}, where we also give an example of a farsighted
$\mu$ for which nevertheless the limit $m_k \to \infty$ makes no
sense.

\paragraph{Summary}
We have introduced several concepts that might be useful for
proving value bounds, including forgetful, relevant, asymptotically
learnable, farsighted, uniform, (generalized) Markovian, factorizable
and (pseudo)passive $\mu$. We have sorted them here, approximately in
the order of decreasing generality. We will call them {\it
separability concepts}. The more general (like relevant,
asymptotically learnable and farsighted) $\mu$ will be called
weakly separable, the more restrictive (like (pseudo) passive and
factorizable) $\mu$ will be called strongly separable, but we will
use these qualifiers in a more qualitative, rather than rigid
sense. Other (non-separability) concepts are deterministic $\mu$
and, of course, the class of all chronological $\mu$.

\indxs{Pareto optimality}{AI$\xi$ model}
\subsection{Pareto Optimality of AI$\xi$}
This subsection shows Pareto-opimtality of AI$\xi$ analogous to
SP. The total $\mu$-expected reward $V_\mu^{p^\xi}$ of
policy $p^\xi$ of the AI$\xi$ model is of central interest in
judging the performance of AI$\xi$. We know that there {\em are}
policies (e.g.\ $p^\mu$ of AI$\mu$) with higher $\mu$-value
($V_\mu^\best\geq V_\mu^{p^\xi}$). In general, every
policy based on an estimate $\rho$ of $\mu$ that is closer to
$\mu$ than $\xi$ is, outperforms $p^\xi$ in environment $\mu$,
simply because it is more tailored toward $\mu$. On the other
hand, such a system probably performs worse than $p^\xi$ in other
environments.
Since we do not know $\mu$ in advance we may ask whether there
exists a policy $p$ with better or equal performance than $p^\xi$
in {\em all} environments $\nu\in\M$ and a strictly better
performance for one $\nu \in\M$. This would clearly render $p^\xi$
suboptimal. One can show that there is no such $p$ \cite{Hutter:02selfopt}

\fdefinition{defaiPareto}{Pareto Optimality}{
A policy $\tilde p$ is called Pareto optimal if there is no other
policy $p$ with $V_{\nu}^p\geq V_{\nu}^{\tilde
p}$ for all $\nu \in \M$ and strict inequality for at
least one $\nu$.
}

\ftheorem{thaiPareto}{Pareto Optimality}{
AI$\xi$ alias $p^\xi$ is Pareto optimal.
}

\noindent Pareto optimality should be regarded as a necessary
condition for an agent aiming to be optimal. From a practical
point of view, a significant increase of $V$ for many environments
$\nu$ may be desirable, even if this causes a small decrease of $V$
for a few other $\nu$. The impossibility of such a ``balanced''
improvement is a more demanding condition on $p^\xi$ than pure
Pareto optimality. In \cite{Hutter:02selfopt} it has been shown
that AI$\xi$ is also balanced Pareto optimal.

\subsection{The Choice of the Horizon}\label{secHorizon}
\indxs{horizon}{problem}%
The only significant arbitrariness in the
AI$\xi$ model lies in the choice of the horizon function $h_k
\equiv m_k - k + 1$. We discuss some choices that seem to be
natural and give preliminary conclusions at the end. We will not
discuss ad hoc choices of $h_k$ for specific problems (like the
discussion in Section~\ref{secSG} in the
context of finite strategic games). We are interested in universal
choices of $m_k$.

\indxs{fixed}{horizon}
\paragraph{Fixed horizon}
If the lifetime of the agent is known to be $m$, which is in
practice always large but finite, then the choice $m_k = m$
maximizes correctly the expected future reward. Lifetime $m$ is usually not
known in advance, as in many cases the time we are willing to run
an agent depends on the quality of its outputs. For this reason,
it is often desirable that good outputs are not delayed too much,
if this results in a marginal reward increase only. This can be
incorporated by damping the future rewards. If, for instance, the
probability of survival in a cycle is $\gamma<1$, an exponential
damping (geometric discount) $r_k := r'_k \cdot \gamma^k$ is
appropriate, where $r'_k$ are bounded, e.g.\ $r'_k \in [0,1]$.
Expression (\ref{ydotxi}) converges for $m_k \to \infty$ in this
case.$\!$\footnote{More precisely, $\displaystyle \hh
y_k=\arg\max_{y_k}\lim_{m_k\to\infty}V_{km_k}^{\best\xi}(\hh y \hh
x_{<k}y_k)$ exists.} But this does not solve the problem, as we
introduced a new arbitrary time scale $(1-\gamma)^{-1}$. Every
damping introduces a time scale. Taking $\gamma\to 1$ is prone
to the same problems as $m_k\to\infty$ in the undiscounted case
discussed below.

\indxs{dynamic}{horizon}
\indxs{universal}{discounting}
\indxs{harmonic}{discounting}
\paragraph{Dynamic horizon (universal \& harmonic discounting)}
The largest horizon with guaranteed finite and enumerable reward
sum can be obtained by the universal discount $r_k \leadsto r_k
\cdot 2^{-K(k)}$. This discount results in truly farsighted agent
with effective horizon that grows faster than any computable
function.
It is similar to a near-harmonic discount $r_k \leadsto
r_k \cdot k^{-(1+\eps)}$, since $2^{-K(k)} \leq 1/k$ for most $k$
and $2^{-K(k)} \geq c/(k\,\log^2 k)$. More
generally, the time-scale invariant damping factor $r_k = r'_k
\cdot k^{-\alpha}$ introduces a dynamic time scale. In cycle $k$
the contribution of cycle $2^{1/\alpha} \cdot k$ is damped by a
factor $\odt$. The effective horizon $h_k$ in this case is $\sim
k$. The choice $h_k = \beta \cdot k$ with $\beta \sim
2^{1/\alpha}$ qualitatively models the same behavior. We have not
introduced an arbitrary time scale $m$, but limited the
farsightedness to some multiple (or fraction) of the length of the
current history. This avoids the preselection of a global
time scale $m$ or ${1\over 1-\gamma}$. This choice has some appeal, as
it seems that humans of age $k$ years usually do not plan their
lives for more than, perhaps, the next $k$ years ($\beta_{human}
\approx 1$). From a practical point of view this model might serve
all needs, but from a theoretical point we feel uncomfortable with
such a limitation in the horizon from the very beginning. Note
that we have to choose $\beta = O(1)$ because otherwise we would
again introduce a number $\beta$, which has to be justified.
We favor the universal discount $\gamma_k=2^{-K(k)}$, since
it allows us, if desired, to ``mimic'' all other more greedy
behaviors based on other discounts $\gamma_k$ by choosing $r_k\in
[0,c\cdot\gamma_k]\subseteq[0,2^{-K(k)}]$.

\indxs{infinite}{horizon}
\paragraph{Infinite horizon}
The naive limit $m_k \to \infty$ in (\ref{ydotxi}) may turn out
to be well defined and the previous discussion superfluous. In the
following, we suggest a limit that is always well defined (for
finite $\Y$). Let $\hh y_k^{(m_k)}$ be defined as in (\ref{ydotxi})
with dependence on $m_k$ made explicit. Further, let $\hh
\Y_k^{(m)} := \{\,\hh y_k^{(m_k)} : m_k \geq m\}$ be the set
of outputs in cycle $k$ for the choices $m_k = m,m+1,m+2,...$.
Because $\hh \Y_k^{(m)} \supseteq \hh \Y_k^{(m+1)} \neq \{\}$,
we have $\hh \Y_k^{(\infty)} := \bigcap_{m=k}^\infty\hh
\Y_k^{(m)} \neq \{\}$. We define the $m_k = \infty$ model to
output any $\hh y_k^{(\infty)} \in \hh \Y_k^{(\infty)}$. This is
the best output consistent with some arbitrary large choice of
$m_k$. Choosing the lexicographically smallest $\hh
y_k^{(\infty)} \in \hh \Y_k^{(\infty)}$ would correspond to the
lower limit $\underline\lim_{m\to\infty}\hh y_k^{(m)}$, which
always exists (for finite $\Y$). Generally $\hh
y_k^{(\infty)} \in \hh \Y_k^{(\infty)}$ is unique, i.e.\ $|\hh
\Y_k^{(\infty)}| = 1$ iff the naive limit $\lim_{m\to\infty}\hh
y_k^{(m)}$ exists. Note that the limit
$\lim_{m\to\infty}V_{km}^\best(y\!x_{<k})$ need not exist for
this construction.

\indxs{average}{reward}\indxs{differential}{gain}
\paragraph{Average reward and differential gain}
Taking the raw average reward $(r_k +...+ r_m)/(m - k + 1)$ and
$m\to\infty$ also does not help: consider an arbitrary policy for the
first $k$ cycles and the/an optimal policy for the remaining
cycles $k+1...\infty$. In e.g.\ i.i.d.\ environments the limit
exists, but all these policies give the same average value, since
changing a finite number of terms does not affect an infinite
average. In {\sc mdp} environments with a single recurrent class
one can define the relative or differential gain
\cite{Bertsekas:96}. In more general environments (we are
interested in) the differential gain can be infinite, which is
acceptable, since differential gains can still be totally ordered.
The major problem is the {\it existence} of the differential gain,
i.e.\ whether it converges for $m\to\infty$ in
$\SetR\cup\{\infty\}$ at all (and does not oscillate). This is
just the old convergence problem in slightly different form.

\indxs{immortal}{agents}\indxs{lazy}{agents}
\paragraph{Immortal agents are lazy}
The construction in the next to previous paragraph leads to a mathematically elegant,
no-parameter AI$\xi$ model. Unfortunately this is not the end of
the story. The limit $m_k \to \infty$ can cause undesirable
results in the AI$\mu$ model for special $\mu$, which might also
happen in the AI$\xi$ model whatever we define $m_k \to \infty$.
Consider an agent who for every $\sqrt{l}$ consecutive days of
work, can thereafter take $l$ days of holiday. Formally, consider
$\Y = \X = \R = \{0,1\}$. Output $y_k = 0$ shall give reward $r_k
= 0$ and output $y_k = 1$ shall give $r_k = 1$ iff $\hh
y_{k-l-\sqrt l}...\hh y_{k-l} = 0...0$ for some $l$, i.e.\ the
agent can achieve $l$ consecutive positive rewards if there was a
preceding sequence of length at least $\sqrt l$ with $y_k = r_k =
0$. If the lifetime of the AI$\mu$ agent is $m$, it outputs $\hh
y_k = 0$ in the first $s$ cycles and then $\hh y_k = 1$ for the
remaining $s^2$ cycles with $s$ such that $s+s^2=m$. This will
lead to the highest possible total reward $V_{1m} = s^2 =
m+\odt-\sqrt{m+^1\!\!/\!_4}$. Any fragmentation of the $0$ and $1$
sequences would reduce $V_{1m}$, e.g.\ alternatingly working for 2
days and taking 4 days off would give $V_{1m}={2\over 3}m$. For $m
\to \infty$ the AI$\mu$ agent can and will delay the point $s$ of
switching to $\hh y_k = 1$ indefinitely and always output $0$
leading to total reward $0$, obviously the worst possible
behavior. The AI$\xi$ agent will explore the above rule after a
while of trying $y_k = 0/1$ and then applies the same behavior as
the AI$\mu$ agent, since the simplest rules covering past data
dominate $\xi$. For finite $m$ this is exactly what we want, but
for infinite $m$ the AI$\xi$ model (probably) fails, just as the
AI$\mu$ model does. The good point is that this is not a weakness
of the AI$\xi$ model in particular, as AI$\mu$ fails too. The bad
point is that $m_k\to\infty$ has far-reaching consequences, even
when starting from an already very large $m_k=m$. This is because
the $\mu$ of this example is highly nonlocal in time, i.e.\
it may violate one of our weak separability conditions.

\paragraph{Conclusions}
We are not sure whether the choice of $m_k$ is of marginal
importance, as long as $m_k$ is chosen sufficiently large and of
low complexity, $m_k=2^{2^{16}}$ for instance, or whether the
choice of $m_k$ will turn out to be a central topic for the
AI$\xi$ model or for the planning aspect of any AI system in
general. We suppose that the limit $m_k\to\infty$ for the
AI$\xi$ model results in correct behavior for weakly separable
$\mu$. A proof of this
conjecture, if true, would probably give interesting insights.

\subsection{Outlook}\label{secAIxiOut}

\indxs{prediction}{expert advice}
\paragraph{Expert advice approach}
We considered expected performance bounds for predictions based on
Solomonoff's prior. The other, dual, currently very popular
approach, is ``prediction with expert advice'' (PEA) invented by
Littlestone and Warmuth (1989)\nocite{Littlestone:89}, and Vovk
(1992)\nocite{Vovk:92}. Whereas PEA performs well in any
environment, but only relative to a given set of experts , our
$\Lambda_\xi$ predictor competes with {\em any} other predictor,
but only in expectation for environments with computable
distribution. It seems philosophically less compromising to make
assumptions on prediction strategies than on the environment,
however weak. One could investigate whether PEA can be generalized
to the case of active agents, which would result in a model dual
to AIXI. We believe the answer to be negative, which on the
positive side would show the necessity of Occam's razor
assumption, and the distinguishedness of AIXI.

\indxs{random}{action}
\paragraph{Actions as random variables}
The uniqueness for the choice of the generalized $\xi$
(\ref{xidef}) in the AIXI model could be explored. From the
originally many alternatives, which could all be ruled out, there
is one alternative which still seems possible. Instead of defining
$\xi$ as in (\ref{uniMAI}) one could treat the agent's
actions $y$ also as universally distributed random variables and
then conditionalize $\xi$ on $y$ by \mrcp.

\indxs{structure}{AI$\xi$ model}
\indxs{axiomatic approach}{AI$\xi$ model}
\paragraph{Structure of AIXI}
The algebraic properties and the structure of AIXI could be
investigated in more depth. This would extract the essentials
from AIXI which finally could lead to an axiomatic
characterization of AIXI. The benefit is as in any axiomatic
approach. It would clearly exhibit the assumptions, separate the
essentials from technicalities, simplify understanding and,
most important, guide in finding proofs.

\index{policy!restricted class}
\paragraph{Restricted policy classes}
The development in this section could be scaled down to restricted
classes of policies $\cal P$. One may define
$V^*=\arg\max_{p\in\cal P}V^p$. For instance, consider a finite
class of quickly computable policies. For {\sc mdp}s, $\xi$ is
quickly computable and $V_\xi^p$ can be (efficiently) computed by
\idx{Monte Carlo} sampling. Maximizing over the finitely many
policies $p\in\cal P$ selects the asymptotically best policy
$p^\xi$ from $\cal P$ for all (ergodic) {\sc mdp}s
\cite{Hutter:02selfopt}.

\subsection{Conclusions}\label{secAIxiCon}
All tasks that require intelligence to be solved can naturally be
formulated as a maximization of some expected utility in the
framework of agents. We gave an explicit expression
(\ref{ydotrec}) of such a decision-theoretic agent. The main
remaining problem is the unknown prior probability distribution
$\mu^\AI$ of the environment(s). Conventional learning algorithms
are unsuitable, because they can neither handle large
(unstructured) state spaces nor do they converge in the
theoretically minimal number of cycles nor can they handle
non-stationary environments appropriately. On the other hand, the
universal semimeasure $\xi$ (\ref{xidef}), based on ideas from
algorithmic information theory, solves the problem of the unknown
prior distribution for induction problems. No explicit learning
procedure is necessary, as $\xi$ automatically converges to $\mu$.
We unified the theory of universal sequence prediction with the
decision-theoretic agent by replacing the unknown true prior
$\mu^\AI$ by an appropriately generalized universal semimeasure
$\xi^\AI$. We gave strong arguments that the resulting AI$\xi$
model is universally optimal. Furthermore, possible solutions to
the horizon problem were discussed. In Section~\ref{chApply}
we present a number of problem classes, and outline how the
AI$\xi$ model can solve them. They include sequence prediction,
strategic games, function minimization and, especially, how
AI$\xi$ learns to learn supervised. In Section~\ref{chTime} we
develop a modified time-bounded (computable) AIXI$tl$ version.

\section{Important Problem Classes}\label{chApply}

In order to give further support for the universality and
optimality of the AI$\xi$ theory, we apply AI$\xi$ in this section
to a number of problem classes. They include sequence
prediction, strategic games, function minimization and,
especially, how AI$\xi$ learns to learn supervised. For some
classes we give concrete examples to illuminate the scope of the
problem class. We first formulate each problem class in its
natural way (when $\mu^{\mbox{\tiny problem}}$ is known) and then
construct a formulation within the AI$\mu$ model and prove its
equivalence. We then consider the consequences of replacing $\mu$
by $\xi$. The main goal is to understand why and how the problems
are solved by AI$\xi$. We only highlight special aspects of each
problem class. Sections~\ref{secSP}--\ref{secOther} together
should give a better picture of the AI$\xi$ model. We do not study
every aspect for every problem class. The subsections may be read
selectively, and are not essential to understand the remainder.

\subsection{Sequence Prediction (SP)}\label{secSP}
We introduced the AI$\xi$ model as a unification of ideas
of sequential decision theory and universal probability
distribution. We might expect AI$\xi$ to behave identically to
SP$\xi$, when faced with a sequence prediction problem, but
things are not that simple, as we will see.

\paragraph{Using the AI$\mu$ model for sequence prediction}
We saw in Section~\ref{chSP} how to predict sequences for
known and unknown prior distribution $\mu^\SP$. Here we consider
binary sequences\footnote{We use $z_k$ to avoid notational
conflicts with the agent's inputs $x_k$.} $z_1z_2z_3...\in
\B^\infty$ with known prior probability
$\mu^\SP(\pb{z_1z_2z_3...})$.

We want to show how the AI$\mu$ model can be used for sequence
prediction. We will see that it makes the same prediction as the
SP$\mu$ agent. For simplicity we only discuss the special
error loss $\ell_{xy}=1-\delta_{xy}$, where $\delta$ is the
Kronecker symbol, defined as $\delta_{ab} = 1$ for $a = b$ and $0$
otherwise. First, we have to specify {\it how} the AI$\mu$ model
should be used for sequence prediction. The following choice is
natural:

The system's output $y_k$ is interpreted as a prediction for the
$k^{th}$ bit $z_k$ of the string under consideration. This
means that $y_k$ is binary ($y_k \in \B =: \Y$). As a
reaction of the environment, the agent receives reward $r_k = 1$
if the prediction was correct ($y_k = z_k$), or $r_k = 0$ if
the prediction was erroneous ($y_k \neq z_k$). The question is
what the observation $o_k$ in the next cycle should be. One choice
would be to inform the agent about the correct $k^{th}$ bit
of the string and set $o_k=z_k$. But as from
the reward $r_k$ in conjunction with the prediction $y_k$, the true
bit $z_k=\delta_{y_kr_k}$ can be inferred, this information is
redundant. There is no
need for this additional feedback. So we set $o_k = \epstr \in \O =
\{\epstr\}$, thus having $x_k \equiv r_k\in\R\equiv\X=\{0,1\}$. The
agent's performance does not change when we include this
redundant information; it merely complicates the notation. The prior
probability $\mu^\AI$ of the AI$\mu$ model is
\beq\label{muaisp}
  \mu^\AI(y_1\pb x_1 ...y_k\pb x_k) \;=\;
  \mu^\AI(y_1\pb r_1...y_k\pb r_k) \;=\;
  \mu^\SP(\pb{\delta_{y_1 r_1}...\delta_{y_k r_k}}) \;=\;
  \mu^\SP(\pb{z_1...z_k})
\eeq
In the following, we will drop the superscripts of $\mu$ because
they are clear from the arguments of $\mu$ and the $\mu$ equal in
any case. It is intuitively clear and can formally be shown
\cite{Hutter:00kcunai,Hutter:04uaibook} that maximizing the future reward
$V_{km}^\mu$ is identical to greedily maximizing the immediate
expected reward $V_{kk}^\mu$. There is no exploration-exploitation
tradeoff in the prediction case. Hence, AI$\mu$ acts with
\beq\label{ebestysp}
  \hh y_k \;=\; \arg\max_{y_k}
  V_{kk}^{\best\mu}(\hh y\!\hh x_{<k}y_k)
  \;=\; \arg\max_{y_k}\sum_{r_k}r_k
  \!\cdot\!\mu^\AI(\hh y\!\hh r_{<k}y\!\pb r_k) \;=\;
  \arg\max_{z_k}\mu^\SP(\hh z_1...\hh z_{k-1}\pb z_k)
\eeq
The first equation is the definition of the agent's action
(\ref{pbestrec}) with $m_k$ replaced by $k$. In the second
equation we used the definition (\ref{defAImuVi}) of $V_{km}$. In
the last equation we used (\ref{muaisp}) and
$r_k=\delta_{y_kz_k}$.

So, the AI$\mu$ model predicts that $z_k$ that has maximal
$\mu$-probability, given $\hh z_1...\hh z_{k-1}$. This prediction
is independent of the choice of $m_k$. It is exactly the
prediction scheme of the sequence predictor SP$\mu$ with known
prior described in Section~\ref{secErr} (with special error loss).
As this model was optimal, AI$\mu$ is optimal too, i.e.\ has
minimal number of expected errors (maximal $\mu$-expected reward)
as compared to any other sequence prediction scheme. From this, it
is clear that the value $V_{km}^{\best\mu}$ must be closely
related to the expected error $E_m^{\smash{\Lambda_\mu}}$
(\ref{rholossi}). Indeed one can show that $V_{1m}^{\best\mu} = m
- E_m^{\smash{\Lambda_\mu}}$, and similarly for general loss
functions.

\paragraph{Using the AI$\xi$ model for sequence prediction}
Now we want to use the universal AI$\xi$ model instead of AI$\mu$
for sequence prediction and try to derive error/loss bounds analogous
to (\ref{thULoss}). Like in the AI$\mu$ case, the agent's output
$y_k$ in cycle $k$ is interpreted as a prediction for the $k^{th}$
bit $z_k$ of the string under consideration. The reward is
$r_k=\delta_{y_kz_k}$ and there are no other inputs
$o_k=\epsilon$. What makes the analysis more difficult is that
$\xi$ is not symmetric in $y_ir_i\leftrightarrow(1-y_i)(1-r_i)$
and (\ref{muaisp}) does not hold for $\xi$. On the other hand,
$\xi^\AI$ converges to $\mu^\AI$ in the limit (\ref{aixitomu}),
and (\ref{muaisp}) should hold asymptotically for $\xi$ in some
sense. So we expect that everything proven for AI$\mu$ holds
approximately for AI$\xi$. The AI$\xi$ model should behave
similarly to Solomonoff
prediction SP$\xi$. In particular, we expect error bounds similar to
(\ref{thULoss}). Making this rigorous seems difficult. Some
general remarks have been made in the last section. Note that
bounds like (\ref{cximu}) cannot hold in general, but could be
valid for AI$\xi$ in (pseudo)passive environments.

\label{defKstar} Here we concentrate on the special case of a
deterministic computable environment, i.e.\ the environment is a
sequence $\hh z = \hh z_1\hh z_2...$ with $K(\hh z_{1:\infty}) <
\infty$. Furthermore, we only consider the simplest horizon model
$m_k = k$, i.e.\ greedily maximize only the next reward. This is sufficient
for sequence prediction, as the reward of cycle $k$ only depends
on output $y_k$ and not on earlier decisions. This choice is in no
way sufficient and satisfactory for the full AI$\xi$ model, as
{\it one} single choice of $m_k$ should serve for {\it all} AI
problem classes. So AI$\xi$ should allow good sequence prediction
for some universal choice of $m_k$ and not only for $m_k = k$,
which definitely does not suffice for more complicated AI
problems. The analysis of this general case is a challenge for the
future.
For $m_k = k$ the AI$\xi$ model (\ref{ydotxi}) with
$o_i=\epstr$ and $r_k\in\{0,1\}$ reduces to
\beq\label{ydotxisp}
  \hh y_k \;=\; \arg\max_{y_k}\sum_{r_k}r_k\!\cdot\!
  \xi(\hh y\!\hh r_{<k}y\!\pb r_k) \;=\;
  \arg\max_{y_k}\xi(\hh y\!\hh r_{<k}y_k\pb 1)
\eeq
The environmental response $\hh r_k$ is given by $\delta_{\hh
y_k\hh z_k}$; it is 1 for a correct prediction $(\hh y_k = \hh
z_k)$ and 0 otherwise. One can show \cite{Hutter:00kcunai,Hutter:04uaibook} that
the number of wrong predictions $E_\infty^{\AI\xi}$ of the AI$\xi$
model (\ref{ydotxisp}) in these environments is bounded by
\beq\label{Ebndsp}
  E_\infty^{\AI\xi} \;\leqm \;
  2^{K(\hh z_{1:\infty})} \;<\; \infty
\eeq
for a computable deterministic environment string $\hh z_1\hh
z_2...$. The intuitive interpretation is that each wrong
prediction eliminates at least one program $p$ of size
$\l(p) \leqa  K(\hh z)$. The size is smaller than $K(\hh z)$, as
larger policies could not mislead the agent to a wrong
prediction, since there is a program of size $K(\hh z)$ making a correct
prediction. There are at most $2^{K(\hh z)+O(1)}$ such policies,
which bounds the total number of errors.

We have derived a finite bound for $E_\infty^{\AI\xi}$, but
unfortunately, a rather weak one as compared to (\ref{thULoss}).
The reason for the strong bound in the SP case was that every
error eliminates half of the programs.

The AI$\xi$ model would not be sufficient for realistic
applications if the bound (\ref{Ebndsp}) were sharp, but we have
the strong feeling (but only weak arguments) that better
bounds proportional to $K(\hh z)$ analogous to (\ref{thULoss})
exist. The current proof technique is not strong enough for
achieving this. One argument for a better bound is the formal
similarity between $\arg\max_{z_k}\xi(\hh z_{<k}\pb z_k)$ and
(\ref{ydotxisp}), the other is that we were unable to construct an
example sequence for which AI$\xi$ makes more than
$O(K(\hh z))$ errors.

\subsection{Strategic Games (SG)}\label{secSG}

\paragraph{Introduction}
A very important class of problems are strategic games (SG). Game
theory considers simple games of chance like roulette, combined
with strategy like backgammon, up to purely strategic games like
\idx{chess} or checkers or go. In fact, what is subsumed under
\idx{game theory} is so general that it includes not only a huge
variety of game types, but can also describe political and
economic competitions and coalitions, Darwinism and many more
topics. It seems that nearly every AI problem could be brought
into the form of a game. Nevertheless, the intention of a game is
that several players perform actions with (partial) observable
consequences. The goal of each player is to maximize some utility
function (e.g.\ to win the game). The players are assumed to be
rational, taking into account all information they posses. The
different goals of the players are usually in conflict. For an
introduction into game theory, see
\cite{Fudenberg:91,Osborne:94,Russell:03,VonNeumann:44}.

If we interpret the AI system as one player and the environment
models the other rational player {\it and} the environment provides
the reinforcement feedback $r_k$, we see that the agent-environment
configuration satisfies all criteria of a game. On the other hand,
the AI models can handle more general situations,
since they interact optimally with an environment, even if the environment
is not a rational player with conflicting goals.

\paragraph{Strictly competitive strategic games}
In the following, we restrict ourselves to deterministic, strictly
competitive strategic\footnote{In game theory, games like chess
are often called `extensive', whereas `strategic' is reserved for
a different kind of game.} games with alternating moves. Player 1
makes move $y_k$ in round $k$, followed by the move $o_k$ of
player 2.$\!$\footnote{We anticipate notationally the later
identification of the moves of player 1/2 with the
actions/observations in the AI models.} So a game with $n$ rounds
consists of a sequence of alternating moves
$y_1o_1y_2o_2...y_no_n$. At the end of the game in cycle $n$ the
game or final board situation is evaluated with
$V(y_1o_1...y_no_n)$. Player 1 tries to maximize $V$, whereas
player 2 tries to minimize $V$. In the simplest case, $V$ is $1$
if player 1 won the game, $V = -1$ if player 2 won and $V = 0$ for
a draw. We assume a fixed game length $n$ independent of the
actual move sequence. For games with variable length but maximal
possible number of moves $n$, we could add dummy moves and pad the
length to $n$. The optimal strategy (Nash equilibrium) of both
players is a minimax strategy
\bqa\label{sgxdot}
  \hh o_k &=& \arg\min_{o_k}\max_{y_{k+1}}\min_{o_{k+1}}...\max_{y_n}\min_{o_n}
  V(\hh y_1\hh o_1...\hh y_ko_k...y_no_n),
\\ \label{sgydot}
  \hh y_k &=& \arg\max_{y_k}\min_{o_k}...\max_{y_n}\min_{o_n}
  V(\hh y_1\hh o_1...\hh y_{k-1}\hh o_{k-1}y_ko_k...y_no_n). \qquad
\eqa
But note that the minimax strategy is only optimal if both players
behave rationally. If, for instance, player 2 has limited capabilites or makes
errors and player 1 is able to discover these (through past moves), he
could exploit these weaknesses and improve his performance
by deviating from the minimax strategy. At least the classical
game theory of Nash equilibria does not take into account limited
rationality, whereas the AI$\xi$ agent should.

\paragraph{Using the AI$\mu$ model for game playing}
In the following, we demonstrate the applicability of the AI model
to games. The AI$\mu$ model takes the position of player 1. The
environment provides the evaluation $V$. For a symmetric situation
we could take a second AI$\mu$ model as player 2, but for simplicity we
take the environment as the second player and assume that this
environmental player behaves according to the minimax strategy (\ref{sgxdot}).
The environment serves as a perfect player {\it and} as a teacher, albeit a
very crude one, as it tells the agent at the end of the game
only whether it won or lost.

The minimax behavior of player 2 can be expressed by a
(deterministic) probability distribution $\mu^\SG$ as the
following:
\beq\label{defmusg}
  \mu^\SG(y_1\pb o_1...y_n\pb o_n) \;:=\;
  \left\{
  \begin{array}{l}
    \displaystyle
    1 \qmbox{if}
    o_k=\arg\min_{o'_k}...\max_{y'_n}\min_{o'_n}
    V(y_1o_1...y_ko'_k...y'_no'_n)
    \;\;\forall\; k 
    \\
    0 \quad\mbox{otherwise}
  \end{array} \right.
\eeq
The probability that player 2 makes move $o_k$ is
$\mu^\SG(\hh y_1 \hh o_1...\hh y_k\pb o_k)$, which is 1 for
$o_k = \hh o_k$ as defined in (\ref{sgxdot}) and 0 otherwise.

Clearly, the AI$\mu$ system receives no feedback, i.e.\ $r_1 =...=
r_{n-1} = 0$, until the end of the game, where it should receive
positive/negative/neutral feedback on a win/loss/draw, i.e.\
$r_n=V(...)$. The environmental prior probability is therefore
\beq\label{muaisg}
  \mu^\AI(y_1\pb x_1...y_n\pb x_n) \;=\;
  \left\{
  \begin{array}{cl}
    \displaystyle
    \mu^\SG(y_1\pb o_1...y_n\pb o_n) & \mbox{if}\quad
    r_1...r_{n-1}\!=\!0 \;\mbox{and}\; r_n=V(y_1o_1...y_no_n)
    \\
    0 & \mbox{otherwise}
  \end{array} \right.
\eeq
where $x_i = r_io_i$. If the environment is a
minimax player (\ref{sgxdot}) plus a crude teacher $V$, i.e.\ if
$\mu^\AI$ is the true prior probability, the question now is,
what is the behavior $\hh y_k^\AI$ of the AI$\mu$ agent. It
turns out that if we set $m_k = n$ the AI$\mu$ agent is also a
minimax player (\ref{sgydot}) and hence optimal ($\hh y_k^\AI=\hh
y_k^\SG$, see \cite{Hutter:00kcunai,Hutter:04uaibook} for a formal proof). Playing
a sequence of games is a special case of a factorizable $\mu$
described in Section~\ref{secFacmu} with identical factors $\mu_r$
for all $r$ and equal episode lengths $n_{r+1} - n_r = n$.

Hence, in a minimax environment AI$\mu$ behaves itself as a
minimax strategy,
\beq\label{yaisgrep}
  \hh y_k^\AI \;=\;
  \arg\max_{y_k}\min_{o_k}...
     \max_{y_{(r+1)n}}\min_{\;o_{(r+1)n}}
     V(\hh y\!\hh o_{rn+1:k-1}...y\!o_{k:(r+1)n})
\eeq
with $r$ such that $rn < k \leq (r + 1)n$ and for any choice of
$m_k$ as long as the horizon $h_k \geq n$.

\paragraph{Using the AI$\xi$ Model for Game Playing}
\index{game playing!with AIXI}%
When going from the specific AI$\mu$ model, where the rules of the
game are explicitly modeled into the prior probability $\mu^\AI$,
to the universal model AI$\xi$, we have to ask whether these rules
can be learned from the assigned rewards $r_k$. Here, the main
reason for studying the case of repeated games rather than just
one game arises. For a single game there is only one cycle of
nontrivial feedback, namely the end of the game, which is too late
to be useful except when further games follow.

We expect that no other learning scheme (with no extra
information) can learn the game more quickly than AI$\xi$, since
$\mu^\AI$ factorizes in the case of
games of fixed length, i.e.\ $\mu^\AI$ satisfies a strong
separability condition. In the case of variable game length the
entanglement is also low. $\mu^\AI$ should still be sufficiently
separable, allowing us to formulate and prove good reward bounds for
AI$\xi$. A qualitative argument goes as follows:

Since initially, AI$\xi$ loses all games, it tries to draw out a
loss as long as possible, without having ever experienced or even
knowing what it means to win. Initially, AI$\xi$ will make a lot
of illegal moves. If illegal moves abort the game resulting in
(non-delayed) negative reward (loss), AI$\xi$ can quickly learn
the typically simple rules concerning legal moves, which usually
constitute most of the rules; just the goal rule is missing. After
having learned the move-rules, AI$\xi$ learns the (negatively
rewarded) losing positions, the positions leading to losing
positions, etc., so it can try to draw out losing games.
For instance, in chess, avoiding being check mated for 20,
30, 40 moves against a master is already quite an achievement. At
this ability stage, AI$\xi$ should be able to win some games by
luck, or speculate about a symmetry in the game that check mating
the opponent will be positively rewarded. Once having found out
the complete rules (moves and goal), AI$\xi$ will right away
reason that playing minimax is best, and henceforth beat all
grandmasters.

\index{feedback!more}%
If a (complex) game cannot be learned in this way in a realistic
number of cycles, one has to provide more feedback. This could be
achieved by intermediate help during the game. The environment
could give positive (negative) feedback for every good (bad) move
the agent makes. The demand on whether a move is to be valuated as
good should be adapted to the gained experience of the agent in
such a way that approximately the better half of the moves are
valuated as good and the other half as bad, in order to maximize
the information content of the feedback.

For more complicated games like \idx{chess}, even more feedback may
be necessary from a practical point of view. One way to increase the
feedback far beyond a few bits per cycle is to train the agent by
teaching it good moves. This is called supervised learning.
Despite the fact that the AI$\mu$ model has only a reward feedback
$r_k$, it is able to learn supervised, as will be shown in
Section~\ref{secEX}. Another way would be to start with more
simple games containing certain aspects of the true game and to
switch to the true game when the agent has learned the simple
game.

No other difficulties are expected when going from
$\mu$ to $\xi$. Eventually $\xi^\AI$ will converge to the
minimax strategy $\mu^\AI$. In the more realistic case, where the
environment is not a perfect minimax player, AI$\xi$ can
detect and exploit the weakness of the opponent.

Finally, we want to comment on the input/output space $\X$/$\Y$ of
the AI models. In practical applications, $\Y$ will possibly include
also illegal moves. If $\Y$ is the set of moves of, e.g.\ a robotic
arm, the agent could move a wrong figure or even knock over the
figures. A simple way to handle illegal moves $y_k$ is by
interpreting them as losing moves, which terminate the game.
Further, if, e.g.\ the input $x_k$ is the image of a video camera
which makes one shot per move, $\X$ is not the set of moves by the
environment but includes the set of states of the game board. The
discussion in this section handles this case as well. There is no
need to explicitly design the systems I/O space $\X/\Y$ for a
specific game.

The discussion above on the AI$\xi$ agent was rather informal for
the following reason: game playing (the SG$\xi$ agent) has
(nearly) the same complexity as fully general AI, and quantitative
results for the AI$\xi$ agent are difficult (but not impossible)
to obtain.

\subsection{Function Minimization (FM)}\label{secFM}

\paragraph{Applications/examples}
There are many problems that can be reduced to function minimization
(FM) problems. The minimum of a (real-valued) function
$f : \Y \to \SetR$ over some domain $\Y$ or a good approximate
to the minimum has to be found, usually with some limited resources.

One popular example is the traveling salesman problem (TSP). $\Y$
is the set of different routes between towns, and $f(y)$ the length
of route $y \in \Y$. The task is to find a route of minimal
length visiting all cities. This problem is NP hard. Getting good
approximations in limited time is of great importance in various
applications. 
Another example is the minimization of production costs (MPC),
e.g.\ of a car, under several constraints. $\Y$ is the set of all
alternative car designs and production methods compatible with the
specifications and $f(y)$ the overall cost of alternative
$y \in \Y$. 
A related example is finding materials or (bio)molecules with
certain properties (MAT), e.g.\ solids with minimal electrical
resistance or maximally efficient chlorophyll modifications, or
aromatic molecules that taste as close as possible to strawberry. 
We can also ask for nice paintings (NPT). $\Y$ is the set of all
existing or imaginable paintings, and $f(y)$ characterizes how much
person $A$ likes painting $y$. The agent should present
paintings which $A$ likes.

For now, these are enough examples. The TSP is very rigorous from a
mathematical point of view, as $f$, i.e.\ an algorithm of $f$, is
usually known. In principle, the minimum could be found by
exhaustive search, were it not for computational resource
limitations. For MPC, $f$ can often be modeled in a reliable and
sufficiently accurate way. For MAT you need very accurate physical
models, which might be unavailable or too difficult to solve or
implement. For NPT all we have is the judgement of person $A$ on
every presented painting. The evaluation function $f$ cannot be
implemented without scanning $A$'s brain, which is not possible with
today's technology.

So there are different limitations, some depending on the
application we have in mind. An implementation of $f$ might not be
available, $f$ can only be tested at some arguments $y$ and $f(y)$
is determined by the environment. We want to (approximately)
minimize $f$ with as few function calls as possible or, conversely,
find an as close as possible approximation for the
minimum within a fixed number of function evaluations. If $f$ is
available or can quickly be inferred by the agent and evaluation
is quick, it is more important to minimize the total time needed to
imagine new trial minimum candidates plus the evaluation time for
$f$. As we do not consider computational aspects of AI$\xi$ till
Section~\ref{secAIXItl} we concentrate on the first
case, where $f$ is not available or dominates the computational
requirements.

\paragraph{The greedy model}
The FM model consists of a sequence $\hh y_1\hh z_1\hh y_2\hh
z_2...$ where $\hh y_k$ is a trial of the FM agent for a minimum
of $f$ and $\hh z_k=f(\hh y_k)$ is the true function value
returned by the environment. We randomize the model by assuming a
probability distribution $\mu(f)$ over the functions. There are
several reasons for doing this. We might really not know the exact
function $f$, as in the NPT example, and model our uncertainty by
the probability distribution $\mu$. What is more important, we want to
parallel the other AI classes, like in the SP$\mu$ model, where we
always started with a probability distribution $\mu$ that was finally
replaced by $\xi$ to get the universal Solomonoff prediction
SP$\xi$. We want to do the same thing here. Further, the probabilistic
case includes the deterministic case by choosing
$\mu(f) = \delta_{ff_0}$, where $f_0$ is the true function. A
final reason is that the deterministic case is trivial when $\mu$
and hence $f_0$ are known, as the agent can internally (virtually)
check all function arguments and output the correct minimum from the very
beginning.

We assume that $\Y$ is countable and that $\mu$ is a
discrete measure, e.g.\ by taking only computable functions. The
probability that the function values of $y_1,...,y_n$ are
$z_1,...,z_n$ is then given by
\beq\label{fmmudef}
  \mu^\FM(y_1\pb z_1...y_n\pb z_n) \;:=\;
  \sum_{f:f(y_i)=z_i\;\forall 1\leq i\leq n} \nq\mu(f)
\eeq
We start with a model that minimizes the expectation
$z_k$ of the function value $f$ for the next output
$y_k$, taking into account previous information:
\beqn
  \hh y_k \;:=\; \arg\min_{y_k}\sum_{z_k} z_k\!\cdot\!
  \mu(\hh y_1\hh z_1...\hh y_{k-1}\hh z_{k-1}y_k\pb z_k)
\eeqn
This type of greedy algorithm, just minimizing the next feedback,
was sufficient for sequence prediction (SP) and is also sufficient
for classification (CF, not described here). It is, however, not
sufficient for function minimization as the following example
demonstrates.

Take $f:\{0,1\} \to \{1,2,3,4\}$. There are 16 different
functions which shall be equiprobable, $\mu(f) = {1\over 16}$.
The function expectation in the first cycle
\beqn
  \langle z_1\rangle \;:=\; \sum_{z_1} z_1\!\cdot\!\mu(y_1\pb z_1) \;=\;
  {\textstyle{1\over 4}}\sum_{z_1}z_1 \;=\;
  {\textstyle{1\over 4}}(1\!+\!2\!+\!3\!+\!4) \;=\; 2.5
\eeqn
is just the arithmetic average of the possible function values and
is independent of $y_1$. Therefore, $\hh y_1 = 0$, if we define
$\arg\min$ to take the lexicographically first minimum in an
ambiguous case like here. Let us assume that $f_0(0) = 2$, where
$f_0$ is the true environment function, i.e.\ $\hh z_1 = 2$. The
expectation of $z_2$ is then
\beqn
  \langle z_2\rangle \;:=\; \sum_{z_2} z_2\!\cdot\!\mu(02y_2\pb z_2)
  \;=\; \left\{
  \begin{array}{c@{\qmbox{for}}l}
    2                      & y_2=0 \\
    2.5                    & y_2=1
  \end{array} \right.
\eeqn
For $y_2 = 0$ the agent already knows $f(0) = 2$, for
$y_2 = 1$ the expectation is, again, the arithmetic average. The
agent will again output $\hh y_2 = 0$ with feedback $\hh
z_2 = 2$. This will continue forever. The agent is not
motivated to explore other $y$'s as $f(0)$ is already smaller than the
expectation of $f(1)$. This is obviously not what we
want. The greedy model fails. The agent ought to be inventive and
try other outputs when given enough time.

The general reason for the failure of the greedy approach is that
the information contained in the feedback $z_k$ depends on the
output $y_k$. A FM agent can actively influence the knowledge it
receives from the environment by the choice in $y_k$. It may be
more advantageous to first collect certain knowledge about $f$ by
an (in greedy sense) nonoptimal choice for $y_k$, rather than to
minimize the $z_k$ expectation immediately. The nonminimality of
$z_k$ might be overcompensated in the long run by exploiting this
knowledge. In SP, the received information is always the current
bit of the sequence, independent of what SP predicts for this bit.
This is why a greedy strategy in the SP case is already optimal.

\paragraph{The general FM$\mu/\xi$ model}\label{secGFMM}
To get a useful model we have to think more carefully about what
we really want. Should the FM agent output a good minimum in the
last output in a limited number of cycles $m$, or should the
average of the $z_1,...,z_m$ values be minimal, or does it suffice
that just one of the $z$ is as small as possible? The subtle and
important differences between these settings have been analyzed
and discussed in detail in \cite{Hutter:00kcunai,Hutter:04uaibook}. In the
following we concentrate on minimizing the average, or
equivalently the sum of function values. We define the FM$\mu$
model as to minimize the sum $z_1 +...+z_m$.
Building the $\mu$ average by summation over
the $z_i$ and minimizing w.r.t.\ the $y_i$ has to be performed in
the correct chronological order. With a similar reasoning as in
(\ref{ebesty}) to (\ref{ydotrec}) we get
\beq\label{fmydot}
  \hh y_k^\FM \;=\; \arg\min_{y_k}\sum_{z_k}...\min_{y_m}\sum_{z_m}
  (z_1\!+...+\!z_m)\!\cdot\!
  \mu(\hh y_1\hh z_1...\hh y_{k-1}\hh z_{k-1}y_k\pb z_k...y_m\pb z_m)
\eeq
By construction, the FM$\mu$ model guarantees optimal results in
the usual sense that no other model knowing only $\mu$
can be expected to produce better results.
The interesting case (in AI) is when $\mu$ is unknown. We
define for this case, the FM$\xi$ model by replacing $\mu(f)$
with some $\xi(f)$, which should assign high probability to
functions $f$ of low complexity. So we might define
$\xi(f) = \sum_{q:\forall x[U(qx)=f(x)]}2^{-\l(q)}$.
The problem with this definition is that it is, in general,
undecidable whether a TM $q$ is an implementation of a function
$f$. $\xi(f)$ defined in this way is uncomputable,
not even approximable. As we only need a $\xi$ analogous to the
l.h.s.\ of (\ref{fmmudef}), the following definition is natural
\beq\label{fmxidef}
  \xi^\FM(y_1\pb z_1...y_n\pb z_n) \;:=\;
  \sum_{q:q(y_i)=z_i\;\forall 1\leq i\leq n} \nq 2^{-\l(q)}
\eeq
$\xi^\FM$ is actually equivalent to inserting the uncomputable
$\xi(f)$ into (\ref{fmmudef}). One can show that $\xi^\FM$ is an
enumerable semimeasure and dominates all enumerable probability
distributions of the form (\ref{fmmudef}).

Alternatively, we could have constrained the sum in (\ref{fmxidef})
by $q(y_1...y_n) = z_1...z_n$ analogous to (\ref{uniMAI}), but these
two definitions are not equivalent. Definition (\ref{fmxidef})
ensures the symmetry\footnote{See \cite{Solomonoff:99} for a discussion
on symmetric universal distributions on unordered data.} in its
arguments and $\xi^\FM(...y\pb z...y\pb z'...) = 0$ for $z\neq z'$.
It incorporates all general knowledge we have about function
minimization, whereas (\ref{uniMAI}) does not. But this extra
knowledge has only low information content (complexity of $O(1)$),
so we do not expect FM$\xi$ to perform much worse when using
(\ref{uniMAI}) instead of (\ref{fmxidef}). But there is no reason
to deviate from (\ref{fmxidef}) at this point.

We can now define a loss $L_m^{\FM\mu}$ as (\ref{fmydot}) with $k
= 1$ and $\arg\min_{y_1}$ replaced by $\min_{y_1}$ and,
additionally, $\mu$ replaced by $\xi$ for $L_m^{\FM\xi}$. We
expect $|L_m^{\FM\xi} - L_m^{\FM\mu}|$ to be bounded in a way that
justifies the use of $\xi$ instead of $\mu$ for computable $\mu$,
i.e.\ computable $f_0$ in the deterministic case. The arguments
are the same as for the AI$\xi$ model.

In \cite{Hutter:00kcunai,Hutter:04uaibook} it has been proven that FM$\xi$ is
inventive in the sense that it never ceases searching for minima,
but will test {\em all} $y\in\Y$ if $\Y$ is finite (and an
infinite set of different $y$'s if $\Y$ is infinite) for
sufficiently large horizon $m$. There are currently no rigorous
results on the {\em quality} of the guesses, but for the FM$\mu$
agent the guesses are optimal by definition. If $K(\mu)$ for the
true distribution $\mu$ is finite, we expect the FM$\xi$ agent to
solve the `exploration versus exploitation' problem in a
universally optimal way, as $\xi$ converges rapidly to $\mu$.

\paragraph{Using the AI Models for Function Mininimization}\label{secUAIFM}
The AI models can be used for function minimization in the
following way. The output $y_k$ of cycle $k$ is a guess for a
minimum of $f$, like in the FM model. The reward $r_k$ should be
high for small function values $z_k = f(y_k)$. The choice $r_k = -
z_k$ for the reward is natural. Here, the feedback is not binary
but $r_k \in \R \subset \SetR$, with $\R$ being a countable
subset of $\SetR$, e.g.\ the computable reals or all rational
numbers. The feedback $o_k$ should be the function value
$f(y_k)$. As this is already provided in the rewards $r_k$ we
could set $o_k=\epstr$ as in Section~\ref{secSP}. For a change and
to see that the choice really does not matter we set $o_k = z_k$
here. The AI$\mu$ prior
probability is
\beq\label{muAIfm}
  \mu^\AI(y_1\pb x_1...y_n\pb x_n)
  \;=\; \left\{
  \begin{array}{cl}
    \mu^\FM(y_1\pb z_1...y_n\pb z_n)
    & \mbox{for } r_k=-z_k,\; o_k=z_k,\; x_k=r_k o_k \\
    0 & \mbox{else}.
  \end{array} \right.
\eeq
Inserting this into (\ref{pbestrec}) with $m_k=m$ one can show
that $\hh y_k^\AI=\hh y_k^\FM$, where $\hh y_k^\FM$ has been
defined in (\ref{fmydot}). The proof is very simple since the FM
model has already a rather general structure, which is similar to
the full AI model.

We expect no problem in going from FM$\xi$ to AI$\xi$. The only
thing the AI$\xi$ model has to learn, is to ignore the $o$
feedbacks as all information is already contained in $r$. This
task is simple as every cycle provides one data point for a simple
function to learn.

\paragraph{Remark on TSP}
The Traveling Salesman Problem (TSP) seems to be trivial in the
AI$\mu$ model but nontrivial in the AI$\xi$ model, because
(\ref{fmydot}) just implements an internal complete
search, as $\mu(f) = \delta_{ff^{\text{\tiny TSP}}}$ contains all
necessary information. AI$\mu$ outputs, from the very beginning,
the exact minimum of $f^{\text{TSP}}$. This ``solution'' is, of
course, unacceptable from a performance perspective. As long as we
give no efficient approximation $\xi^c$ of $\xi$, we have not
contributed anything to a solution of the TSP by using AI$\xi^c$.
The same is true for any other problem where $f$ is computable and
easily accessible. Therefore, TSP is not (yet) a good example
because all we have done is to replace an NP complete problem with
the uncomputable AI$\xi$ model or by a computable AI$\xi^c$ model,
for which we have said nothing about computation time yet. It is
simply an overkill to reduce simple problems to AI$\xi$. TSP is a
simple problem in this respect, until we consider the AI$\xi^c$
model seriously. For the other examples, where $f$ is inaccessible
or complicated, an AI$\xi^c$ model would provide a true solution
to the minimization problem as an explicit definition of $f$ is
not needed for AI$\xi$ and AI$\xi^c$. A computable version of
AI$\xi$ will be defined in Section~\ref{secAIXItl}.

\subsection{Supervised Learning from Examples (EX)}\label{secEX}

The developed AI models provide a frame for reinforcement
learning. The environment provides feedback $r$, informing the
agent about the quality of its last (or earlier) output $y$; it
assigns reward $r$ to output $y$. In this sense, reinforcement
learning is explicitly integrated into the AI$\mu/\xi$ models.
AI$\mu$ maximizes the true expected reward, whereas the AI$\xi$
model is a universal, environment-independent reinforcement
learning algorithm.

There is another type of learning method: Supervised learning by
presentation of examples (EX). Many problems learned by this
method are association problems of the following type. Given some
examples $o \in R\subset \O$, the agent should reconstruct, from a
partially given $o'$, the missing or corrupted parts, i.e.\
complete $o'$ to $o$ such that relation $R$ contains $o$. In many
cases, $\O$ consists of pairs $(z,v)$, where $v$ is the possibly
missing part.

\paragraph{Applications/examples}
Learning functions by presenting $(z,f(z))$ pairs and asking for
the function value of $z$ by presenting $(z,?)$ falls into the
category of supervised learning from examples, e.g.\ $f(z)$ may be
the class label or category of $z$.

A basic example is learning properties of geometrical objects
coded in some way. For instance, if there are 18 different objects
characterized by their size (small or big), their colors (red,
green, or blue) and their shapes (square, triangle, or circle), then
$(object,property) \in  R$ if the $object$ possesses the
$property$. Here, $R$ is a relation that is not the graph of a
single-valued function.

When teaching a child by pointing to objects and saying ``this is
a tree'' or ``look how green'' or ``how beautiful'', one
establishes a relation of $(object,property)$ pairs in $R$.
Pointing to a (possibly different) tree later and asking ``what is
this ?'' corresponds to a partially given pair $(object,?)$, where
the missing part ``?'' should be completed by the
child saying ``tree''.

A final example we want to give is chess. We have seen that, in
principle, chess can be learned by reinforcement learning. In the
extreme case the environment only provides reward $r = 1$ when
the agent wins. The learning rate is probably inacceptable from
a practical point of view, due to the low amount of
information feedback. A more practical method of teaching chess is
to present example games in the form of sensible
$(board\mbox{-}state,move)$
sequences. They contain information about legal and good moves
(but without any explanation). After several games have been presented, the
teacher could ask the agent to make its own move by presenting
$(board\mbox{-}state,?)$ and then evaluate the answer of the agent.

\paragraph{Supervised learning with the AI$\mu/\xi$ model}
Let us define the EX model as follows: The environment presents
inputs $o_{k-1} = z_kv_k \equiv (z_k,v_k) \in
R \cup ({\cal Z} \times \{?\}) \subset
 {\cal Z} \times (\Y \cup \{?\}) = \O$
to the agent in cycle $k - 1$. The agent is expected to output
$y_k$ in the next cycle, which is evaluated with $r_k = 1$ if
$(z_k,y_k) \in R$ and 0 otherwise. To simplify the discussion,
an output $y_k$ is expected and evaluated even when $v_k(\neq?)$
is given. To complete the description of the environment, the
probability distribution $\mu_R(\pb{o_1...o_n})$ of the examples
and questions $o_i$ (depending on $R$) has to be given. Wrong
examples should not occur, i.e.\ $\mu_R$ should be 0 if
$o_i \not\in R \cup ({\cal Z} \times \{?\})$ for some
$1 \leq i \leq n$. The relations $R$ might also be probability
distributed with $\sigma(\pb R)$. The example prior probability in
this case is
\beq\label{exmudef}
  \mu(\pb{o_1...o_n}) \;=\;
  \sum_R \mu_R(\pb{o_1...o_n}) \!\cdot\! \sigma(\pb R)
\eeq
The knowledge of the valuation $r_k$ on output $y_k$
restricts the possible relations $R$, consistent with
$R(z_k,y_k) = r_k$, where $R(z,y) := 1$ if $(z,y) \in R$ and 0
otherwise. The prior probability for the input sequence
$x_1...x_n$ if the output sequence of AI$\mu$ is $y_1...y_n$, is
therefore
\beqn
  \mu^\AI(y_1\pb x_1...y_n\pb x_n) \;=\;
  \sum_{R:\forall 1<i\leq n[R(z_i,y_i)=r_i]}
  \mu_R(\pb{o_1...o_n})\!\cdot\!\sigma(\pb R)
\eeqn
where $x_i = r_i o_i$ and $o_{i-1} = z_iv_i$ with $v_i \in \Y \cup \{?\}$.
In the I/O sequence $y_1x_1y_2x_2...=y_1r_1z_2v_2y_2r_2z_3v_3...$
the $y_1r_1$ are dummies, after that regular behavior starts
with example $(z_2,v_2)$.

The AI$\mu$ model is optimal by construction of $\mu^\AI$. For
computable prior $\mu_R$ and $\sigma$, we expect a near-optimal
behavior of the universal AI$\xi$ model if $\mu_R$ additionally
satisfies some separability property. In the following, we give
some motivation why the AI$\xi$ model takes into account the
supervisor information contained in the examples and why it learns
faster than by reinforcement.

We keep $R$ fixed and assume
$\mu_R(o_1...o_n) = \mu_R(o_1) \cdot...\cdot \mu_R(o_n) \neq 0
\Leftrightarrow o_i \in R \cup ({\cal Z} \times \{?\})\;\forall i$
to simplify the discussion. Short codes $q$ contribute most to
$\xi^\AI(y_1\pb x_1...y_n\pb x_n)$. As $o_1...o_n$ is
distributed according to the computable probability distribution
$\mu_R$, a short code of $o_1...o_n$ for large enough $n$ is a
Huffman code w.r.t.\ the distribution $\mu_R$. So we expect
$\mu_R$ and hence $R$ to be coded in the dominant contributions to
$\xi^\AI$ in some way, where the plausible assumption was made
that the $y$ on the input tape do not matter. Much more than one
bit per cycle will usually be learned, i.e.\ relation $R$ will be
learned in $n \ll K(R)$ cycles by appropriate examples. This
coding of $R$ in $q$ evolves independently of the feedbacks $r$.
To maximize the feedback $r_k$, the agent has to learn to output
a $y_k$ with $(z_k,y_k) \in R$. The agent has to invent
a program extension $q'$ to $q$, which extracts $z_k$ from
$o_{k-1} = (z_k,?)$ and searches for and outputs a $y_k$ with
$(z_k,y_k) \in R$. As $R$ is already coded in $q$, $q'$ can
reuse this coding of $R$ in $q$. The size of the extension $q'$
is, therefore, of order 1. To learn this $q'$, the agent requires
feedback $r$ with information content $O(1) = K(q')$ only.

Let us compare this with reinforcement learning, where only $o_{k-1} = (z_k,?)$
pairs are presented. A coding of $R$ in a short code $q$ for
$o_1...o_n$ is of no use and will therefore be absent. Only the
rewards $r$ force the agent to learn $R$. $q'$ is therefore
expected to be of size $K(R)$. The information content in the
$r$'s must be of the order $K(R)$. In practice, there are often only very few
$r_k = 1$ at the beginning of the learning phase, and the
information content in $r_1...r_n$ is much less than $n$ bits. The
required number of cycles to learn $R$ by reinforcement is,
therefore, at least but in many cases much larger than $K(R)$.

Although AI$\xi$ was never designed or told to learn
supervised, it learns how to take advantage of the examples from
the supervisor.  $\mu_R$ and $R$ are learned from the examples; the
rewards $r$ are not necessary for this process. The remaining task
of learning how to learn supervised is then a simple task of
complexity $O(1)$, for which the rewards $r$ are necessary.

\subsection{Other Aspects of Intelligence}\label{secOther}

In AI, a variety of general ideas and methods have been developed.
In the previous subsections, we saw how several problem
classes can be formulated within AI$\xi$. As we claim universality
of the AI$\xi$ model, we want to illuminate which of and how the
other AI methods are incorporated in the AI$\xi$ model by looking
at its structure. Some methods are directly included, while others are
or should be emergent. We do not claim the following list to be
complete.

{\it Probability theory} and {\it utility theory} are the heart of
the AI$\mu/\xi$ models. The probability $\xi$ is a universal
belief about the true environmental behavior $\mu$. The utility
function is the total expected reward, called value, which should
be maximized. Maximization of an expected utility function in a
probabilistic environment is usually called {\it sequential
decision theory}, and is explicitly integrated in full generality
in our model. In a sense this includes probabilistic (a
generalization of deterministic) {\it reasoning}, where the
objects of reasoning are not true and false statements, but the
prediction of the environmental behavior. {\it Reinforcement
Learning} is explicitly built in, due to the rewards. Supervised
learning is an emergent phenomenon (Section~\ref{secEX}). {\it
Algorithmic information theory} leads us to use $\xi$ as a
universal estimate for the prior probability $\mu$.

For horizon $> 1$, the expectimax series
in (\ref{pbestrec}) and the process of selecting maximal
values may be interpreted as abstract {\it planning}. The expectimax
series is a form of {\it informed search}, in the case of AI$\mu$, and {\it
heuristic search}, for AI$\xi$, where $\xi$ could be interpreted as
a heuristic for $\mu$. The minimax strategy of {\it game playing}
in case of AI$\mu$ is also subsumed. The AI$\xi$ model converges
to the minimax strategy if the environment is a minimax player, but
it can also take advantage of environmental players with limited
rationality. {\it Problem solving} occurs (only) in the form of
how to maximize the expected future reward.

{\it Knowledge} is accumulated by AI$\xi$ and is stored in some
form not specified further on the work tape. Any kind of
information in any representation on the inputs $y$ is
exploited. The problem of {\it knowledge engineering} and
{\it representation} appears in the form of how to train the AI$\xi$
model. More practical aspects, like {\it language or image
processing}, have to be learned by AI$\xi$ from scratch.

Other theories, like {\it fuzzy logic}, {\it possibility theory},
{\it Dempster-Shafer theory}, ... are partly outdated and partly
reducible to Bayesian probability theory
\cite{Cheeseman:85,Cheeseman:88}. The interpretation and
consequences of the evidence gap $g := 1 -
\sum_{x_k}\xi(y\!x_{<k}y\!\pb x_k) > 0$ in $\xi$ may be similar to
those in Dempster-Shafer theory. Boolean logical reasoning about
the external world plays, at best, an emergent role in the AI$\xi$
model.

Other methods that do not seem to be contained in the AI$\xi$ model
might also be emergent phenomena. The AI$\xi$ model has to
construct short codes of the environmental behavior, and
AIXI$tl$ (see next section) has to construct
short action programs. If we would analyze and interpret these
programs for realistic environments, we might find some of the
unmentioned or unused or new AI methods at work in these
programs. This is, however, pure speculation at this point. More
important: when trying to make AI$\xi$ practically usable,
some other AI methods, like genetic algorithms or neural nets,
especially for I/O pre/postprocessing, may be useful.

The main thing we wanted to point out is that the AI$\xi$ model
does not lack any important known property of intelligence or
known AI methodology. What {\it is} missing, however, are
computational aspects, which are addressed in the next section.

\section{Time-Bounded AIXI Model}\label{secAIXItl}\label{chTime}

Until now, we have not bothered with the non-computability of the
universal probability distribution $\xi$. As all universal models
in this paper are based on $\xi$, they are not effective in this
form. In this section, we outline how the previous models and
results can be modified/generalized to the time-bounded case.
Indeed, the situation is not as bad as it could be. $\xi$
is enumerable and $\hh y_k$ is still approximable, i.e.\
there exists an algorithm that will produce a
sequence of outputs eventually converging to the exact output $\hh
y_k$, but we can never be sure whether we have already reached it.
Besides this, the convergence is extremely slow, so this type of
asymptotic computability is of no direct (practical) use, but will
nevertheless be important later.

Let $\tilde p$ be a program that calculates within a reasonable
time $\tilde t$ per cycle, a reasonable intelligent output, i.e.\
$\tilde p(\hh x_{<k}) = \hh y_{1:k}$. This sort of computability
assumption, that a general-purpose computer of sufficient power is
able to behave in an intelligent way, is the very basis of AI,
justifying the hope to be able to construct agents that eventually
reach and outperform human intelligence. For a contrary viewpoint
see \cite{Lucas:61,Penrose:89,Penrose:94}. It is not necessary to
discuss here what is meant by `reasonable time/intelligence' and
`sufficient power'. What we are interested in, in this section, is
whether there is a computable version AIXI$\tilde t$ of the
AI$\xi$ agent that is superior or equal to any $p$ with
computation time per cycle of at most $\tilde t$. By `superior',
we mean `more intelligent', so what we need is an order relation
for intelligence, like the one in Definition \ref{defaiorder}.

The best result we could think of would be an AIXI$\tilde t$ with
computation time $\leq \tilde t$ at least as intelligent as any
$p$ with computation time $\leq \tilde t$. If AI is possible at
all, we would have reached the final goal: the construction of the
most intelligent algorithm with computation time $\leq \tilde t$.
Just as there is no universal measure in the set of computable
measures (within time $\tilde t$), neither may such an AIXI$\tilde
t$ exist.

What we can realistically hope to construct is an AIXI$\tilde t$
agent of computation time $c \cdot \tilde t$ per cycle for some
constant $c$. The idea is to run all programs $p$ of length $\leq
\tilde l := \l(\tilde p)$ and time $\leq \tilde t$ per cycle and
pick the best output. The total computation time is $c \cdot
\tilde t$ with $c=2^{\tilde l}$. This sort of idea of `\idx{typing
monkeys}' with one of them eventually writing Shakespeare, has
been applied in various forms and contexts in theoretical computer
science.
\index{vote!democratic}%
The realization of this {\it best vote} idea, in our
case, is not straightforward and will be outlined in this section.
A related idea is that of basing the decision on the
majority of algorithms. This `democratic vote' idea was used
in \cite{Littlestone:94,Vovk:92} for sequence prediction, and is
referred to as `weighted majority'.

\subsection{Time-Limited Probability Distributions}
\index{Kolmogorov complexity!time-limited}%
\index{semimeasure!universal, time-limited}%
\index{universal!time-limited semimeasure}%
In the literature one can find time-limited versions of Kolmogorov
complexity \cite{Daley:73,Daley:77,Ko:86} and the time-limited
universal semimeasure \cite{Li:91,Li:97,Schmidhuber:02speed}. In
the following, we utilize and adapt the latter and see how far we
get. One way to define a time-limited universal chronological
semimeasure is as a mixture over enumerable chronological
semimeasures computable within time $\tilde t$ and of size at most
$\tilde l$.
\beq\label{aixitl}
  \xi^{\tilde t\tilde l}(y\!\pb x_{1:n})
  \;:=\; \nq\sum_{\quad\rho\;:\;\l(\rho)\leq\tilde l\;\wedge\;t(\rho)\leq\tilde t}
  \nq\nq 2^{-\l(\rho)}\rho(y\!\pb x_{1:n})
\eeq
One can show that $\xi^{\tilde t\tilde l}$ reduces to $\xi^\AI$
defined in (\ref{uniMAI}) for $\tilde t, \tilde l \to\infty$. Let
us assume that the true environmental prior probability $\mu^\AI$ is
equal to or sufficiently accurately approximated by a $\rho$ with
$\l(\rho) \leq \tilde l$ and $t(\rho) \leq \tilde t$ with $\tilde
t$ and $\tilde l$ of reasonable size. There are several AI
problems that fall into this class. In function minimization of
Section~\ref{secFM}, the computation of $f$ and $\mu^\FM$ are
often feasible. In many cases, the sequences of Section~\ref{secSP} that should be predicted, can be easily calculated
when $\mu^\SP$ is known. In a classification problem, the probability
distribution $\mu^\CF$, according to which examples are
presented, is, in many cases, also elementary. But not all AI
problems are of this `easy' type. For the strategic games of
Section~\ref{secSG}, the environment itself is usually a highly
complex strategic player with a $\mu^\SG$ that is difficult to
calculate, although one might argue that the environmental player
may have limited capabilities too. But it is easy to think of a
difficult-to-calculate physical (probabilistic) environment like
the chemistry of biomolecules.

The number of interesting applications makes this restricted class
of AI problems, with time- and space-bounded environment
$\mu^{\tilde t\tilde l}$, worthy of study. Superscripts to a
probability distribution except for $\xi^{\tilde t\tilde l}$
indicate their length and maximal computation time. $\xi^{\tilde
t\tilde l}$ defined in (\ref{aixitl}), with a yet to be determined
computation time, multiplicatively dominates all $\mu^{\tilde
t\tilde l}$ of this type. Hence, an AI$\xi^{\tilde t\tilde l}$
model, where we use $\xi^{\tilde t\tilde l}$ as prior probability,
is universal, relative to all AI$\mu^{\tilde t\tilde l}$ models in
the same way as AI$\xi$ is universal to AI$\mu$ for all enumerable
chronological semimeasures $\mu$. The $\arg\max_{y_k}$ in
(\ref{ydotxi}) selects a $y_k$ for which $\xi^{\tilde t\tilde l}$
has the highest expected utility $V_{km_k}$, where $\xi^{\tilde
t\tilde l}$ is the weighted average over the $\rho^{\tilde t\tilde
l}$; i.e.\ output $\hh y_k^{\AI\xi^{\tilde t\tilde l}}$ is determined by a
weighted majority. We expect AI$\xi^{\tilde t\tilde l}$ to
outperform all (bounded) AI$\rho^{\tilde t\tilde l}$, analogous to the
unrestricted case.

In the following we analyze the computability properties of
$\xi^{\tilde t\tilde l}$ and AI$\xi^{\tilde t\tilde l}$, i.e.\ of
$\hh y_k^{\AI\xi^{\tilde t\tilde l}}$. To compute $\xi^{\tilde
t\tilde l}$ according to the definition (\ref{aixitl}) we have to
enumerate all chronological enumerable semimeasures $\rho^{\tilde
t\tilde l}$ of length $\leq \tilde l$ and computation time $\leq
\tilde t$. This can be done similarly to the unbounded case as
described in \cite{Li:97,Hutter:00kcunai,Hutter:04uaibook}. All $2^{\tilde l}$
enumerable functions of length $\leq \tilde l$, computable within
time $\tilde t$ have to be converted to chronological probability
distributions. For this, one has to evaluate each function for
$|\X| \cdot k$ different arguments. Hence, $\xi^{\tilde t\tilde
l}$ is computable within time\footnote{We assume that a (Turing) machine can be
simulated by another in linear time.}
$
  t(\xi^{\tilde t\tilde l}(y\!\pb x_{1:k}))  =
  O(|\X| \cdot k \cdot 2^{\tilde l} \cdot \tilde t)
$.
The computation time of $\hh y_k^{\AI\xi^{\tilde t\tilde l}}$
depends on the size of $\X$, $\Y$ and $m_k$.
$\xi^{\tilde t\tilde l}$ has to be
evaluated $|\Y|^{h_k}|\X|^{h_k}$ times in (\ref{ydotxi}).
It is possible to
optimize the algorithm and perform the computation within time
\beq\label{tyaixi}
  t(\hh y_k^{\AI\xi^{\tilde t\tilde l}}) \;=\;
  O(|\Y|^{h_k}|\X|^{h_k}\!\cdot\!2^{\tilde l}\!\cdot\!\tilde t)
\eeq
per cycle. If we assume that the computation time of $\mu^{\tilde
t\tilde l}$ is exactly $\tilde t$ for all arguments, the
brute-force time $\bar t$ for calculating the sums and maxs in
(\ref{ydotrec}) is $\bar t(\hh y_k^{\AI\mu^{\tilde t\tilde l}})
\geq |\Y|^{h_k}|\X|^{h_k} \cdot \tilde t$. Combining this with
(\ref{tyaixi}), we get
\beqn
  t(\hh y_k^{\AI\xi^{\tilde t\tilde l}}) \;=\;
  O(2^{\tilde l}\!\cdot\!
  \bar t(\hh y_k^{\AI\mu^{\tilde t\tilde l}}))
\eeqn
This result has the proposed structure, that there is a universal
AI$\xi^{\tilde t\tilde l}$ agent with computation time
$2^{\tilde l}$ times the computation time of a special
AI$\mu^{\tilde t\tilde l}$ agent.

Unfortunately, the class of AI$\mu^{\tilde t\tilde l}$ systems
with brute-force evaluation of $\hh y_k$ according to
(\ref{ydotrec}) is completely uninteresting from a practical point
of view. For instance, in the context of chess, the above result says that
the AI$\xi^{\tilde t\tilde l}$ is superior within time $2^{\tilde
l} \cdot \tilde t$ to any brute-force minimax strategy of computation time
$\tilde t$. Even if the factor of $2^{\tilde l}$ in computation
time would not matter, the AI$\xi^{\tilde t\tilde l}$ agent is,
nevertheless practically useless, as a brute-force minimax chess
player with reasonable time $\tilde t$ is a very poor player.

\index{algorithm!non-incremental}%
Note that in the case of binary sequence prediction ($h_k = 1$,
$|\Y| = |\X| = 2$) the computation time of $\rho$ coincides with
that of $\hh y_k^{\AI\rho}$ within a factor of 2. The class
AI$\rho^{\tilde t\tilde l}$ includes {\it all} non-incremental
sequence prediction algorithms of length $\leq \tilde l$ and
computation time $\leq \tilde t/2$. By non-incremental, we mean
that no information of previous cycles is taken into account for
speeding up the computation of $\hh y_k$ of the current cycle.

The shortcomings (mentioned and unmentioned ones) of this approach
are cured in the next subsection by deviating from the standard
way of defining a time-bounded $\xi$ as a sum over functions or
programs.

\subsection{The Idea of the Best Vote Algorithm}
\index{vote!best}\index{algorithm!best vote}
\indxs{algorithm}{incremental}%
A general agent is a chronological program $p(x_{<k})=y_{1:k}$.
This form, introduced in Section~\ref{secAIfunc}, is general
enough to include any AI system (and also less intelligent
systems). In the following, we are interested in programs $p$ of
length $\leq \tilde l$ and computation time $\leq \tilde t$ per
cycle. One important point in the time-limited setting is that $p$
should be incremental, i.e.\ when computing $y_k$ in cycle $k$,
the information of the previous cycles stored on the work tape can
be reused. Indeed, there is probably no practically interesting,
non-incremental AI system at all.

In the following, we construct a policy $p^\best$, or more
precisely, policies $p_k^\best$ for every cycle $k$ that
outperform all time- and length-limited AI systems $p$. In cycle $k$,
$p_k^\best$ runs all $2^{\tilde l}$ programs $p$ and selects the
one with the best output $y_k$. This is a `best vote' type of
algorithm, as compared to the `weighted majority' type algorithm of the
last subsection. The ideal measure for the quality of the output
would be the $\xi$-expected future reward
\beq
 V_{km}^{p\xi}(\hh y\!\hh x_{<k}) \;:=\; \sum_{q\in\hh Q_k}2^{-\l(q)}V_{km}^{pq}
 \quad,\quad
  V_{km}^{pq} \;:=\; r(x_k^{pq})+...+r(x_m^{pq})
\eeq
The program $p$ that maximizes $V_{km_k}^{p\xi}$ should be selected.
We have dropped the normalization $\cal N$ unlike in (\ref{cxi}),
as it is independent of $p$ and does not change the order relation
in which we are solely interested here. Furthermore, without
normalization, $V^{*\xi}_{km}(\hh y\!\hh x_{<k}):=\max_{p\in\hh
P}V_{km}^{p\xi}(\hh y\!\hh x_{<k})$ is enumerable, which will be
important later.

\subsection{Extended Chronological Programs}
\index{program!extended chronological}\index{policy!extended chronological}%
In the functional form of the AI$\xi$ model it
was convenient to maximize $V_{km_k}$ over all $p \in \hh P_k$,
i.e.\ all $p$ consistent with the current history $\hh y\!\hh
x_{<k}$. This was not a restriction, because for every possibly
inconsistent program $p$ there exists a program $p' \in \hh P_k$
consistent with the current history and identical to $p$ for all
future cycles $\geq k$. For the time-limited best vote algorithm
$p^\best$ it would be too restrictive to demand $p \in \hh P_k$.
To prove universality, one has to compare {\it all} $2^{\tilde l}$
algorithms in every cycle, not just the consistent ones. An
inconsistent algorithm may become the best one in later cycles.
For inconsistent programs we have to include the $\hh y_k$ into
the input, i.e.\ $p(\hh y\!\hh x_{<k}) = y_{1:k}^p$ with $\hh y_i
\neq y_i^p$ possible. For $p \in \hh P_k$ this was not necessary,
as $p$ knows the output $\hh y_k\equiv y_k^p$ in this case. The
$r_i^{pq}$ in the definition of $V_{km}$ are the rewards
emerging in the I/O sequence, starting with $\hh y\!\hh x_{<k}$
(emerging from $p^\best$) and then continued by applying $p$ and
$q$ with $\hh y_i := y_i^p$ for $i \geq k$.

Another problem is that we need $V_{km_k}$ to select the best
policy, but unfortunately $V_{km_k}$ is uncomputable. Indeed, the
structure of the definition of $V_{km_k}$ is very similar to that
of $\hh y_k$, hence a brute-force approach to approximate
$V_{km_k}$ requires too much computation time as for $\hh y_k$. We
solve this problem in a similar way, by supplementing each $p$ with
a program that estimates $V_{km_k}$ by $w_k^p$ within time
$\tilde t$. We combine the calculation of $y_k^p$ and $w_k^p$ and
extend the notion of a chronological program once again to
\beq\label{extprog}
  p(\hh y\!\hh x_{<k}) \;=\; w_1^py_1^p...w_k^py_k^p
\eeq
with chronological order $w_1^py_1^p\hh y_1\hh x_1
w_2^py_2^p\hh y_2\hh x_2...$.

\subsection{Valid Approximations}
\indxs{value}{valid approximation}
\index{approximation!value, valid}%
Policy $p$ might suggest any output $y_k^p$ but it is not allowed
to rate it with an arbitrarily high $w_k^p$ if we want $w_k^p$ to
be a reliable criterion for selecting the best $p$. We demand that
no policy is allowed to claim that it is better than it actually
is. We define a (logical) predicate VA($p$) called {\it valid
approximation}, which is true if and only if $p$ always satisfies
$w_k^p \leq V_{km_k}^{p\xi}$, i.e.\ never overrates itself.
\beq\label{vadef}
  \mbox{VA}(p) \;\equiv\;
  [\forall k\forall w_1^py_1^p\hh y_1\hh x_1...w_k^py_k^p :
  p(\hh y\!\hh x_{<k}) \!=\! w_1^py_1^p...w_k^py_k^p
  \Rightarrow
  w_k^p\leq V_{km_k}^{p\xi}(\hh y\!\hh x_{<k})]
\eeq
In the following, we restrict our attention to programs $p$, for
which VA($p$) can be proven in some formal axiomatic system. A very
important point is that $V^{*\xi}_{km_k}$ is enumerable. This
ensures the existence of sequences of programs $p_1, p_2, p_3, ...$
for which VA($p_i$) can be proven and $\lim_{i\to\infty}w_k^{p_i}
= V^{*\xi}_{km_k}$ for all $k$ and all I/O sequences. $p_i$ may be
defined as the naive (nonhalting) approximation scheme (by
enumeration) of $V^{*\xi}_{km_k}$ terminated after $i$ time steps
and using the approximation obtained so far for $w_k^{p_i}$ together
with the corresponding output $y_k^{p_i}$. The convergence
$w_k^{p_i}\toinfty{i} V^{*\xi}_{km_k}$ ensures that
$V^{*\xi}_{km_k}$, which we claimed to be the universally optimal
value, can be approximated by $p$ with provable VA$(p)$
arbitrarily well, when given enough time. The approximation is not
uniform in $k$, but this does not matter as the selected $p$ is
allowed to change from cycle to cycle.

Another possibility would be to consider only those $p$ that
check $w_k^p \leq V_{km_k}^{p\xi}$ online in every cycle, instead
of the pre-check VA($p$), either by constructing a proof (on the
work tape) for this special case, or $w_k^p \leq
V_{km_k}^{p\xi}$ is already evident by the construction of
$w_k^p$. In cases where $p$ cannot guarantee $w_k^p \leq
V_{km_k}^{p\xi}$ it sets $w_k = 0$ and, hence, trivially satisfies
$w_k^p \leq V_{km_k}^{p\xi}$. On the other hand, for these $p$ it
is also no problem to prove VA($p$) as one has simply to analyze
the internal structure of $p$ and recognize that $p$ shows the
validity internally itself, cycle by cycle, which is easy by
assumption on $p$. The cycle-by-cycle check is therefore a
special case of the pre-proof of VA($p$).

\subsection{Effective Intelligence Order Relation}
\index{intelligence!effective order}
\index{order relation!effective intelligence}%
In Section~\ref{secAIxi} we introduced an intelligence order
relation $\succeq$ on AI systems, based on the expected reward
$V_{km_k}^{p\xi}$. In the following we need an order relation
$\succeq^c$ based on the claimed reward $w_k^p$ which might
be interpreted as an approximation to $\succeq$.

\fdefinition{effaiord}{Effective intelligence order relation}{
We call $p$
{\it effectively more or equally intelligent} than $p'$ if
\bqan
  p\succeq^c\!p' &:\Leftrightarrow&
  \forall k\forall \hh y\!\hh x_{<k}
  \exists w_{1:n}w'_{1:n} :
\\
  & & p(\hh y\!\hh x_{<k}) \!=\! w_1\!*...w_k\!* \;\wedge\;
  p'(\hh y\!\hh x_{<k}) \!=\! w'_1\!*...w'_k\!* \;\wedge\;
  w_k\!\geq\!w'_k,
\eqan
i.e.\ if $p$ always claims higher reward estimate $w$ than $p'$.
}
Relation $\succeq^c$ is a co-enumerable partial order relation on
extended chronological programs. Restricted to valid
approximations it orders the policies w.r.t.\ the quality of their
outputs {\it and} their ability to justify their outputs with high
$w_k$.

\subsection{The Universal Time-Bounded AIXI$tl$ Agent}
\index{time bound!AIXI$tl$}%
In the following, we describe the algorithm $p^\best$ underlying
the universal time-bounded AIXI$\tilde t\tilde l$ agent. It
is essentially based on the selection of the best algorithms
$p_k^\best$ out of the time ${\tilde t}$ and length ${\tilde l}$
bounded $p$, for which there exists a proof of VA($p$) with length
$\leq l_P$.

\begin{enumerate}\parskip=0ex\parsep=0ex\itemsep=0ex
\item Create all binary strings of length $l_P$ and interpret each
as a coding of a mathematical proof in the same formal logic system in
which VA($\cdot$) was formulated. Take those strings
that are proofs of VA($p$) for some $p$ and keep the
corresponding programs $p$.
\item Eliminate all $p$ of length $> \tilde l$.
\item Modify the behavior of all retained $p$ in each cycle $k$ as follows:
Nothing is changed if $p$ outputs some $w_k^py_k^p$ within $\tilde
t$ time steps. Otherwise stop $p$ and write $w_k=0$ and some
arbitrary $y_k$ to the output tape of $p$. Let $P$ be the set of
all those modified programs.
\item Start first cycle: $k := 1$.
\item\label{pbestloop} Run every $p \in P$ on extended input
$\hh y\!\hh x_{<k}$, where all outputs are redirected to some
auxiliary tape: $p(\hh y\!\hh x_{<k}) = w_1^py_1^p...w_k^py_k^p$.
This step is performed incrementally by adding $\hh y\!\hh
x_{k-1}$ for $k > 1$ to the input tape and continuing the
computation of the previous cycle.
\item Select the program $p$ with highest claimed reward $w_k^p$:
$p_k^\best := \arg\max_pw_k^p$.
\item Write $\hh y_k := y_k^{p_k^\best}$ to the output tape.
\item Receive input $\hh x_k$ from the environment.
\item Begin next cycle: $k := k + 1$, goto step
\ref{pbestloop}.
\end{enumerate}

\noindent It is easy to see that the following theorem holds.

\indxs{optimality}{AIXI$tl $}
\ftheorem{thAIXItl}{Optimality of AIXI${tl}$}{
Let $p$ be any extended chronological (incremental) program like
(\ref{extprog}) of length $\l(p) \leq \tilde l$ and computation
time per cycle $t(p) \leq \tilde t$, for which there exists a
proof of VA($p$) defined in (\ref{vadef}) of length $\leq l_P$.
The algorithm $p^\best$ constructed in the last paragraph,
which depends on $\tilde l$, $\tilde t$ and $l_P$ but not on $p$, is
effectively more or equally intelligent, according to $\succeq^c$
(see Definition~\ref{effaiord}) than any such $p$. The size of
$p^\best$ is $\l(p^\best) = O(\log(\tilde l \cdot \tilde
t \cdot  l_P))$, the setup-time is
$t_{setup}(p^\best) = O(l_P^2 \cdot 2^{l_P})$ and the computation
time per cycle is $t_{cycle}(p^\best) = O(2^{\tilde
l} \cdot \tilde t)$.
}

\noindent Roughly speaking, the theorem says that if there exists
a computable solution to some or all AI problems at all, the
explicitly constructed algorithm $p^\best$ is such a solution.
Although this theorem is quite general, there are some limitations
and open questions that we discuss in the next subsection.

The construction of the algorithm $p^*$ needs the specification of
a formal logic system
$(\forall,\lambda,y_i,c_i,f_i,R_i,\rightarrow,\wedge,=,...)$, and
axioms, and inference rules. A proof is a sequence of formulas,
where each formula is either an axiom or inferred from previous
formulas in the sequence by applying the inference rules. Details
can be found in \cite{Hutter:01fast} in a related
construction or in 
any textbook on logic or proof theory, e.g.\
\cite{Fitting:96,Shoenfield:67}. We only need to know that {\em
provability} and {\em Turing Machines} can be formalized. The
setup time in the theorem is just the time needed to verify
the $2^{l_P}$ proofs, each needing time $O(l_P^2)$.

\subsection{Limitations and Open Questions}
\begin{itemize}\parskip=0ex\parsep=0ex
\item Formally, the total computation time of $p^\best$ for cycles
$1...k$ increases linearly with $k$, i.e.\ is of order $O(k)$ with
a coefficient $2^{\tilde l} \cdot \tilde t$. The unreasonably
large factor $2^{\tilde l}$ is a well-known drawback in
best/democratic vote models and will be taken without further
comments, whereas the factor ${\tilde t}$ can be assumed to be of
reasonable size. If we do not take the limit $k \to \infty$ but
consider reasonable $k$, the practical significance of the time bound
on $p^\best$ is somewhat limited due to the additional additive
constant $O(l_P^2 \cdot 2^{l_P})$. It is much larger than $k \cdot
2^{\tilde l} \cdot \tilde t$ as typically $l_P \gg \l($VA$(p))
\geq \l(p) \equiv \tilde l$.
\item\index{value!justification} $p^\best$ is superior only
to those $p$ that justify their
outputs (by large $w_k^p$). It might be possible that there are
$p$ that produce good outputs $y_k^p$ within reasonable time, but
it takes an unreasonably long time to justify their outputs by
sufficiently high $w_k^p$. We do not think that (from a certain
complexity level onwards) there are policies where the process of
constructing a good output is completely separated from some sort
of justification process. But this justification might not be
translatable (at least within reasonable time) into a reasonable
estimate of $V_{km_k}^{p\xi}$.
\item The (inconsistent) programs $p$ must be able to continue
strategies started by other policies. It might happen that a
policy $p$ steers the environment to a direction for which $p$ is
specialized. A ``foreign'' policy might be able to displace $p$
only between loosely connected episodes. There is probably no
problem for factorizable $\mu$. Think of a chess game, where it is
usually very difficult to continue the game or strategy of a
different player. When the game is over, it is usually advantageous
to replace a player by a better one for the next game. There might
also be no problem for sufficiently separable $\mu$.
\item There might be (efficient) valid approximations $p$ for which
VA($p$) is true but not provable, or for which only a very long
($> l_P$) proof exists.
\end{itemize}

\subsection{Remarks}
\begin{itemize}\parskip=0ex\parsep=0ex
\item The idea of suggesting outputs and justifying them by proving
reward bounds implements one aspect of human thinking. There are
several possible reactions to an input. Each reaction possibly has
far-reaching consequences. Within a limited time one tries to estimate the
consequences as well as possible. Finally,
each reaction is valuated, and the best one is selected. What
is inferior to human thinking is that the estimates $w_k^p$ must
be rigorously proved and the proofs are constructed by blind
exhaustive search, further, that {\it all} behaviors $p$ of length
$\leq \tilde l$ are checked. It is inferior ``only'' in the sense of
necessary computation time but not in the sense of the quality of
the outputs.
\item In practical applications there are often cases with
short and slow programs $p_s$ performing some task $T$, e.g.\
the computation of the digits of $\pi$, for which there exist
long but quick programs $p_l$ too. If it is not too difficult to
prove that this long program is equivalent to the short one, then it is
possible to prove $K^{t(p_l)}(T) \leqa \l(p_s)$
with $K^t$ being the time-bounded Kolmogorov complexity.
Similarly, the method of proving bounds $w_k$ for $V_{km_k}$ can
give high lower bounds without explicitly executing these short
and slow programs, which mainly contribute to $V_{km_k}$.
\item Dovetailing all length- and time-limited programs is a
well-known elementary idea (e.g.\ typing monkeys). The crucial part
that was developed here, is the selection criterion for the
most intelligent agent.
\item The construction of AIXI$\tilde t\tilde l$ and the
enumerability of $V_{km_k}$ ensure arbitrary close
approximations of $V_{km_k}$, hence we expect that the behavior of
AIXI$\tilde t\tilde l$ converges to the behavior of AI$\xi$
in the limit $\tilde t,\tilde l,l_P\to\infty$, in some sense.
\item Depending on what you know or assume that a program $p$ of size
$\tilde l$ and computation time per cycle $\tilde t$ is able to
achieve, the computable AIXI$\tilde t\tilde l$ model will have
the same capabilities. For the strongest assumption of the
existence of a Turing machine that outperforms human
intelligence, AIXI$\tilde t\tilde l$ will do too, within
the same time frame up to an (unfortunately very large) constant
factor.
\end{itemize}

\section{Discussion}\label{chDisc}

This section reviews what has been achieved in the article and
discusses some otherwise unmentioned topics of general interest.
We remark on various topics, including
concurrent actions and perceptions, the choice of the I/O spaces,
treatment of encrypted information, and peculiarities of mortal
embodies agents. We continue with an outlook on further research.
Since many ideas have already been presented in the various
sections, we concentrate on nontechnical open questions of
general importance, including optimality, down-scaling,
implementation, approximation, elegance, extra knowledge, and
training of/for AIXI($tl$). We also include some (personal)
remarks on non-computable physics, the number of wisdom $\Omega$,
and consciousness. As it should be, the article concludes with
conclusions.

\subsection{General Remarks}


\paragraph{Game theory}
\index{game theory}%
\index{theory|see{particular theory}}%
\index{concurrent!actions and perceptions}%
\index{actions!concurrent}%
\index{perceptions!concurrent}%
In game theory \cite{Osborne:94} one often wants to model the
situation of simultaneous actions, whereas the AI$\xi$ models have
serial I/O. Simultaneity can be simulated by withholding the
environment from the current agent's output $y_k$, until $x_k$ has
been received by the agent. Formally, this means that
$\mu(y\!x_{<k}y\!\pb x_k)$ is independent of the last output
$y_k$. The AI$\xi$ agent is already of simultaneous type in an
abstract view if the behavior $p$ is interpreted as the action.
In this sense, AIXI is the action $p^\best$ that maximizes the
utility function (reward), under the assumption that the
environment acts according to $\xi$. The situation is different
from game theory, as the environment $\xi$ is not a second
`player' that tries to optimize his own utility (see Section~\ref{secSG}).

\paragraph{Input/output spaces}
\index{input space!choice}%
\index{output space!choice}%
In various examples we have chosen differently specialized input
and output spaces $\X$ and $\Y$. It should be clear that, in
principle, this is unnecessary, as large enough spaces $\X$ and
$\Y$ (e.g.\ the set of strings of length $2^{32}$) serve every
need and can always be Turing-reduced to the specific presentation
needed internally by the AIXI agent itself. But it is clear that,
using a generic interface, such as camera and monitor for
learning tic-tac-toe, for example, adds the task of learning vision
and drawing.

\paragraph{How AIXI($tl$) deals with encrypted information}
\index{cryptography}
\index{decryption}
\indxs{RSA}{cryptography}
\indxs{encrypted}{information}%
Consider the task of decrypting a message that was encrypted by a
public key encrypter like RSA. A message $m$ is encrypted using a
product $n$ of two large primes $p_1$ and $p_2$, resulting in
encrypted message $c=$RSA$(m|n)$. RSA is a simple algorithm of
size $O(1)$. If AIXI is given the public key $n$ and encrypted
message $c$, in order to reconstruct the original message $m$ it
only has to ``learn'' the function
RSA$^{-1}(c|n):=\overline{\mbox{RSA}}(c|p_1,p_2)=m$. RSA$^{-1}$
can itself be described in length $O(1)$, since
$\overline{\mbox{RSA}}$ is $O(1)$ and $p_1$ and $p_2$ can be
reconstructed from $n$. Only very little information is needed to
learn $O(1)$ bits. In this sense decryption is easy for AIXI (like
TSP, see Section \ref{secFM}). The problem is that while
$\overline{\mbox{RSA}}$ is efficient, RSA$^{-1}$ is an extremely
slow algorithm, since it has to find the prime factors from the
public key.
But note, in AIXI we are not talking about computation time, we
are only talking about information efficiency (learning in the least
number of interaction cycles).
One of the key insights in this article that allowed for an
elegant theory of AI was this separation of data efficiency from
computation time efficiency. Of course, in the real world
computation time matters, so we invented AIXI$tl$.
AIXI$tl$ can do every job as well as the best length $l$ and time
$t$ bounded agent, apart from time factor $2^l$ and a huge offset
time. No practical offset time is sufficient to find the factors
of $n$, but in theory, enough offset time allows also AIXI$tl$ to
(once-and-for-all) find the factorization, and then, decryption is
easy of course.

\paragraph{Mortal embodied agents}
\indxs{mortal}{agents}%
\indxs{embodied}{agents}%
\index{agents!bodiless}%
\indxs{real}{environment}%
\indxs{autonomous}{robots}%
The examples we gave in this article, particularly those in
Section~\ref{chApply}, were mainly bodiless agents: predictors,
gamblers, optimizers, learners. There are some peculiarities with
reinforcement learning autonomous embodied robots in real
environments.

We can still reward the robot according to how well it solves the
task we want it to do. A minimal requirement is that the robot's
hardware functions properly. If the robot starts to malfunction
its capabilities degrade, resulting in lower reward. So, in an
attempt to maximize reward, the robot will also maintain itself.
The problem is that some parts will malfunction rather quickly
when no appropriate actions are performed, e.g.\ flat batteries,
if not recharged in time. Even worse, the robot may work perfectly
until the battery is nearly empty, and then suddenly stop its
operation (death), resulting in zero reward from then
on. There is too little time to learn how to maintain itself
before it's too late. An autonomous embodied robot cannot start
from scratch but must have some rudimentary built-in capabilities
(which may not be that rudimentary at all) that allow it to at
least survive.
\index{reflex}%
\index{behahiour!innate}%
\indxs{internal}{reward}%
\index{animals}%
\index{humans}%
Animals survive due to reflexes, innate behavior, an internal
reward attached to the condition of their organs, and a guarding
environment during childhood. Different species emphasize
different aspects. Reflexes and innate behaviors are stressed in
lower animals versus years of safe childhood for humans. The same
variety of solutions is available for constructing autonomous
robots (which we will not detail here).

\index{manipulation}%
Another problem connected, but possibly not limited to embodied
agents, especially if they are rewarded by humans, is the
following: Sufficiently intelligent agents may increase their
rewards by psychologically manipulating their human ``teachers'',
or by threatening them. This is a general sociological problem
which successful AI will cause, which has nothing specifically to
do with AIXI. Every intelligence superior to humans is capable of
manipulating the latter. In the absence of manipulable humans, e.g.\
where the reward structure serves a survival function, AIXI may
directly hack into its reward feedback. Since this is unlikely to
increase its long-term survival, AIXI will probably resist this
kind of manipulation (just as most humans don't take hard drugs,
due to their long-term catastrophic consequences).

\subsection{Outlook \& Open Questions}\label{secOutlook}

Many ideas for further studies were already stated in the
various sections of the article. This outlook only contains
nontechnical open questions regarding AIXI($tl$) of general
importance.

\paragraph{Value bounds}\indxs{value}{bounds}
Rigorous proofs for non-asymptotic value bounds for AI$\xi$ are
the major theoretical challenge -- general ones, as well as tighter
bounds for special environments $\mu$, e.g.\ for rapidly mixing
{\sc mdp}s, and/or other performance criteria have to be found
and proved. Although not necessary from a practical point of view,
the study of continuous classes $\M$, restricted policy classes,
and/or infinite $\Y$, $\X$ and $m$ may lead to useful insights.

\paragraph{Scaling AIXI down}\index{scaling!AIXI down}
A direct implementation of the AIXI$tl$ model is, at best,
possible for small-scale (toy) environments due to the large factor $2^l$ in
computation time. But there are other applications of the AIXI
theory. We saw in several examples how to integrate problem
classes into the AIXI model. Conversely, one can downscale the
AI$\xi$ model by using more restricted forms of $\xi$. This could be
done in the same way as the theory of universal induction was
downscaled with many insights to the Minimum Description Length
principle \cite{Li:92b,Rissanen:89} or to the domain of finite
automata \cite{Feder:92}. The AIXI model might similarly serve as
a supermodel or as the very definition of (universal unbiased)
intelligence, from which specialized models could be derived.

\paragraph{Implementation and approximation}
\indxs{implementation}{AIXI model}%
\indxs{approximation}{AIXI model}%
With a reasonable computation time, the AIXI model would be a
solution of AI (see the next point if you disagree). The AIXI$tl$
model was the first step, but the elimination of the factor $2^l$
without giving up universality will almost certainly be a very
difficult task.$\!$\footnote{But see \cite{Hutter:01fast} for an
elegant {\em theoretical} solution.} One could try to select
programs $p$ and prove VA($p$) in a more clever way than by mere
enumeration, to improve performance without destroying
universality. All kinds of ideas like \idx{genetic algorithms},
advanced theorem provers\index{theorem provers}\index{proof} and
many more could be incorporated. But now we have a problem.

\paragraph{Computability}\indxs{computability}{AIXI model}
We seem to have transferred the AI problem just to a different
level. This shift has some advantages (and also some
disadvantages) but does not present a practical solution. Nevertheless,
we want to stress that we have reduced the AI problem to (mere)
computational questions. Even the most general other systems the
author is aware of depend on some (more than complexity)
assumptions about the environment or it is far from clear whether
they are, indeed, universally optimal. Although computational
questions are themselves highly complicated, this reduction is a
nontrivial result. A formal theory of something, even if not
computable, is often a great step toward solving a problem and
also has merits of its own, and AI should not be different in this
respect (see previous item).

\paragraph{Elegance}
\index{artificial intelligence!elegant$\leftrightarrow$complex}%
Many researchers in AI believe that intelligence is something
complicated and cannot be condensed into a few formulas. It is
more a combining of enough {\it methods} and much explicit {\it
knowledge} in the right way. From a theoretical point of view we
disagree, as the AIXI model is simple and seems to serve all
needs. From a practical point of view we agree to the following
extent: To reduce the computational burden one should provide
special-purpose algorithms ({\it methods}) from the very
beginning, probably many of them related to reduce the complexity
of the input and output spaces $\X$ and $\Y$ by appropriate
pre/postprocessing {\it methods}.

\paragraph{Extra knowledge}\index{knowledge!incorporate}
There is no need to incorporate extra {\it knowledge} from the
very beginning. It can be presented in the first few cycles in
{\it any} format. As long as the algorithm to interpret the data
is of size $O(1)$, the AIXI agent will ``understand'' the data
after a few cycles (see Section~\ref{secEX}). If the environment
$\mu$ is complicated but extra knowledge $z$ makes $K(\mu|z)$
small, one can show that the bound (\ref{eukdistxi}) reduces
roughly to $\ln 2 \cdot K(\mu|z)$ when $x_1 \equiv z$, i.e.\ when
$z$ is presented in the first cycle. The special-purpose
algorithms could be presented in $x_1$ too, but it would be
cheating to say that no special-purpose algorithms were
implemented in AIXI. The boundary between implementation and
training is unsharp in the AIXI model.

\paragraph{Training}\indxs{training}{sequence}
We have not said much about the training process itself, as it is
not specific to the AIXI model and has been discussed in
literature in various forms and disciplines
\cite{Solomonoff:86,Schmidhuber:02bias,Schmidhuber:04oops}. By a
training process we mean a sequence of simple-to-complex tasks to
solve, with the simpler ones helping in learning the
more complex ones. A serious discussion would be out of place. To
repeat a truism, it is, of course, important to present enough
knowledge $o_k$ and evaluate the agent output $y_k$ with $r_k$ in
a reasonable way. To maximize the information content in the
reward, one should start with simple tasks and give positive
reward to approximately the better half of the outputs $y_k$.

\subsection{The Big Questions}

This subsection is devoted to the {\it big} questions of AI in
general and the AIXI model in particular with a personal touch.

\paragraph{On non-computable physics \& brains}
\indxs{physics}{non-computable}%
\indxs{brain}{non-computable}%
There are two possible objections to AI in general and, therefore,
to AIXI in particular.
Non-computable physics (which is not too weird) could make Turing
computable AI impossible. As at least the world that is relevant
for humans seems mainly to be computable we do not believe that it
is necessary to integrate non-computable devices into an AI
system.\index{G\"odel incompleteness} The (clever and
nearly convincing) G\"odel argument by Penrose
\cite{Penrose:89,Penrose:94}, refining Lucas
\cite{Lucas:61}, that non-computational physics {\it
must} exist and {\it is} relevant to the brain, has (in our
opinion convincing) loopholes.

\paragraph{Evolution \& the number of wisdom}
\index{evolution}\index{number of wisdom}%
A more serious problem is the evolutionary information-gathering
process. It has been shown that the `number of wisdom' $\Omega$
contains a very compact tabulation of $2^n$ undecidable problems
in its first $n$ binary digits \cite{Chaitin:91}. $\Omega$ is
only enumerable with computation time increasing more rapidly with
$n$ than any recursive function. The enormous computational power
of evolution could have developed and coded something like
$\Omega$ into our genes, which significantly guides human
reasoning. In short: Intelligence could be something complicated,
and evolution toward it from an even cleverly designed algorithm
of size $O(1)$ could be too slow. As evolution has already taken
place, we could add the information from our genes or brain
structure to any/our AI system, but this means that the important
part is still missing, and that it is principally impossible to
derive an efficient algorithm from a simple formal definition of
AI.

\paragraph{Consciousness}\index{consciousness}
\index{wave function collapse}%
\index{physics!wave function collapse}%
For what is probably the {\it biggest question}, that of {\it
consciousness}, we want to give a physical analogy. Quantum
(field) theory is the most accurate and universal physical theory
ever invented. Although already developed in the 1930s, the {\it
big} question, regarding the interpretation of the wave function
collapse, is still open. Although this is extremely interesting from a
philosophical point of view, it is completely irrelevant from a
practical point of view.$\!$\footnote{In the Theory of Everything, the
collapse might become of `practical' importance and must or will
be solved.} We believe the same to be valid for {\it
consciousness} in the field of Artificial Intelligence:
philosophically highly interesting but practically unimportant.
Whether consciousness {\it will} be explained some day is another
question.

\subsection{Conclusions}\label{secDiscCon}

The major theme of the article was to develop a mathematical
foundation of Artificial Intelligence. This is not an easy task
since intelligence has many (often ill-defined) faces.
More specifically, our goal was to develop a theory for rational
agents acting optimally in any environment. Thereby we touched
various scientific areas, including reinforcement learning,
algorithmic information theory, Kolmogorov complexity,
computational complexity theory, information theory and
statistics, Solomonoff induction, Levin search, sequential
decision theory, adaptive control theory, and many more.

We started with the observation that all tasks that require
intelligence to be solved can naturally be formulated as a
maximization of some expected utility in the framework of agents.
We presented a functional (\ref{pbestfunc}) and an iterative
(\ref{ydotrec}) formulation of such a decision-theoretic agent in
Section~\ref{chAImu}, which is general enough to cover all AI
problem classes, as was demonstrated by several examples. The
main remaining problem is the unknown prior probability
distribution $\mu$ of the environment(s). Conventional learning
algorithms are unsuitable, because they can neither handle large
(unstructured) state spaces, nor do they converge in the
theoretically minimal number of cycles, nor can they handle
non-stationary environments appropriately. On the other hand,
Solomonoff's universal prior $\xi$ (\ref{xidef}), rooted in
algorithmic information theory, solves the problem of
the unknown prior distribution for induction problems as was
demonstrated in Section~\ref{chSP}. No explicit learning procedure
is necessary, as $\xi$ automatically converges to $\mu$. We
unified the theory of universal sequence prediction with the
decision-theoretic agent by replacing the unknown true prior $\mu$
by an appropriately generalized universal semimeasure $\xi$ in
Section~\ref{chAIxi}. We gave various arguments that the resulting
AIXI model is the most intelligent, parameter-free and
environmental/application-independent model possible. We defined
an intelligence order relation (Definition \ref{defaiorder}) to give a
rigorous meaning to this claim. Furthermore, possible solutions to
the horizon problem have been discussed. In Section~\ref{chApply}
we outlined how the AIXI model solves various problem classes.
These included sequence prediction, strategic games, function
minimization and, especially, learning to learn supervised.
The list could easily be extended to other problem classes like
classification, function inversion and many others. The major
drawback of the AIXI model is that it is uncomputable, or more
precisely, only asymptotically computable, which makes an
implementation impossible. To overcome this problem, we
constructed a modified model AIXI$tl$, which is still effectively
more intelligent than any other time $t$ and length $l$ bounded
algorithm (Section~\ref{secAIXItl}). The computation time of
AIXI$tl$ is of the order $t\cdot 2^l$. A way of overcoming the
large multiplicative constant $2^l$  was presented in
\cite{Hutter:01fast} at the expense of an (unfortunately even
larger) additive constant. Possible further research was discussed. The main
directions could be to prove general and special reward bounds,
use AIXI as a supermodel and explore its relation to other
specialized models, and finally improve performance with or without
giving up universality.

All in all, the results show that Artificial Intelligence can be
framed by an elegant mathematical theory. Some progress has also
been made toward an elegant {\em computational} theory of
intelligence.

\addcontentsline{toc}{section}{Annotated Bibliography}
\section*{Annotated Bibliography}\label{secHR}

\paragraph{Introductory textbooks}
The book of Hopcroft and Ullman, and in the new revision co-authored
by Motwani \cite{Hopcroft:01}, is a very readable elementary
introduction to automata theory, formal languages, and computation
theory.
The Artificial Intelligence book \cite{Russell:03} by Russell and
Norvig gives a comprehensive overview over AI approaches in
general.
For an excellent introduction to Algorithmic Information Theory,
Kolmogorov complexity, and Solomonoff induction one should consult
the book of Li and Vit\'anyi \cite{Li:97}.
The Reinforcement Learning book by Sutton and Barto
\cite{Sutton:98} requires no background knowledge, describes the
key ideas, open problems, and great applications of this field. A
tougher and more rigorous book by Bertsekas and Tsitsiklis on
sequential decision theory provides all (convergence) proofs
\cite{Bertsekas:96}.

\paragraph{Algorithmic information theory}
Kolmogorov \cite{Kolmogorov:65} suggested to define the
information content of an object as the length of the shortest
program computing a representation of it. Solomonoff
\cite{Solomonoff:64} invented the closely related universal prior
probability distribution and used it for binary sequence
prediction \cite{Solomonoff:64,Solomonoff:78} and function
inversion and minimization \cite{Solomonoff:86}. Together with
Chaitin \cite{Chaitin:66,Chaitin:75}, this was the invention of
what is now called Algorithmic Information theory. For further
literature and many applications see \cite{Li:97}. Other
interesting applications can be found in
\cite{Chaitin:91,Schmidt:99,Vovk:98}. Related topics are the
Weighted Majority algorithm invented by Littlestone and Warmuth
\cite{Littlestone:94}, universal forecasting by Vovk
\cite{Vovk:92}, Levin search \cite{Levin:73search}, PAC-learning
introduced by Valiant \cite{Valiant:84} and Minimum Description
Length \cite{Li:92b,Rissanen:89}. Resource-bounded complexity is
discussed in \cite{Daley:73,Daley:77,Feder:92,Ko:86,Pintado:97},
resource-bounded universal probability in
\cite{Li:91,Li:97,Schmidhuber:02speed}. Implementations are rare
and mainly due to Schmidhuber
\cite{Conte:97,Schmidhuber:97nn,Schmidhuber:97bias,Schmidhuber:02bias,Schmidhuber:04oops}.
Excellent reviews with a philosophical touch are
\cite{Li:92a,Solomonoff:97}. For an older general review of
inductive inference see Angluin \cite{Angluin:83}.

\paragraph{Sequential decision theory}
The other ingredient in our AI$\xi$ model is sequential decision
theory. We do not need much more than the maximum expected utility
principle and the expectimax algorithm
\cite{Michie:66,Russell:03}. The book of von Neumann and
Morgenstern \cite{VonNeumann:44} might be seen as the initiation
of game theory, which already contains the expectimax algorithm as
a special case. The literature on reinforcement learning and
sequential decision theory is vast and we refer to the references given
in the textbooks \cite{Sutton:98,Bertsekas:96}.

\paragraph{The author's contributions}
Details on most of the issues addressed in this article can be
found in various reports or publications or the book
\cite{Hutter:04uaibook} by the author:
The AI$\xi$ model was first introduced and discussed in March
2000 in \cite{Hutter:00kcunai} in a 62-page-long report. More
succinct descriptions were published in
\cite{Hutter:01aixi,Hutter:01decision}. The AI$\xi$ model has been
argued to formally solve a number of problem classes, including
sequence prediction, strategic games, function minimization,
reinforcement and supervised learning \cite{Hutter:00kcunai}. A
variant of AI$\xi$ has recently been shown to be self-optimizing
and Pareto optimal \cite{Hutter:02selfopt}. The construction of a
general fastest algorithm for all well-defined problems
\cite{Hutter:01fast} arose from the construction of the
time-bounded AIXI$tl$ model \cite{Hutter:01aixi}. Convergence
\cite{Hutter:03unipriors} and tight \cite{Hutter:03optisp} error
\cite{Hutter:99errbnd,Hutter:01alpha} and loss
\cite{Hutter:01loss,Hutter:02spupper} bounds for Solomonoff's
universal sequence prediction scheme have been proven. Loosely
related ideas on a market/economy-based reinforcement learner
\cite{Hutter:01market} and gradient-based reinforcement planner
\cite{Hutter:01grep} were implemented.
These and other papers are available at
http://www.idsia.ch/$^{_{_\sim}}\!$marcus/ai.

\paragraph{Acknowledgements}
I am indebted to Shane Legg who proof-read this \ifagibook
chapter\else article\fi.


\addcontentsline{toc}{section}{References}
\def\bibindent{ABCD}\def\bibopening{\parskip=0ex}

\begin{small}

\end{small}


\addcontentsline{toc}{section}{Index}

\begin{footnotesize}
\begin{theindex}

  \item accessibility, 16
  \item action, 6
    \subitem random, 33
  \item actions
    \subitem concurrent, 55
  \item adaptive
    \subitem control, 24
  \item agent
    \subitem most intelligent, 24
  \item agents, 6
    \subitem bodiless, 56
    \subitem embodied, 56
    \subitem immortal, 32
    \subitem lazy, 32
    \subitem mortal, 56
  \item AI$\mu $ model
    \subitem equivalence, 13
    \subitem recursive \& iterative form, 12
    \subitem special aspects, 13
  \item AI$\xi $ model, 22
    \subitem axiomatic approach, 33
    \subitem general Bayes mixture, 25
    \subitem optimality, 24
    \subitem Pareto optimality, 29
    \subitem structure, 33
  \item AIXI model
    \subitem approximation, 57
    \subitem computability, 57
    \subitem implementation, 57
  \item AIXI$tl $
    \subitem optimality, 52
  \item algorithm
    \subitem best vote, 50
    \subitem incremental, 50
    \subitem non-incremental, 49
  \item alphabet, 7
  \item Angluin, D., 61
  \item animals, 56
  \item approximation
    \subitem AIXI model, 57
    \subitem value, valid, 51
  \item artificial intelligence
    \subitem elegant$\leftrightarrow$complex, 57
  \item asymmetry, 8
  \item asymptotic
    \subitem convergence, 24
    \subitem learnability, 28
  \item autonomous
    \subitem robots, 56
  \item average
    \subitem reward, 31
  \item axiomatic approach
    \subitem AI$\xi $ model, 33

  \indexspace

  \item Bandit problem, 25
  \item Barto, A.\nobreakspace  {}G., 6, 16, 60, 61, 64
  \item Bayes mixture
    \subitem general, 25
  \item behahiour
    \subitem innate, 56
  \item Bellman, R.\nobreakspace  {}E., 6, 16, 61
  \item Bertsekas, D.\nobreakspace  {}P., 6, 16, 31, 60, 61
  \item bias, 23
  \item boosting
    \subitem bound, 28
  \item bound
    \subitem boost, 28
    \subitem value, 26
  \item bounds
    \subitem value, 56
  \item brain
    \subitem non-computable, 58

  \indexspace

  \item chain rule, 11, 19
  \item Chaitin, G.\nobreakspace  {}J., 18, 58, 60, 61
  \item chaos, 15
  \item Cheeseman, P., 47, 62
  \item chess, 36, 39
  \item chronological, 7
    \subitem function, 8
    \subitem order, 11
    \subitem Turing machine, 8
  \item complete
    \subitem history, 16
  \item complexity
    \subitem input sequence, 15
    \subitem Kolmogorov, 17
  \item computability
    \subitem AIXI model, 57
  \item concept class
    \subitem restricted, 25
  \item concepts
    \subitem separability, 26
  \item concurrent
    \subitem actions and perceptions, 55
  \item consciousness, 59
  \item consistency, 24
  \item consistent
    \subitem policy, 23
  \item constants, 14
  \item Conte{ et al.}, M., 61, 62
  \item control
    \subitem adaptive, 24
  \item convergence
    \subitem asymptotic, 24
    \subitem finite, 24
    \subitem uniform, 29
  \item Cox, R.\nobreakspace  {}T., 19, 62
  \item cryptography, 55
    \subitem RSA, 55
  \item cybernetic systems, 6
  \item cycle, 7

  \indexspace

  \item Daley, R.\nobreakspace  {}P., 48, 60, 62
  \item Dawid, A.\nobreakspace  {}P., 17, 62
  \item decision
    \subitem suboptimal, 28
    \subitem wrong, 28
  \item decryption, 55
  \item degree of belief, 19
  \item deterministic, 6
    \subitem environment, 8
  \item differential
    \subitem gain, 31
  \item discounting
    \subitem harmonic, 31
    \subitem universal, 31
  \item dynamic
    \subitem horizon, 31

  \indexspace

  \item efficiency, 24
  \item embodied
    \subitem agents, 56
  \item encrypted
    \subitem information, 55
  \item environment
    \subitem deterministic, 8
    \subitem factorizable, 29
    \subitem farsighted, 29
    \subitem forgetful, 29
    \subitem inductive, 27
    \subitem Markov, 29
    \subitem passive, 27
    \subitem probabilistic, 9
    \subitem pseudo-passive, 26, 27
    \subitem real, 56
    \subitem stationary, 29
    \subitem uniform, 29
  \item environmental class
    \subitem limited, 25
  \item episode, 14
  \item evolution, 58
  \item expected
    \subitem utility, 16
  \item expectimax
    \subitem algorithm, 13
    \subitem tree, 13
  \item experiment, 19
  \item expert advice
    \subitem prediction, 32
  \item exploitation, 16
  \item exploration, 16

  \indexspace

  \item factorizable
    \subitem environment, 13, 29
  \item fair coin flips, 20
  \item farsighted
    \subitem environment, 29
  \item farsightedness
    \subitem dynamic, 10, 12
  \item Feder, M., 57, 60, 62
  \item feedback
    \subitem more, 39
    \subitem negative, 8
    \subitem positive, 8
  \item finite
    \subitem convergence, 24
  \item Fitting, M.\nobreakspace  {}C., 53, 62
  \item fixed
    \subitem horizon, 30
  \item forgetful
    \subitem environment, 29
  \item Fudenberg, D., 36, 62
  \item Fuentes, E., 60, 64
  \item functional form, 11

  \indexspace

  \item G\"odel incompleteness, 58
  \item gain
    \subitem differential, 31
  \item game playing
    \subitem with AIXI, 38
  \item game theory, 36, 55
  \item general
    \subitem Bayes mixture, 25
  \item general Bayes mixture
    \subitem AI$\xi $ model, 25
  \item generalization techniques, 16
  \item generalized universal prior, 22
  \item genetic algorithms, 57
  \item greedy, 10
  \item Gutman, M., 57, 60, 62
  \item G{\'a}cs, P., 18, 62

  \indexspace

  \item harmonic
    \subitem discounting, 31
  \item HeavenHell example, 26
  \item history, 9
    \subitem complete, 16
  \item Hopcroft, J.\nobreakspace  {}E., 60, 62
  \item horizon, 12
    \subitem choice, 15
    \subitem dynamic, 31
    \subitem fixed, 30
    \subitem infinite, 31
    \subitem problem, 30
  \item human, 14
  \item humans, 56
  \item Hutter, M., 4, 13, 18, 21, 23, 25, 30, 33, 35, 36, 38, 42, 43,
    49, 53, 57, 60--63

  \indexspace

  \item I/O sequence, 8
  \item image, 14
  \item immortal
    \subitem agents, 32
  \item imperfect, 15
  \item implementation
    \subitem AIXI model, 57
  \item inconsistent
    \subitem policy, 23
  \item incremental
    \subitem algorithm, 50
  \item independent
    \subitem episodes, 14
  \item inductive
    \subitem environment, 27
  \item infinite
    \subitem horizon, 31
  \item information
    \subitem encrypted, 55
  \item input, 6
    \subitem device, 15
    \subitem regular, 8
    \subitem reward, 8
    \subitem word, 7
  \item input space
    \subitem choice, 15, 55
  \item intelligence, 6
    \subitem effective order, 51
    \subitem intermediate, 24
    \subitem order relation, 23
  \item intermediate
    \subitem intelligence, 24
  \item internal
    \subitem reward, 56
  \item iterative formulation, 11

  \indexspace

  \item Kaelbling, L.\nobreakspace  {}P., 16, 63
  \item knowledge
    \subitem incorporate, 57
  \item Ko, K.-I., 48, 60, 63
  \item Kolmogorov complexity, 17, 18
    \subitem time-limited, 48
  \item Kolmogorov, A.\nobreakspace  {}N., 18, 60, 63
  \item Kumar, P.\nobreakspace  {}R., 24, 26, 63
  \item Kwee, I., 61, 63

  \indexspace

  \item lazy
    \subitem agents, 32
  \item learnable
    \subitem asymptotically, 28
    \subitem task, 23
  \item learning
    \subitem by reinforcement, 16
    \subitem rate, 16
  \item Levin, L.\nobreakspace  {}A., 18, 60, 63
  \item Li, M., 17, 19, 48, 49, 57, 60, 61, 64
  \item lifetime, 8, 15
  \item limited
    \subitem environmental class, 25
  \item limits, 14
  \item Littlestone, N., 48, 60, 64
  \item Littman, M.\nobreakspace  {}L., 16, 63
  \item Lucas, J.\nobreakspace  {}R., 47, 58, 63

  \indexspace

  \item manipulation, 56
  \item Markov, 16
    \subitem $k$-th order, 29
    \subitem environment, 29
  \item maximize
    \subitem reward, 8
  \item Merhav, N., 57, 60, 62
  \item Michie, D., 13, 61, 64
  \item model
    \subitem AI$\xi$, 22
    \subitem universal, 22
  \item monitor, 15
  \item Monte Carlo, 33
  \item Moore, A.\nobreakspace  {}W., 16, 63
  \item Morgenstern, O., 6, 36, 61, 64
  \item mortal
    \subitem agents, 56
  \item most intelligent
    \subitem agent, 24
  \item Motwani, R., 60, 62

  \indexspace

  \item Neumann, J.\nobreakspace  {}V., 6, 36, 61, 64
  \item noise, 15
  \item noisy world, 15
  \item non-computable
    \subitem brain, 58
    \subitem physics, 58
  \item nondeterministic world, 15
  \item Norvig, P., 5, 13, 16, 36, 60, 61, 64
  \item number of wisdom, 58

  \indexspace

  \item objectivist, 19
  \item OnlyOne example, 27
  \item optimal
    \subitem policy, 16
  \item optimality
    \subitem AI$\xi $ model, 24
    \subitem AIXI$tl $, 52
    \subitem by construction, 25
    \subitem universal, 23, 24
  \item order relation
    \subitem effective intelligence, 51
    \subitem intelligence, 23
    \subitem universal, 24
  \item Osborne, M.\nobreakspace  {}J., 36, 55, 64
  \item output, 6
    \subitem device, 15
    \subitem word, 7
  \item output space
    \subitem choice, 15, 55

  \indexspace

  \item Pareto optimality, 24
    \subitem AI$\xi $ model, 29
  \item passive
    \subitem environment, 27
  \item Penrose, R., 47, 58, 64
  \item perception, 6
  \item perceptions
    \subitem concurrent, 55
  \item perfect, 15
  \item physical random processes, 19
  \item physics
    \subitem non-computable, 58
    \subitem quantum, 15
    \subitem wave function collapse, 59
  \item Pintado, X., 60, 64
  \item policy, 7, 16
    \subitem consistent, 23
    \subitem extended chronological, 50
    \subitem inconsistent, 23
    \subitem optimal, 16
    \subitem restricted class, 33
    \subitem self-optimizing, 25
  \item policy iteration, 16
  \item posterization, 28
  \item prediction
    \subitem expert advice, 32
  \item prefix property, 7
  \item prequential approach, 17
  \item probabilistic
    \subitem environment, 9
  \item probability
    \subitem distribution, 11
  \item probability distribution, 11
    \subitem conditional, 11
  \item problem
    \subitem horizon, 30
    \subitem relevant, 28
    \subitem solvable, 23
  \item program
    \subitem extended chronological, 50
  \item proof, 57
  \item pseudo-passive
    \subitem environment, 26, 27

  \indexspace

  \item quantum physics, 15

  \indexspace

  \item random
    \subitem action, 33
  \item real
    \subitem environment, 56
  \item recursive formulation, 11
  \item reduction
    \subitem state space, 16
  \item reflex, 56
  \item reinforcement learning, 16
  \item relevant
    \subitem problem, 28
  \item restricted
    \subitem concept class, 25
  \item restricted domains, 16
  \item reward, 6
    \subitem average, 31
    \subitem future, 8
    \subitem internal, 56
    \subitem maximize, 8
    \subitem total, 8
  \item Rissanen, J.\nobreakspace  {}J., 19, 57, 60, 64
  \item robots
    \subitem autonomous, 56
  \item RSA
    \subitem cryptography, 55
  \item Rubenstein, A., 36, 55, 64
  \item Russell, S.\nobreakspace  {}J., 5, 13, 16, 36, 60, 61, 64

  \indexspace

  \item scaling
    \subitem AIXI down, 57
  \item Schmidhuber, J., 48, 58, 60, 61, 63--65
  \item Schmidt, M., 60, 64
  \item self-optimization, 24
  \item self-optimizing
    \subitem policy, 25
  \item self-tunability, 24
  \item semimeasure
    \subitem universal, time-limited, 48
  \item separability
    \subitem concepts, 26
  \item sequence, 7
    \subitem training, 58
  \item set
    \subitem prefix-free, 7
  \item Shoenfield, J.\nobreakspace  {}R., 53, 65
  \item Smith, C.\nobreakspace  {}H., 61
  \item Solomonoff, R.\nobreakspace  {}J., 20, 42, 58, 60, 61, 65
  \item solvable
    \subitem problem, 23
  \item state
    \subitem environmental, 16
    \subitem internal, 6
  \item stationarity, 16
  \item stationary
    \subitem environment, 29
  \item string
    \subitem empty, 7
    \subitem length, 7
  \item strings, 7
  \item structure
    \subitem AI$\xi $ model, 33
  \item subjectivist, 19
  \item suboptimal
    \subitem decision, 28
  \item Sutton, R.\nobreakspace  {}S., 6, 16, 60, 61, 64

  \indexspace

  \item tape
    \subitem bidirectional, 7
    \subitem unidirectional, 7
  \item task
    \subitem learnable, 23
  \item theorem provers, 57
  \item theory, \see{particular theory}{55}
  \item time bound
    \subitem AIXI$tl$, 52
  \item Tirole, J., 36, 62
  \item training
    \subitem sequence, 58
  \item Tsitsiklis, J.\nobreakspace  {}N., 6, 16, 31, 60, 61
  \item Turing machine, 7
    \subitem chronological, 8
    \subitem head, 7
  \item typing monkeys, 48

  \indexspace

  \item Ullman, J.\nobreakspace  {}D., 60, 62
  \item unbiasedness, 24
  \item underline, 11
  \item uniform
    \subitem convergence, 29
    \subitem environment, 29
  \item universal
    \subitem AI$\xi$ model, 22
    \subitem discounting, 31
    \subitem generalized prior, 22
    \subitem optimality, 23, 24
    \subitem order relation, 24
    \subitem time-limited semimeasure, 48
  \item universe, 16
  \item utility, 8
    \subitem expected, 16

  \indexspace

  \item Valiant, L.\nobreakspace  {}G., 60, 65
  \item valid approximation
    \subitem value, 51
  \item value
    \subitem bound, 26
    \subitem bounds, 56
    \subitem justification, 53
    \subitem valid approximation, 51
  \item value iteration, 16
  \item Varaiya, P.\nobreakspace  {}P., 24, 26, 63
  \item video camera, 14
  \item Vit\'anyi, P. M.\nobreakspace  {}B., 17, 19, 48, 49, 57, 60, 61,
    64
  \item vote
    \subitem best, 50
    \subitem democratic, 48
  \item Vovk, V.\nobreakspace  {}G., 48, 60, 65

  \indexspace

  \item Warmuth, M.\nobreakspace  {}K., 48, 60, 64
  \item Watkins, C., 60, 65
  \item wave function collapse, 59
  \item Wiering, M.\nobreakspace  {}A., 61, 65

  \indexspace

  \item Zhao, J., 61, 65

\end{theindex}
\end{footnotesize}

\end{document}